\documentclass[runningheads]{llncs}
\usepackage[T1]{fontenc}
\usepackage{accv}

\usepackage{graphicx}
\usepackage{graphicx}
\usepackage{booktabs}

\usepackage[accsupp]{axessibility}  %

\usepackage{graphicx}
\usepackage{amsmath}
\usepackage{booktabs}

\usepackage{epsfig}
\usepackage{amssymb}
\usepackage[dvipsnames]{xcolor}
\usepackage{colortbl}

\usepackage[justification=centering]{subcaption}
\usepackage{multirow}
\usepackage{comment}
\usepackage{kotex}

\usepackage{xspace}
\newcommand{\eg}{\textit{e.g.}\xspace}
\newcommand{\ie}{\textit{i.e.}\xspace}

\definecolor{dark-gray}{gray}{0.45}
\newcommand\ours{\texttt{MatchMe}\xspace}

\usepackage[pagebackref,breaklinks,colorlinks,citecolor=accvblue]{hyperref}
\usepackage{orcidlink}

\renewcommand\footnotemark{}

\begin{document}
\title{Match me if you can: Semi-Supervised Semantic Correspondence Learning with Unpaired Images}
\titlerunning{MatchMe}
\author{
Jiwon Kim\inst{1,2}$^*$\thanks{$^*$ Work done while at NAVER AI Lab, currently at LG AI Research. \\ Correspondence to Dongyoon Han.} \ \
Byeongho Heo\inst{1} \ \ 
Sangdoo Yun\inst{1} \ \
Seungryong Kim\inst{3} \ \
Dongyoon Han\inst{1}
}
\authorrunning{Kim et al.}
\institute{
$^{1}$ NAVER AI Lab \ \  $^{2}$ LG AI Research \ \ $^{3}$ KAIST
}
\maketitle              %
\begin{abstract}
Semantic correspondence methods have advanced to obtaining high-quality correspondences employing complicated networks, aiming to maximize the model capacity. However, despite the performance improvements, they may remain constrained by the scarcity of training keypoint pairs, a consequence of the limited training images and the sparsity of keypoints. This paper builds on the hypothesis that there is an inherent data-hungry matter in learning semantic correspondences and uncovers the models can be more trained by employing densified training pairs. We demonstrate a simple machine annotator reliably enriches paired key points via machine supervision, requiring neither extra labeled key points nor trainable modules from unlabeled images. Consequently, our models surpass current state-of-the-art models on semantic correspondence learning benchmarks like SPair-71k, PF-PASCAL, and PF-WILLOW and enjoy further robustness on corruption benchmarks. Our code is available at \url{https://github.com/naver-ai/matchme}.

\end{abstract}
\section{Introduction}
\label{sec:intro}
Learning dense correspondence between image pairs is a fundamental problem that facilitates many computer vision tasks~\cite{yang2019volumetric,kim2019semantic,lee2020reference,li2020online,min2021hypercorrelation,xie2021few,kokkinos2021learning}. In contrast to classical tasks, where images are captured in geometrically constrained settings such as different views of the same scene~\cite{hosni2012fast,melekhov2019dgc} or neighboring frames in a video~\cite{ilg2017flownet,sun2018pwc,hui2018liteflownet}, the semantic correspondence task~\cite{liu2010sift,bristow2015dense,hur2015generalized,ham2017proposal} finds pixel-wise visual correspondences between images containing the same object or semantic meaning. Due to these unconstrained settings, it should handle the additional challenges of large intra-class variations in appearance and background clutter. Recent methods~\cite{min2019hyperpixel,liu2020semantic,li2020correspondence,min2021convolutional,li2021probabilistic,zhao2021multi,min2020learning,cho2021semantic,kim2022transformatcher,cho2022cats++,kim2024efficient} were generally trained to fit on full-labeled datasets~\cite{ham2017proposal,min2019spair} providing limited training pairs with manually annotated keypoint pairs.
The rigorous requirements of pixel-level semantic correspondences lead to considerable time and expense in manual annotation by experts. This results in a limited quantity of available training data; we call it a data-hungry problem in semantic correspondence learning.

Various methods have focused on unsupervised strategies~\cite{rocco2017convolutional,laskar2018semi,truong2022probabilistic,kim2022semimatch,huang2022learning} to increase the amount of correspondence supervision on unlabeled data in a self-supervised or weakly-supervised way. In particular, the weakly-supervised methods attempted to solve the problem by using a cycle consistency~\cite{laskar2018semi,truong2022probabilistic}
or pseudo-labels~\cite{kim2022semimatch,huang2022learning} on real image pairs for unsupervised loss signal, but they still only rely on image pairs in the training set. The capability of heavy matching networks hinges on data quantity at first, but the training data remains significantly smaller than other computer vision tasks (\eg, 1.2M images in ImageNet-1K~\cite{russakovsky2015imagenet}). 
Therefore, we argue that previous approaches~\cite{laskar2018semi,kim2022semimatch,truong2022probabilistic,huang2022learning}, attempting to densify points for training, may not be an underlying solution for the data-hungry problem.

\begin{figure}[t!]
    \centering
    \begin{subfigure}{0.95\linewidth}
    \centering
\includegraphics[width=.9\linewidth]{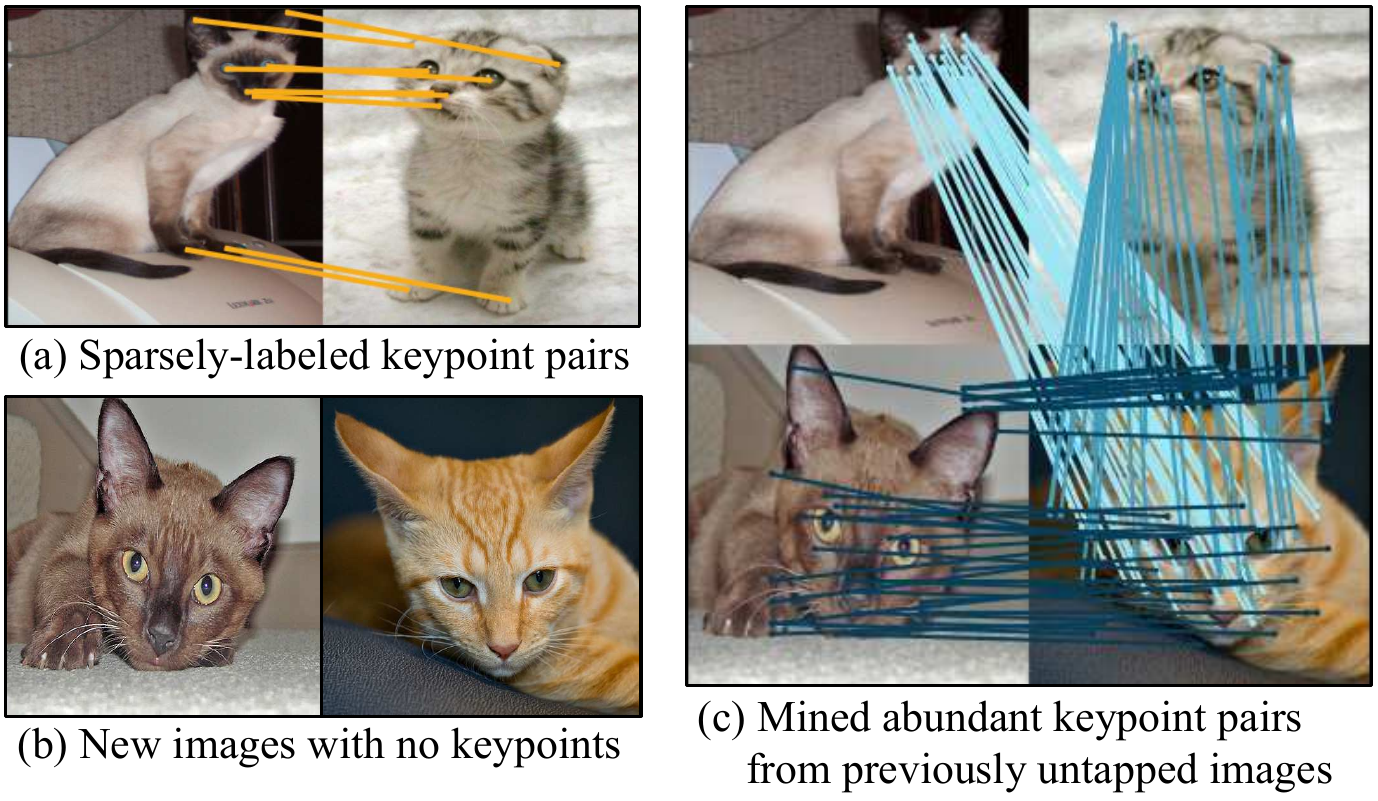}%
    \end{subfigure}
\caption{\textbf{Untapped annotation gems.} Semantic correspondence learning usually suffers from data hunger, so few sparsely paired keypoints {drawn by \color{BurntOrange} yellow lines} in labeled data inherently limit the performance. (a) Labeled images in the SPair-71k benchmark~\protect\cite{min2019spair} contain sparse manually annotated keypoint pairs. (b) Unlabeled images would become hidden supplementary sources for potentially increasing the density of pairs. (c) Newly expanded image pairs can provide abundant densified points to alleviate the underlying data-hungry matter. 
(c) illustrates that a wealth of novel machine-annotated keypoint pairs (indicated by {\color{TealBlue} blue-type lines}) are generated by simply incorporating new unlabeled images. }  
\label{fig:intro_mot}
\vspace{-.5em}
\end{figure}

In this paper, we present a fundamental approach dubbed \ours \ focusing on overcoming the insufficiency of both image and point pairs. 
We utilize unlabeled image pairs, having potentially rich semantic information that has remained unannotated. The unlabeled images newly supplied can be utilized to generate a bunch of novel pairs with originally labeled (Fig.~\ref{fig:intro_mot}(a)) or other unlabeled images (Fig.~\ref{fig:intro_mot}(b)); the newly created keypoint pairs densify labels for training (Fig.~\ref{fig:intro_mot}(c)). 
We adopt machine annotators~\cite{tarvainen2017mean,xie2020self,pham2021meta,li2022rethinking} to acquire densified labels for simplicity as well. We conjecture that a machine annotator could offer reliable labels based on the findings from~\cite{xie2020self,yun2021re}. 
Additionally, our framework allows for improved label quality by iteratively updating the annotator with the current trained model in successive training cycles. 

Our proposed method is demonstrated by applying it to recent matching architectures~\cite{cho2021semantic,cho2022cats++} to show applicability. Experimental results prove that our method is effective and achieves state-of-the-art performance on every benchmark, including PF-PASCAL~\cite{ham2017proposal}, PF-WILLOW~\cite{ham2017proposal}, and SPair-71k~\cite{min2019spair}.
\ours \ achieves state-of-the-art performance on all semantic correspondence benchmarks, showing accuracy gain of {2.0}\% and {2.4}\% on PF-WILLOW and SPair-71k (PCK@$\alpha=0.1$).%

\section{Background}
\subsection{Task Definition}
The semantic correspondence task aims to predict the matching probability $P$ between a semantically similar image pair. Given a training image pair $\mathcal{S}$ with source image $I_s \in \mathbb{R}^{H_s \times W_s}$ and target image $I_t \in \mathbb{R}^{H_t \times W_t}$, a matching function $f$ with the network parameters $\theta$ predicts $P_{s,t} = f(I_{s}, I_{t}; \theta) \in \mathbb{R}^{H_s W_s \times H_t W_t}$ by considering the feature similarities across all the points in $I_{s}$ and $I_{t}$. It minimizes the following problem with image pairs $\mathcal{S}$ and supervision $\hat{P}_{s, t} \in \mathbb{R}^{H_s W_s \times H_t W_t}$ between two images $(I_s, I_t)$:
\begin{equation}
   \hspace{-.5em}
   \mathcal{L}_\mathcal{S} =  \frac{1}{|\mathcal{S}|} \sum_{(I_s,I_t) \in \mathcal{S}} \sum_{i=1}^{H_t W_t} 
    \hat{M}_{s,t}(i){\mathcal{D}( P_{s,t}(\cdot , i) , \hat{P}_{s,t}(\cdot , i))},
    \label{eq:loss_sup}
\end{equation}
where $(\cdot, i)$ indicates the $i$-th column of a matrix and $\mathcal{D}(\cdot,\cdot)$ is a distance function. 
$\hat{M}_{s,t} \in \mathbb{B}^{H_t W_t}$ denotes a binary mask vector, in which $\hat{M}_{s,t}(i)$ corresponds to the existence of $\hat{P}_{s,t}(\cdot , i)$; we have 
\begin{equation}
\hat{M}_{s,t}({i})= \begin{cases}1, & \text { if } || \hat{P}_{s,t}(\cdot, i) ||_\infty > 0 , \\ 0, & \text { otherwise. }\end{cases}
\end{equation}
In a supervised learning framework, $(I_s, I_t)$ is tied together, so minimization in Eq.\eqref{eq:loss_sup} gives a matching function $f$ under the fixed and limited image pairs $\mathcal{S}$. 

\subsection{Motivation}
Previous methods traditionally aimed to design a novel matching network to gain a high-quality correlation map based on high-dimensional
convolutions or Transformers. However, %
they leverage complicated learning frameworks with large models and distinct data augmentations due to relying on limited annotated keypoint pairs. Sophisticated learning frameworks~\cite{kim2022semimatch,huang2022learning,truong2022probabilistic} or heavy models~\cite{rocco2020ncnet,min2021convolutional,cho2021semantic,cho2022cats++} with a matching function $f$, can fit a model to insufficient data space more; involving data augmentation methods~\cite{kim2022semimatch} diversify the images $I_s$ and $I_t$. However, they do not take into account the cardinality of the image pairs $|\mathcal{S}|$; still, the recent unsupervised methods that try to densify keypoint pairs~\cite{kim2022semimatch,huang2022learning} in each image pair do not consider the additional image pairs.

We argue using fixed annotated pairs in training inherently restricts performance; this is more likely because the annotated pairs are very sparse~\cite{ham2017proposal,min2019spair} (see Fig. 1(a)). Furthermore, the available image count is inadequate to offset the limited number of pairs. 
Given the constraints, we reframe the issue as a sample optimization problem instead of Eq.\eqref{eq:loss_sup} as:
\begin{equation}
    \hspace{-.3em}
    \min_{\theta,\,S'}  \frac{1}{|\mathcal{S'}|}\sum_{(I_s,I_t)  \in \mathcal{S'}} \sum_{i=1}^{H_t W_t} 
    \hat{M}_{s,t}(i){\mathcal{D}( P_{s,t}(\cdot , i),  \hat{P}_{s,t}(\cdot , i))},
    \label{eq:loss_intermediate}
\end{equation}
where the objective has a newly added variable $\mathcal{S'}$ and corresponding supervision. However, a direct optimization of this problem seems like an NP-hard problem. Therefore, we relax the problem by managing the image pair variable to be expanded, having a lower bound of Eq.\eqref{eq:loss_sup} by simply untying the link between source and target image pairs:
\begin{equation}
    \hspace{-.3em}
    \min_{\theta}  \frac{1}{|\mathcal{S'}|} \sum_{(I_s,I_t)  \in \mathcal{S'}} \sum_{i=1}^{H_t W_t} 
   \hat{M}_{s,t}(i){\mathcal{D}( P_{s,t}(\cdot , i), \hat{P}_{s,t}(\cdot , i))},
   \label{eq:loss_expanded}
\end{equation}
where $S' \supseteq S$. Minimizing Eq.\eqref{eq:loss_expanded} will give the trained weight with a lower value than minimizing Eq.\eqref{eq:loss_sup}, where a larger cardinality has a lower objective value. Our concern now moves on to how to acquire additional pair sets in $\mathcal{S'}$ over the original paired images $\mathcal{S}$.

\section{Method}
\begin{figure}[t]
\centering
\includegraphics[width=0.65\textwidth]{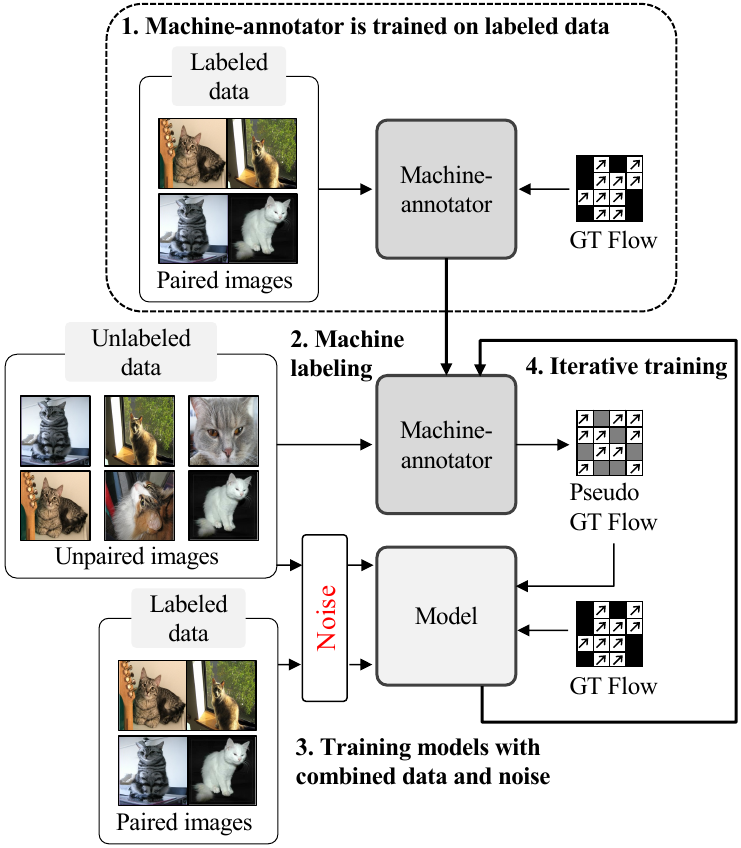}
\caption{\textbf{Schematic illustration of our method.} Unlabeled images are iteratively labeled by a progressively evolving machine annotator, where incrementally increasing {\color{Red} noise} are injected to challenge the training. Therefore, 
by learning increasingly challenging images, the model's generalization ability continues to improve.}
  \label{fig:net_arch}
  \vspace{-15pt}
\end{figure}

This section introduces how we enlarge the training pairs by effectively using unlabeled data that lack annotations, as illustrated in Fig.~\ref{fig:net_arch}. It is worth noting that our method serves as a demonstration of the intended purpose, suggesting that more complex approaches could outperform ours.
\subsection{Mining Untapped Annotation Gems}
Suppose we have a superset that contains images with or without labels having $C$ object classes;
there are $n_c$ samples in each class $c$. Ideally, $n_c(n_c - 1)$ pairs for each class could be utilized for supervised training (with labels). 
Namely, the possible image pairs for class $c$ is 
\begin{equation}
      \mathcal{U}_{c}=\{(I_s, I_t) \mid s \in c, t \in c, s \neq t\},
\end{equation}
and the set of entire pairs in a training set is $\mathcal{U}{=}\bigcup_{c=1}^{C} \mathcal{U}_c$. Due to the impracticality of labeling full-image-pairs in $\mathcal{U}$, sparsely and partially labeled keypoint-level supervisions are typically available. A set of labeled data $\mathcal{S}_c$ for supervised training consists of image pairs for each class label $c$, which can be defined as a subset of $U_c$:
\begin{equation}
      \mathcal{S}_{c}=\{(I_s, I_t) \in \mathcal{U}_c \mid || \hat{M}_{s,t} ||_\infty > 0 \},
\end{equation}
and the total set of image pairs is $\mathcal{S}{=}\{\mathcal{S}_{c}\}_{c=1}^{C}$. Prior methods~\cite{min2020learning,zhao2021multi,min2021convolutional,cho2021semantic,kim2022transformatcher,cho2022cats++,kim2024efficient} only trained with training image pairs $\mathcal{S}$ under the shared protocol based upon semantic correspondence benchmarks~\cite{ham2017proposal,min2019spair}. This bottlenecks the model's performance due to insufficient training image pairs to instantiate full dense correspondences under large intra-class variations~\cite{min2020learning,lee2021patchmatch,cho2022cats++}. Therefore, expanding the training pairs towards the entire pairs $\mathcal{U}$ by mining new annotation gems is a straightforward yet underlying solution. There can be approaches to picking which image pairs to use in training, but we alternatively involve the image set closer to the entire image pairs $\mathcal{U}$ to demonstrate that such a simple choice works. We further conjecture that this is more likely to cover different appearances and difficulty levels between image pairs, enabling the learned model to have improved generalizability.

To mine correspondence supervisions for novel training pairs having no human annotations, we utilize a self-training technique~\cite{xie2020self,yun2021re} using a machine annotator $f_\mathrm{T}$, which is trained on labeled training pairs $\mathcal{S}$, generates rough correspondences $\bar{P}_{s,t}= f_\mathrm{T}(I_s, I_t;\theta_\mathrm{T}) \in \mathbb{R}^{H_s W_s \times H_t W_t}$ on newly involved training pairs where a large fraction of pairs do not belong to the existing training pairs. The model $f$ is then trained to learn the novel correspondences with the loss:
\begin{equation}
    \mathcal{L}_\mathcal{U} = \sum_{(I_s,I_t)  \in \mathcal{U}} \sum_{i=1}^{H_t W_t} 
    \bar{M}_{s,t}(i){\mathcal{D}( P_{s,t}(\cdot , i) , \bar{P}_{s,t}(\cdot , i))},
    \label{eq:loss_unsup}
\end{equation}
where $P_{s,t} = f(I_{s}, I_{t}; \theta)$ is the predicted correspondence from the model. We use $\bar{M}_{s,t}$ as the binary mask for gating the correspondence based on a confidence-based strategy, widely used in ~\cite{xie2020unsupervised,sohn2020fixmatch,rizve2021defense,xu2021dash,zhang2021flexmatch}. By using a matching confidence $\bar{C}_{s,t}(\cdot, i)$, which defined from pair-wise scores between all locations in $I_s$ and $i$ in $I_t$ with soft-argmax operation, it is thus formulated as:
\begin{equation}
\bar{M}_{s,t}({i})= \begin{cases}1, & \text { if } || \bar{C}_{s,t}(\cdot, i) ||_\infty > \tau , \\ 0, & \text { otherwise, }\end{cases}
\end{equation}
where $\tau$ is the score threshold. The correspondences generated with high confidence encourage the model to be trained without erroneous supervision from ambiguous or textureless areas. In this work, different from the previous methods~\cite{huang2019dynamic,min2019hyperpixel,li2020correspondence,min2020learning,zhao2021multi,min2021convolutional,cho2021semantic,cho2022cats++,lee2021patchmatch}, trained on sparse keypoint pairs per limited labeled image pairs, we can significantly increase the number of image pairs available for training by including unlabeled data. Additionally, we use a large corpus of keypoint pairs, densely filling most object parts by using dense predictions from a machine annotator without the expense of manual annotation made by a human expert.

Our training objective is $\mathcal{L}=\mathcal{L}_\mathcal{U} + \lambda \mathcal{L}_\mathcal{S}$, combining the loss function in Eq.\eqref{eq:loss_sup} and Eq.\eqref{eq:loss_unsup} to leverage both the existing labeled pairs and the ones annotated by machine annotator. The weighting parameter $\lambda$ adjusts the learning dynamics between the losses; we use $\lambda{=}1$ for simplicity. Note that recent approaches~\cite{luo2024diffusion,zhang2024tale,hedlin2024unsupervised} employed features from pre-trained text-to-image diffusion models for semantic correspondence, which somewhat resembles our machine annotator. However, our model is much lighter and is trained on much smaller data under a simpler setup.

\begin{table*}[t]
    \caption{\textbf{Comparison with state-of-the-art methods on SPair-71k}. Per-class and overall PCK ($\alpha_\text{bbox}=0.1$) results are shown in the table. Numbers in bold indicate the best performance, and underlined ones are the second best. The averaged PCK of each \ours \ significantly improves the baseline by a large margin, surpassing the state-of-the-art methods. This superiority is mostly consistent across various regimes, including supervised regimes such as a recent work HCCNet, semi-supervised regimes such as SemiMatch and SCORRSAN, and unsupervised regimes such as DIFT. %
    }
    \label{tab:spair}
    \centering
    \tabcolsep=0.05em
    \resizebox{1\linewidth}{!}{
    \begin{tabular}{@{}l|c|cccccccccccccccccc@{}}
        \toprule
        Methods  & All & Aero & Bike &Bird & Boat &Bottle& Bus & Car &Cat& Chair&Cow&Dog&Horse&MBike&Person&Plant&Sheep&Train&TV   \\  
        \midrule
        HPF~\cite{min2019hyperpixel}   &28.2 &25.2 &18.9 &52.1 &15.7& 38.0& 22.8 &19.1 &52.9 &17.9& 33.0 &32.8 &20.6& 24.4 &27.9 &21.1 &15.9& 31.5& 35.6 \\
        SCOT~\cite{liu2020semantic} &35.6   &34.9 &20.7 &63.8 &21.1& 43.5& 27.3 &21.3 &63.1 &20.0& 42.9 &42.5 &31.1& 29.8 &35.0 &27.7 &24.4& 48.4& 40.8\\
        DHPF~\cite{min2020learning} & 37.3 &38.4  &23.8 & 68.3  &18.9 & 42.6 & 27.9 & 20.1  &61.6  &22.0 & 46.9 & 46.1 & 33.5 & 27.6 & 40.1  &27.6 &28.1 &49.5  &46.5 \\
        PMD~\cite{li2021probabilistic} & 37.4 &38.5& 23.7& 60.3& 18.1& 42.7& 39.3& 27.6& 60.6& 14.0& 54.0& 41.8& 34.6& 27.0& 25.2& 22.1& 29.9& 70.1& 42.8\\
        
        MMNet~\cite{zhao2021multi} & 40.9  &43.5& 27.0& 62.4& 27.3& 40.1& 50.1& 37.5& 60.0& 21.0& 56.3& 50.3& 41.3& 30.9& 19.2& 30.1& 33.2& 64.2& 43.6\\
    
        CHM~\cite{min2021convolutional} & 46.3&49.6 &29.3& 68.7& 29.7& 45.3& 48.4& 39.5& 64.9& 20.3& 60.5& 56.1& 46.0& 33.8& 44.3& 38.9& 31.4& 72.2& 55.5\\
        PMNC~\cite{lee2021patchmatch} & 50.4&54.1& 35.9& 74.9& {36.5}& 42.1& 48.8& 40.0& 72.6& 21.1& 67.6& 58.1& {50.5}& 40.1& {54.1} & 43.3& 35.7& 74.5& 59.9\\        
        CATs~\cite{cho2021semantic} & 49.9 &52.0 &34.7& 72.2& 34.3& 49.9& {57.5}& {43.6}& 66.5& {24.4}& 63.2& 56.5 & 52.0 & 42.6& 41.7& 43.0& 33.6& 72.6& 58.0\\
        TransforMatcher~\cite{kim2022transformatcher} & {53.7}&{59.2} &{39.3} &{73.0} &{41.2} &{52.5} &{66.3} &\underline{55.4} &{67.1}&{26.1}&{67.1}&{56.6} &{53.2}&{45.0} &39.9&{42.1} &{35.3}&{75.2} &{68.6} \\
        CATs++~\cite{cho2022cats++}&\underline{59.8}&{60.6}& {46.9}& \underline{82.5}&\underline{41.6}& \textbf{56.7}& \underline{65.1}& {50.4}& {72.8} & {29.2}& \underline{75.9} &\underline{65.3}& \underline{62.6}& \underline{50.9} &\underline{56.1}& \underline{54.6}& \underline{48.0} & {80.8} & \underline{75.0} \\

        SemiMatch~\cite{kim2022semimatch}&50.7&53.6 &37.0 &74.6 &32.3 &47.5 &57.7 &42.4 &67.4 &23.7& 64.2 &57.3 &51.7 &43.8 &40.4 &45.3 &33.1 &74.1 &65.9  \\
        SCORRSAN~\cite{huang2022learning} & {55.3}&{57.1} &{40.3} &{78.3} &{38.1} &{51.8} &{57.8} &{47.1} &{67.9}&{25.2} &{71.3}&{63.9} &49.3&{45.3} &{49.8}&{48.8} &{40.3}&{77.7} &{69.7}\\

        DIFT$_{sd}$ ~\cite{tang2023emergent} & {52.9}&\underline{61.2} & \textbf{53.2} &{79.5} &{31.2} &{45.3} &{39.8} &{33.3} & \textbf{77.8}&\textbf{34.7} &{70.1}&{51.5} & 57.2 &{50.6} &{41.4}&{51.9} &{46.0}&{67.6} &{59.5}\\
        HCCNet~\cite{kim2024efficient} & {54.8}&{59.9} &{40.6} &{70.5} &{39.8} &\underline{55.9} &\underline{65.1} &\textbf{56.8} &{66.6}&{25.6} &{69.2}&{59.6} & 58.7&{46.7} &{40.3}&{43.6} &{39.6}& \underline{82.2} &{65.4}\\
        \midrule
        
        \rowcolor{gray!10} 
        \rowcolor{gray!10} 
        \ours (ours) & %
        \textbf{62.0} & 
        \textbf{63.4} & \underline{51.1} & \textbf{83.2} & \textbf{44.8}& {53.1}& \textbf{66.9} &  {53.4} & \underline{74.8} & \underline{30.4} & \textbf{76.8} & \textbf{66.6} & \textbf{68.1} & \textbf{55.2} & \textbf{60.7} & \textbf{59.1} & \textbf{48.1} & \textbf{84.9} & \textbf{75.9}   
        \\
        \bottomrule
    \end{tabular}
    }
\end{table*}

\subsection{Iterative Labeling and Training}
Our framework enjoys further improvements via iterative training. Specifically, each iteration repeats the training process using the model trained in the previous iteration as a new machine annotator and trains a new model. We define ${l}$-th training iteration as $(f_\mathrm{T}^l, f^l)$, consisting of a pair of the machine annotator and in-training-model. The first generation of annotator model $f_\mathrm{T}^0$ is trained from scratch on labeled training pairs $ \mathcal{S}$. The subsequent annotator models use the model trained in the preceding generations, \ie, $f_\mathrm{T}^l = f^{l-1}$.

During training, we augment the input images of the model to diversify them further and let the model learn with more challenging ones.
We use photometric augmentation $\mathcal{N}_p$~\cite{cho2021semantic,cho2022cats++,kim2022semimatch} and geometric augmentation $\mathcal{N}_g$~\cite{rocco2017convolutional,kim2022semimatch,truong2022probabilistic}, to the source and target image for training. As a result, the model is trained on more challenging image pairs than those in previous steps, which enhances the model's generalization ability, resulting in a superior model that surpasses the performance of the previous step. It can be defined as $P_{s,t} = f(\mathcal{N}_p(I_s), \mathcal{N}_p(\mathcal{N}_g(I_t))))$. Photometric augmentation is applied to both the source and target images, while geometric augmentation is applied only to the target image, considering computational efficiency. The machine-generated labels $\bar{P}_{s,t}$ are warped to align spatial position changes by applying the same geometric transformation.

\section{Experiments} 

\subsection{Experimental Setups}
\noindent\textbf{Benchmarks.}
Experiments are conducted on three standard benchmarks for semantic correspondence learning: PF-PASCAL~\cite{ham2017proposal}, PF-WILLOW~\cite{ham2017proposal}, and SPair-71k~\cite{min2019spair} consisting of image pairs with human-annotated keypoints
from 20, 4, and 18 categories, respectively. As in previous works~\cite{han2017scnet}, we split the PF-PASCAL dataset~\cite{ham2017proposal} into about 700, 300, and 300 images for training, validation, and testing, respectively. For the SPair-71k dataset~\cite{min2019spair}, we use 53,340 for training, 5,384 for validation, and 12,234 for testing. To verify generalization capacity, the PF-WILLOW dataset~\cite{ham2017proposal} is used for testing only.

\noindent\textbf{Evaluation metric.}
Following \cite{min2019hyperpixel}, the percentage of correct keypoint (PCK@$\alpha_k$) is used for the evaluation metric by setting $\alpha_k$, a tolerance margin, having a value $\in \{0,1\}$. PCK can be computed as the ratio of correctly estimated keypoint pairs to the total number of keypoint pairs using the Euclidean distance between them within the pixel margin $\alpha_k \cdot \max (H_k, W_k)$. By setting $k \in \{\text{img}, \text{bbox}\}$, $H_k$ and $W_k$ are the width and height of either image or the object's bounding box.

\begin{figure*}[t!]
  \centering
  \footnotesize
  \begin{subfigure}[b]{0.19\linewidth}
    \centering
    \includegraphics[width=\linewidth]{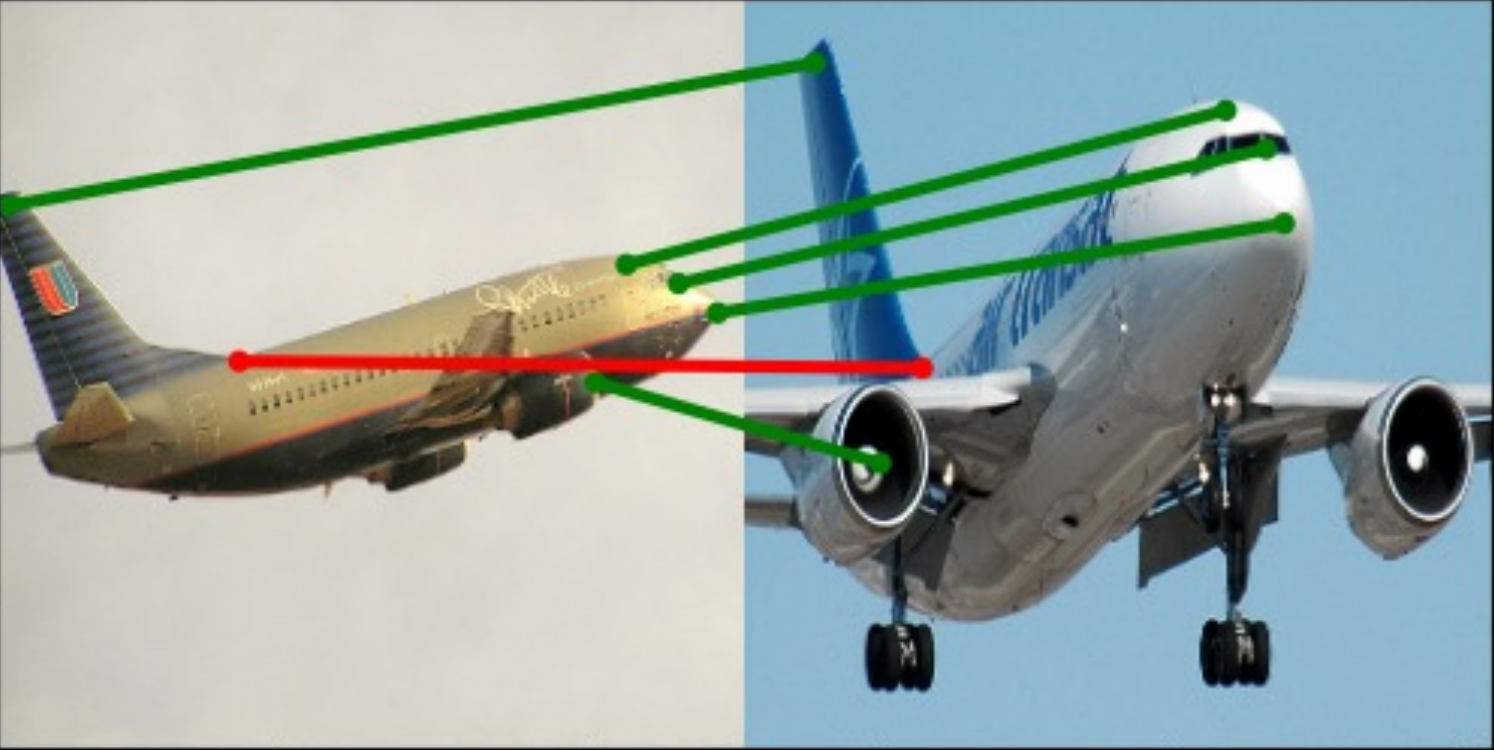}
  \end{subfigure}
  \begin{subfigure}[b]{0.19\linewidth}
    \centering
    \includegraphics[width=\linewidth]{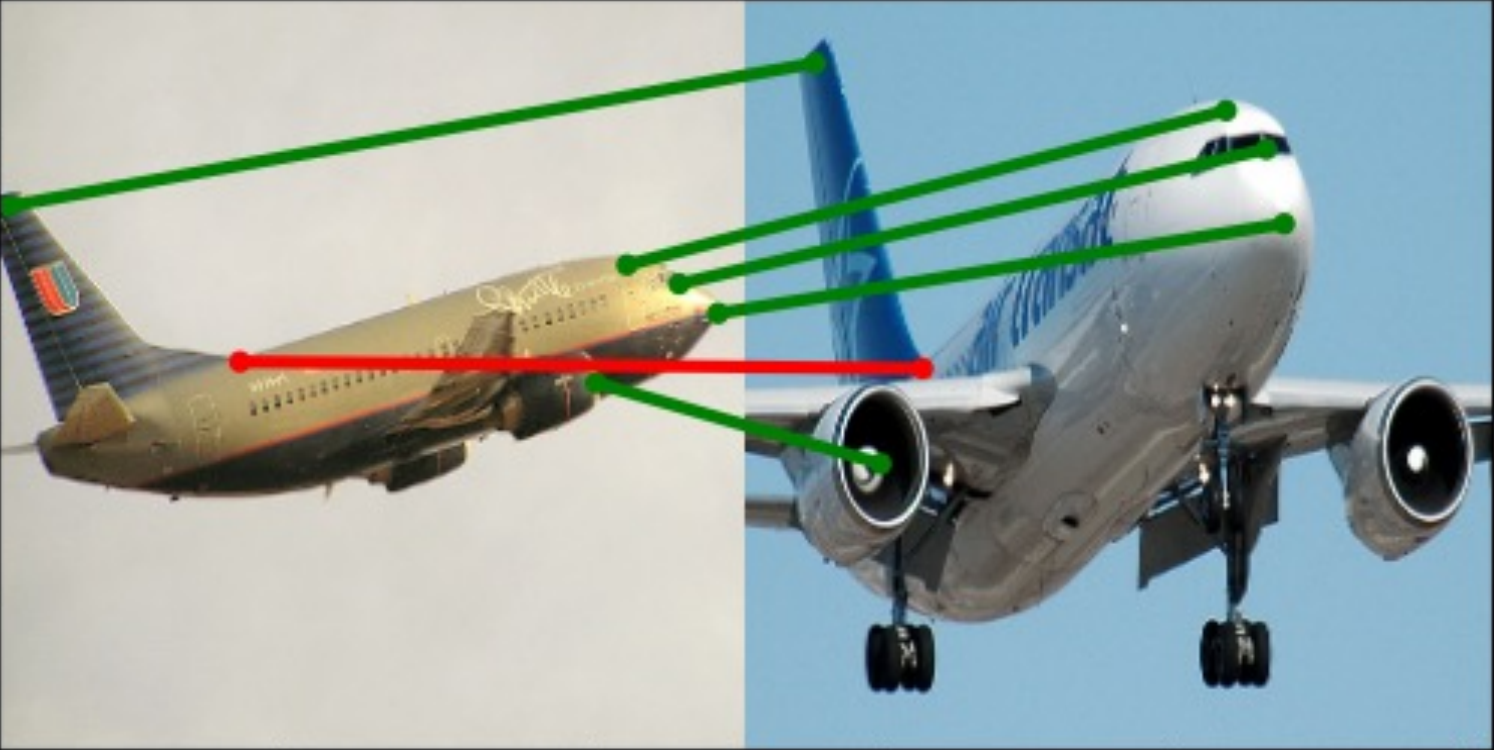}
  \end{subfigure}
  \begin{subfigure}[b]{0.19\linewidth}
    \centering
    \includegraphics[width=\linewidth]{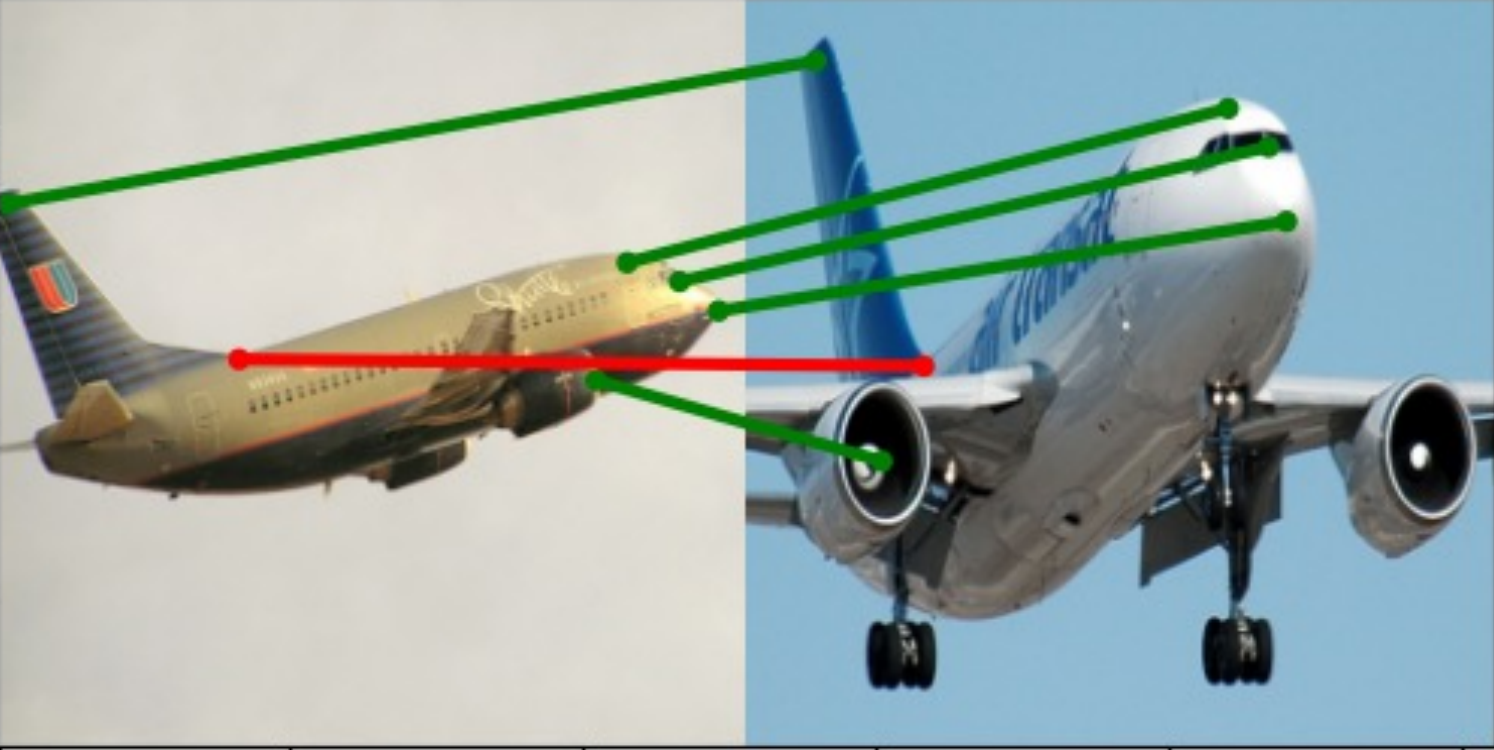}
  \end{subfigure}
  \begin{subfigure}[b]{0.19\linewidth}
    \centering
    \includegraphics[width=\linewidth]{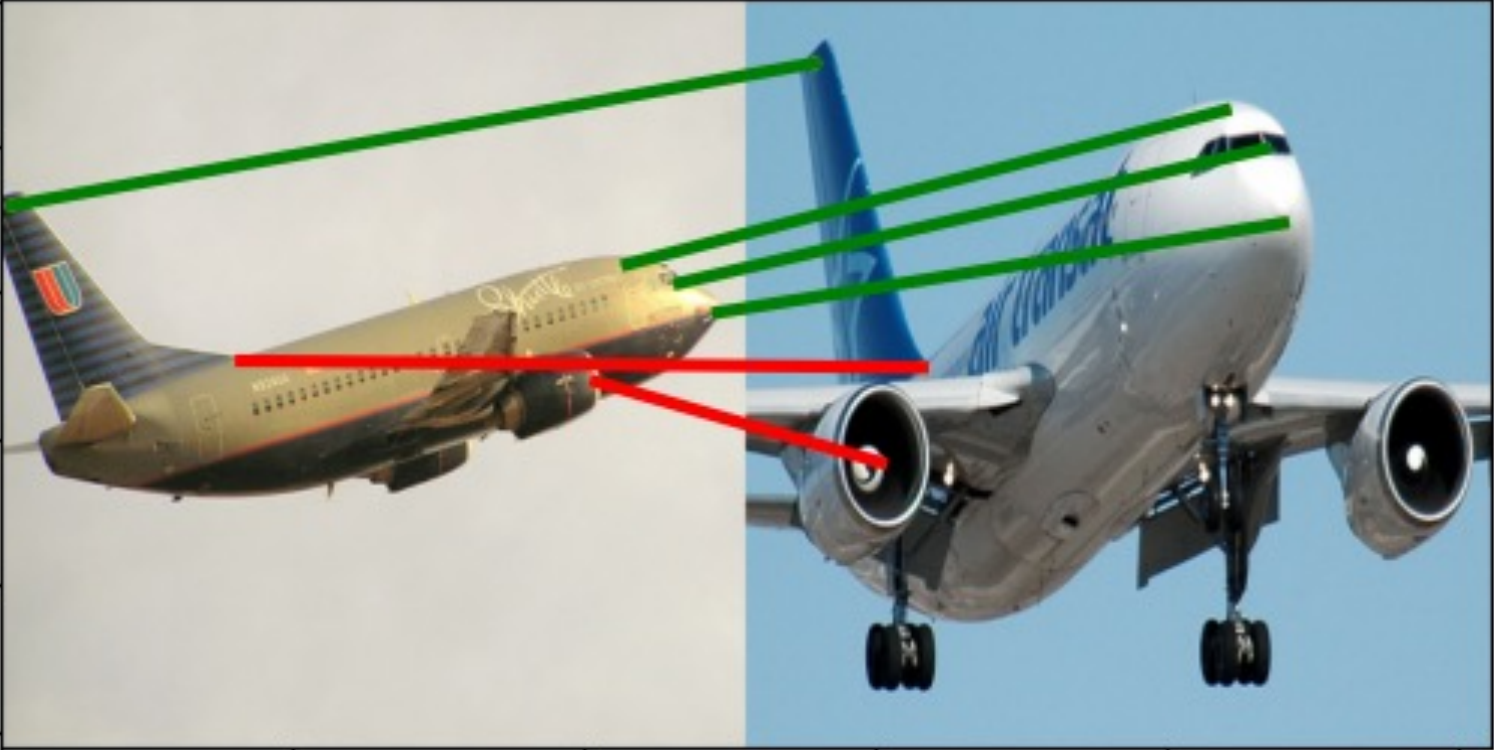}
  \end{subfigure}
    \begin{subfigure}[b]{0.19\linewidth}
    \centering
    \includegraphics[width=\linewidth]{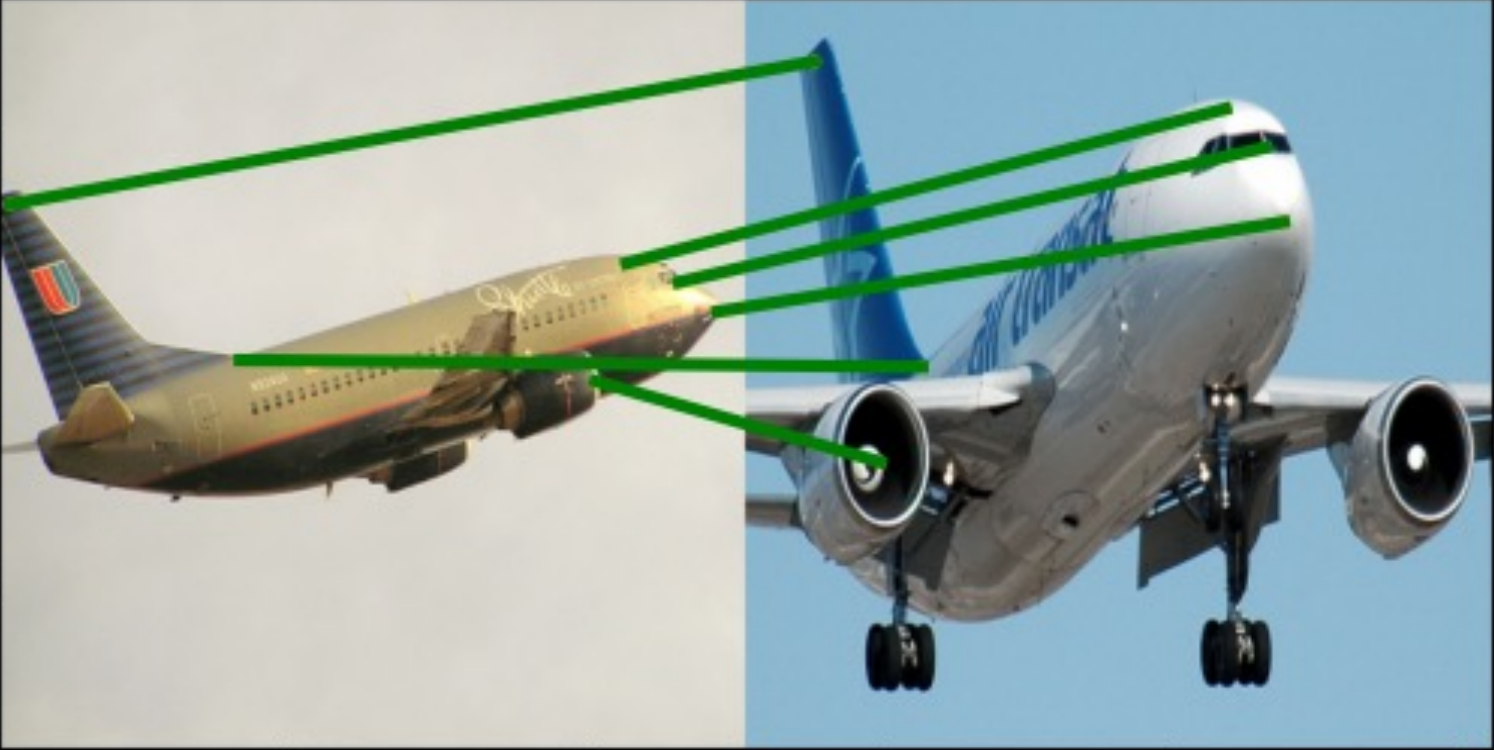}
  \end{subfigure} \\
  \begin{subfigure}[b]{0.19\textwidth}
    \centering
    \includegraphics[width=\textwidth]{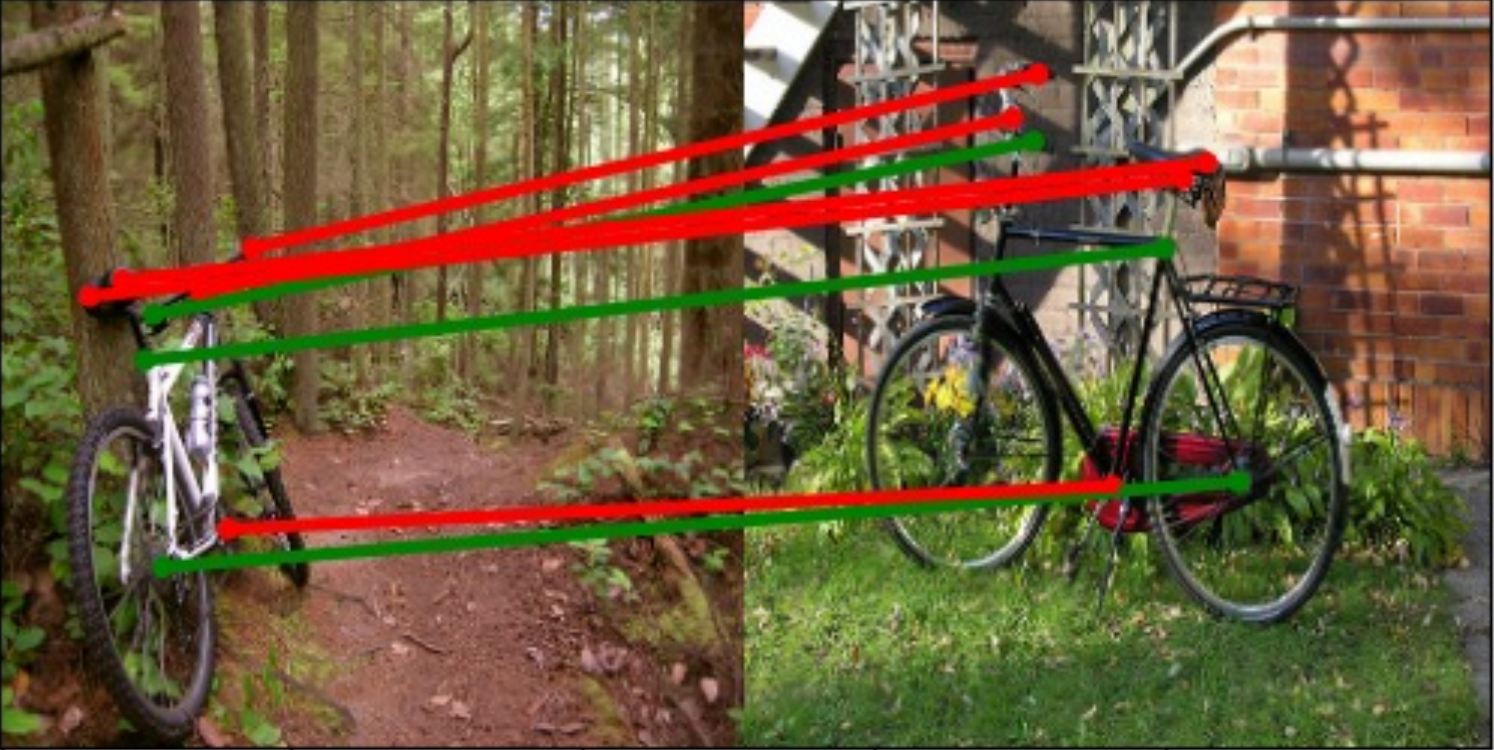}
  \end{subfigure} 
  \begin{subfigure}[b]{0.19\textwidth}
    \centering
    \includegraphics[width=\textwidth]{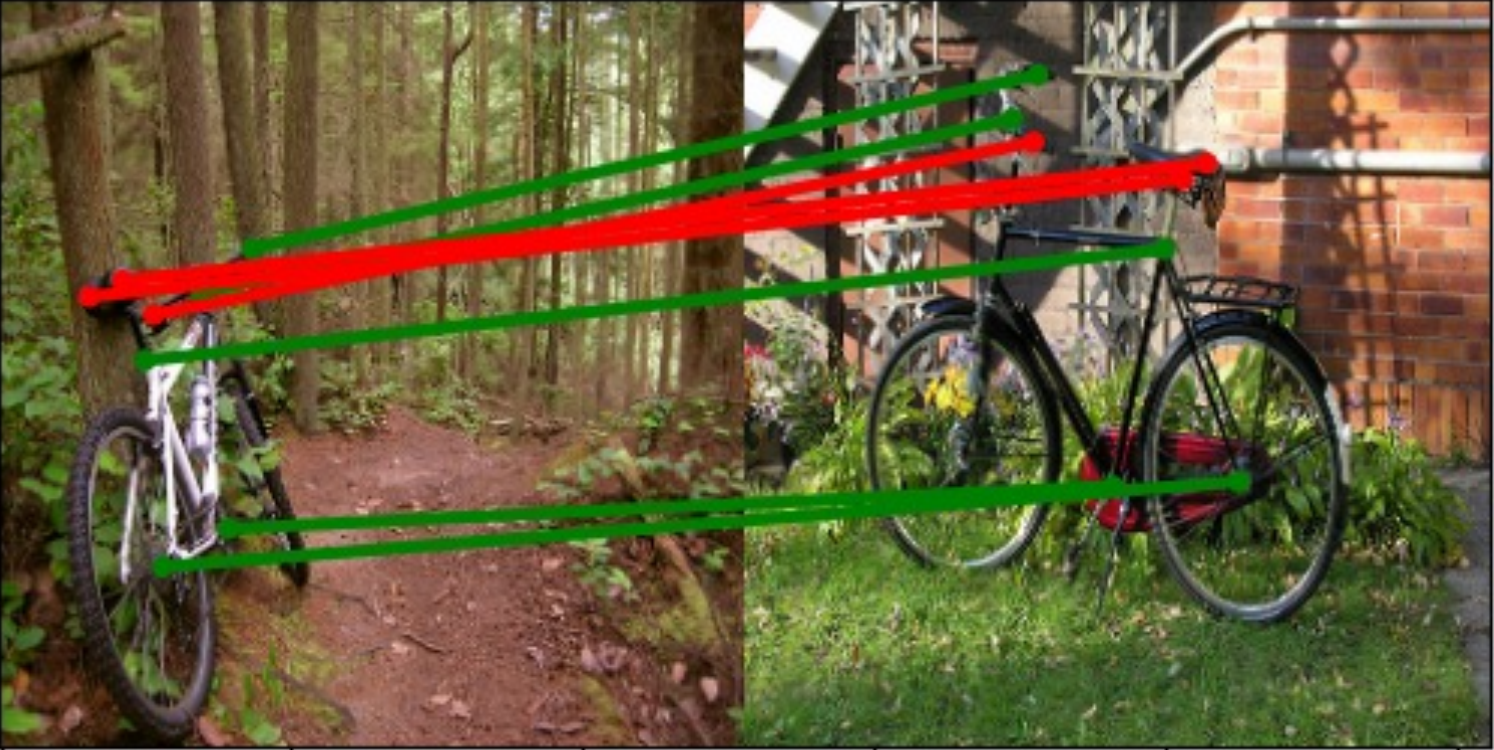}
  \end{subfigure}
  \begin{subfigure}[b]{0.19\textwidth}
    \centering
    \includegraphics[width=\textwidth]{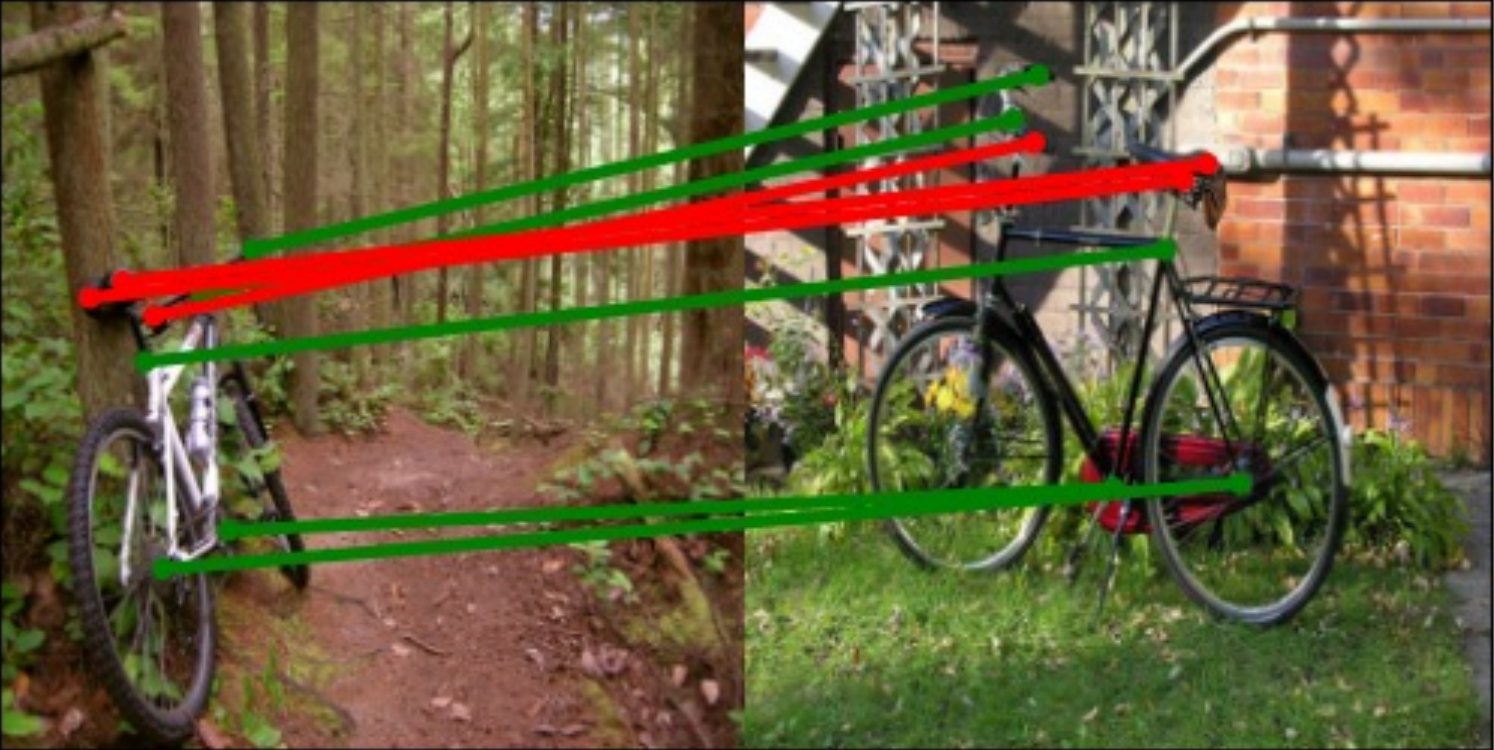}
  \end{subfigure}
  \begin{subfigure}[b]{0.19\textwidth}
    \centering
    \includegraphics[width=\textwidth]{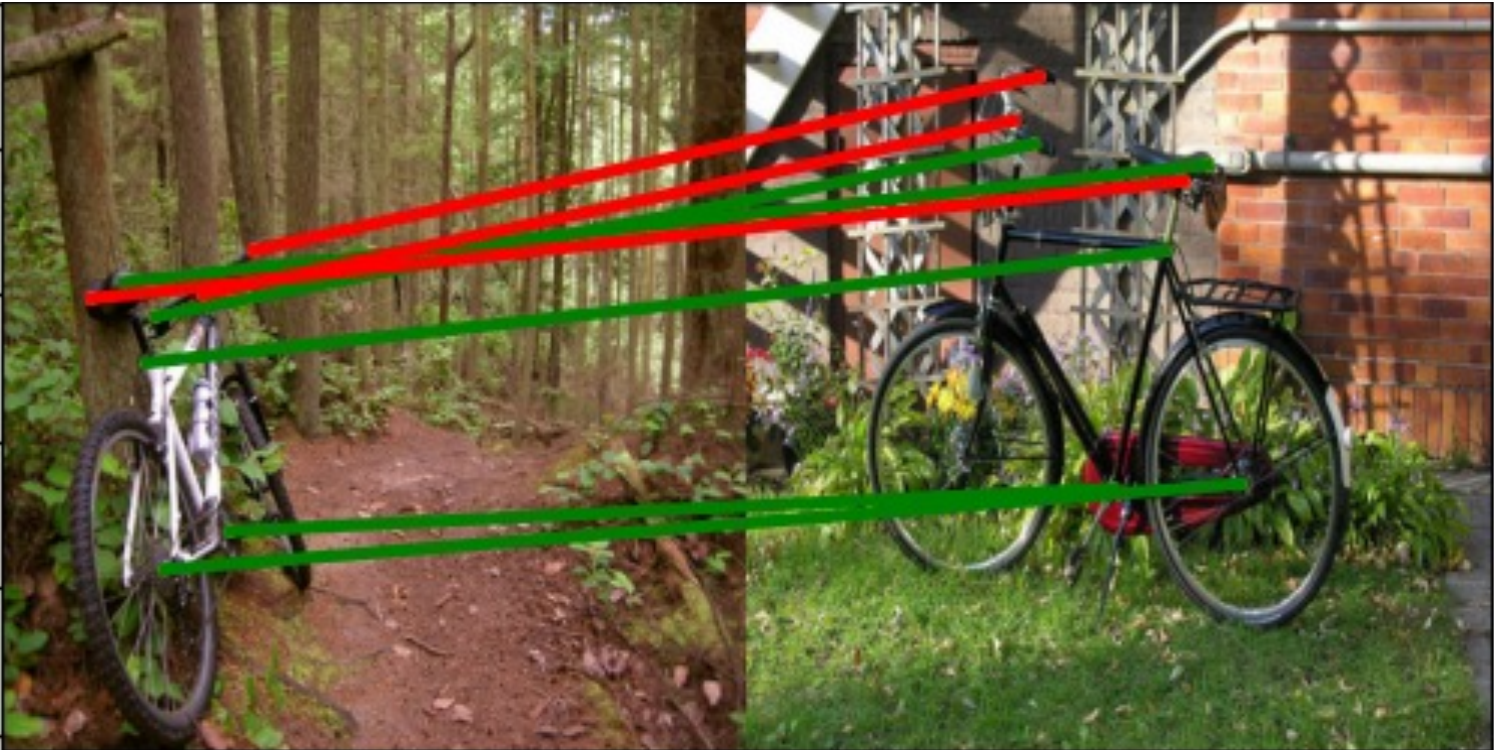}
  \end{subfigure}
    \begin{subfigure}[b]{0.19\textwidth}
    \centering
    \includegraphics[width=\textwidth]{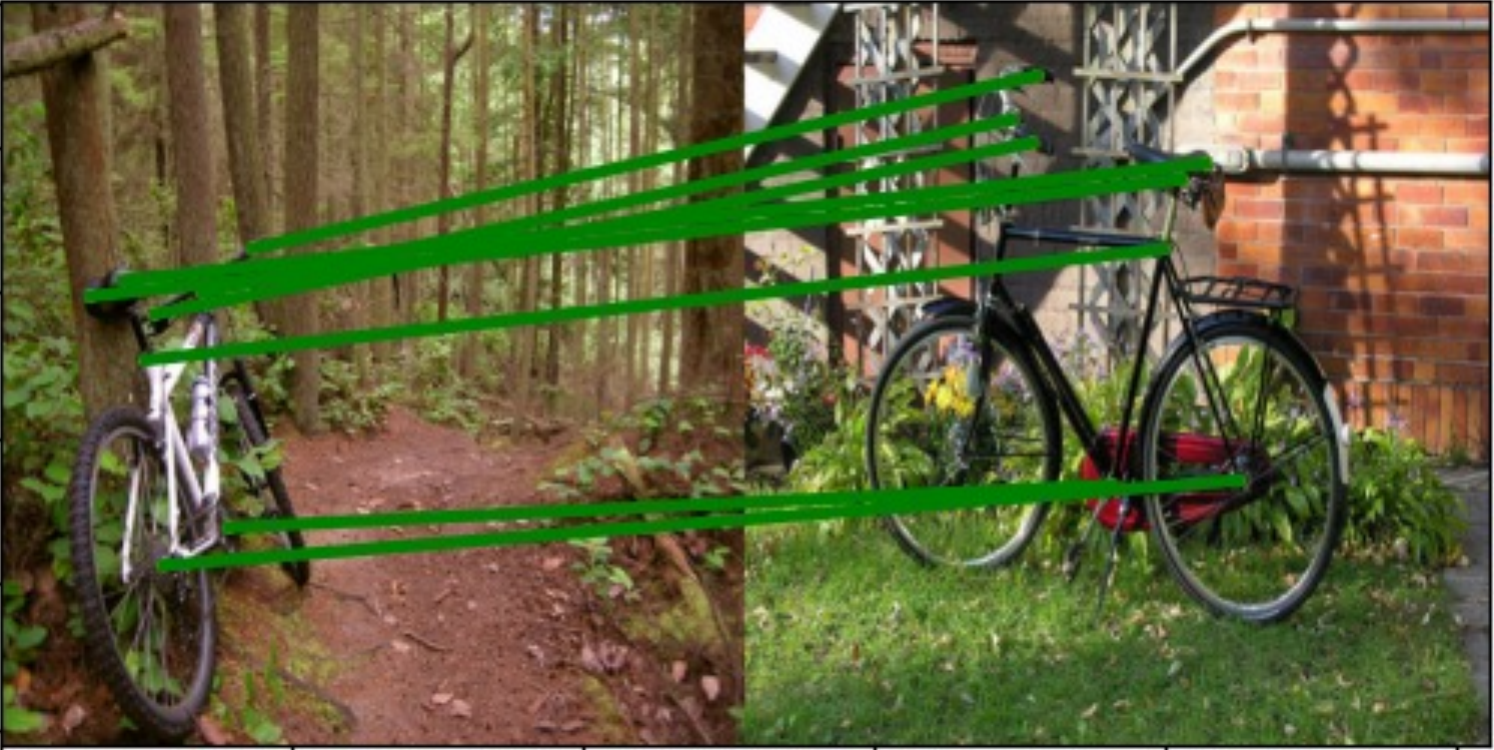}
  \end{subfigure} 
   \begin{subfigure}[b]{0.19\textwidth}
    \centering
    \includegraphics[width=\textwidth]{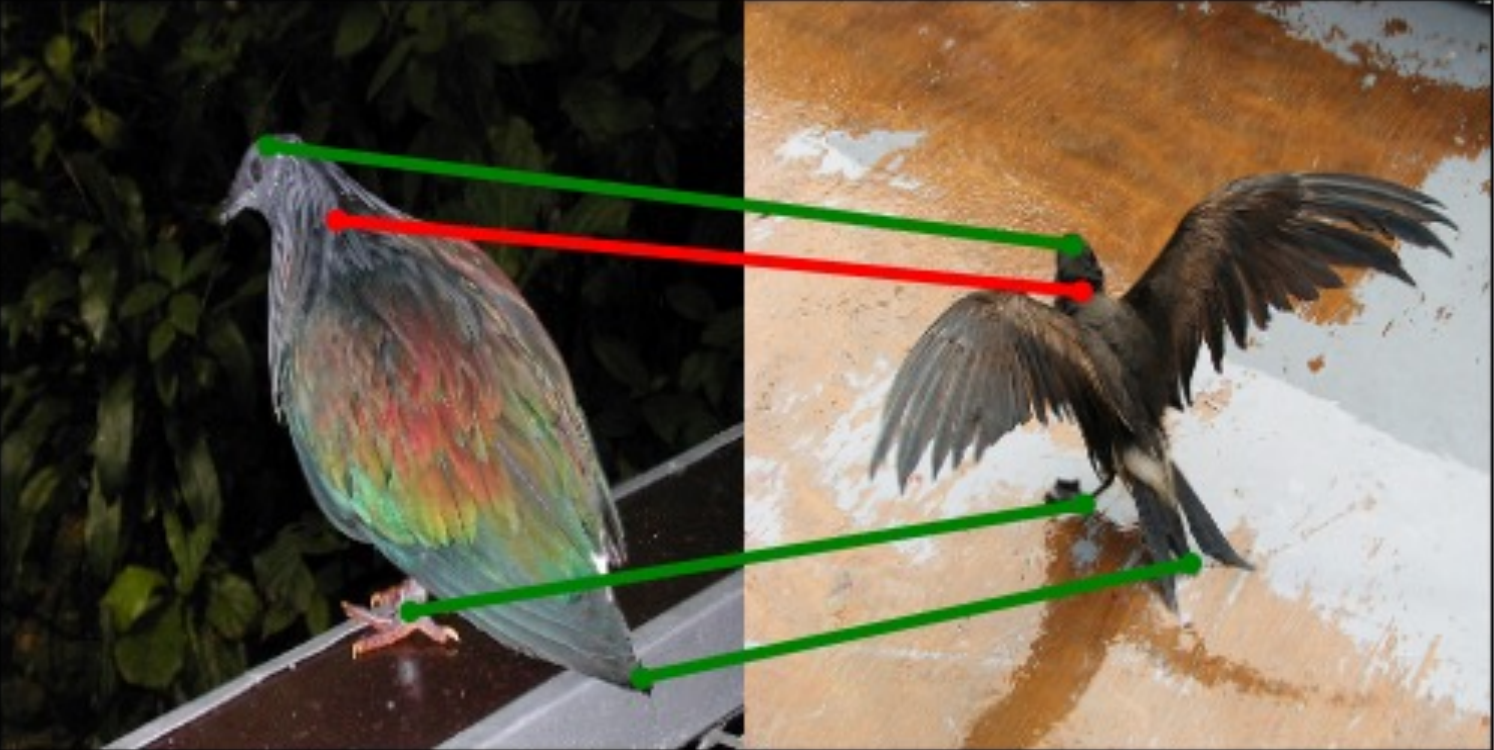}
  \end{subfigure}
  \begin{subfigure}[b]{0.19\textwidth}
    \centering
    \includegraphics[width=\textwidth]{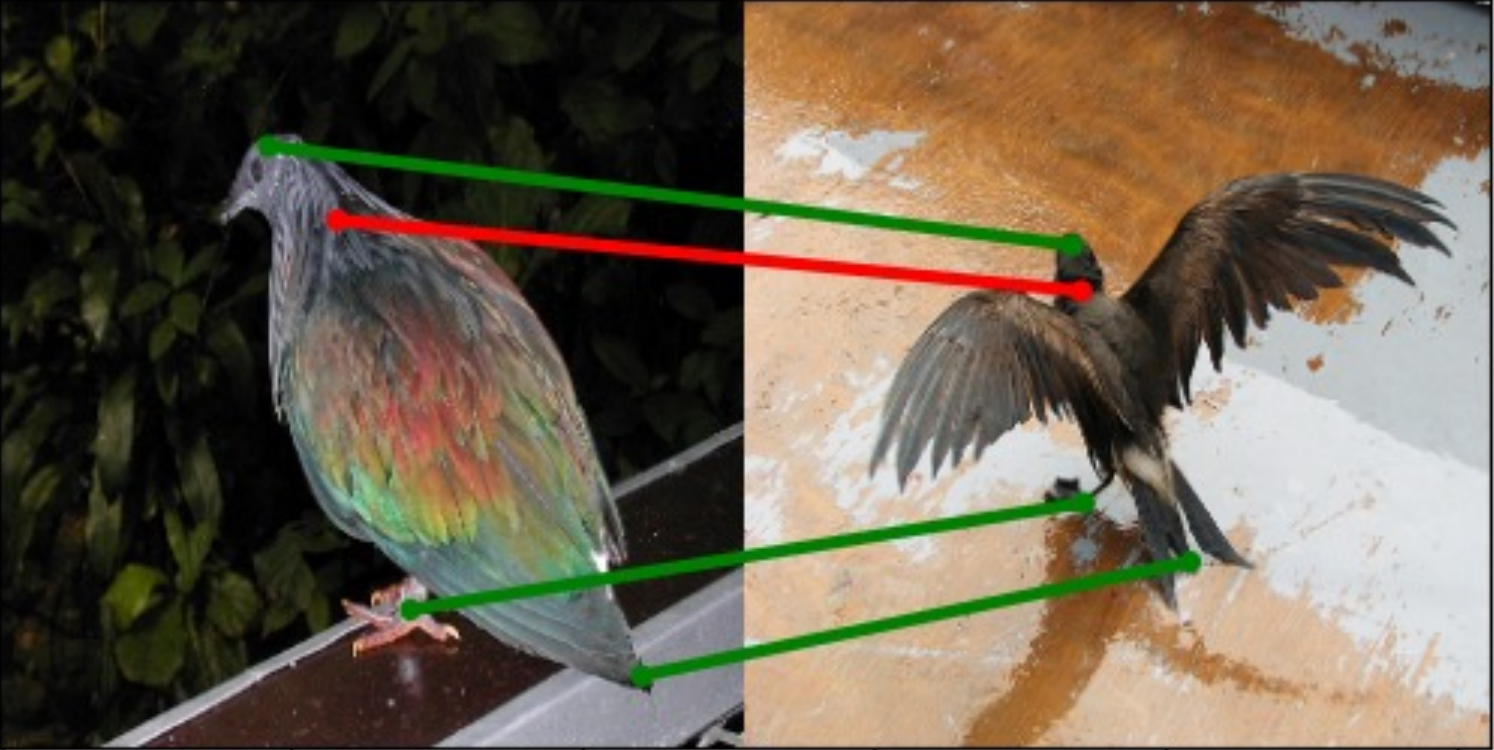}
  \end{subfigure}
  \begin{subfigure}[b]{0.19\textwidth}
    \centering
    \includegraphics[width=\textwidth]{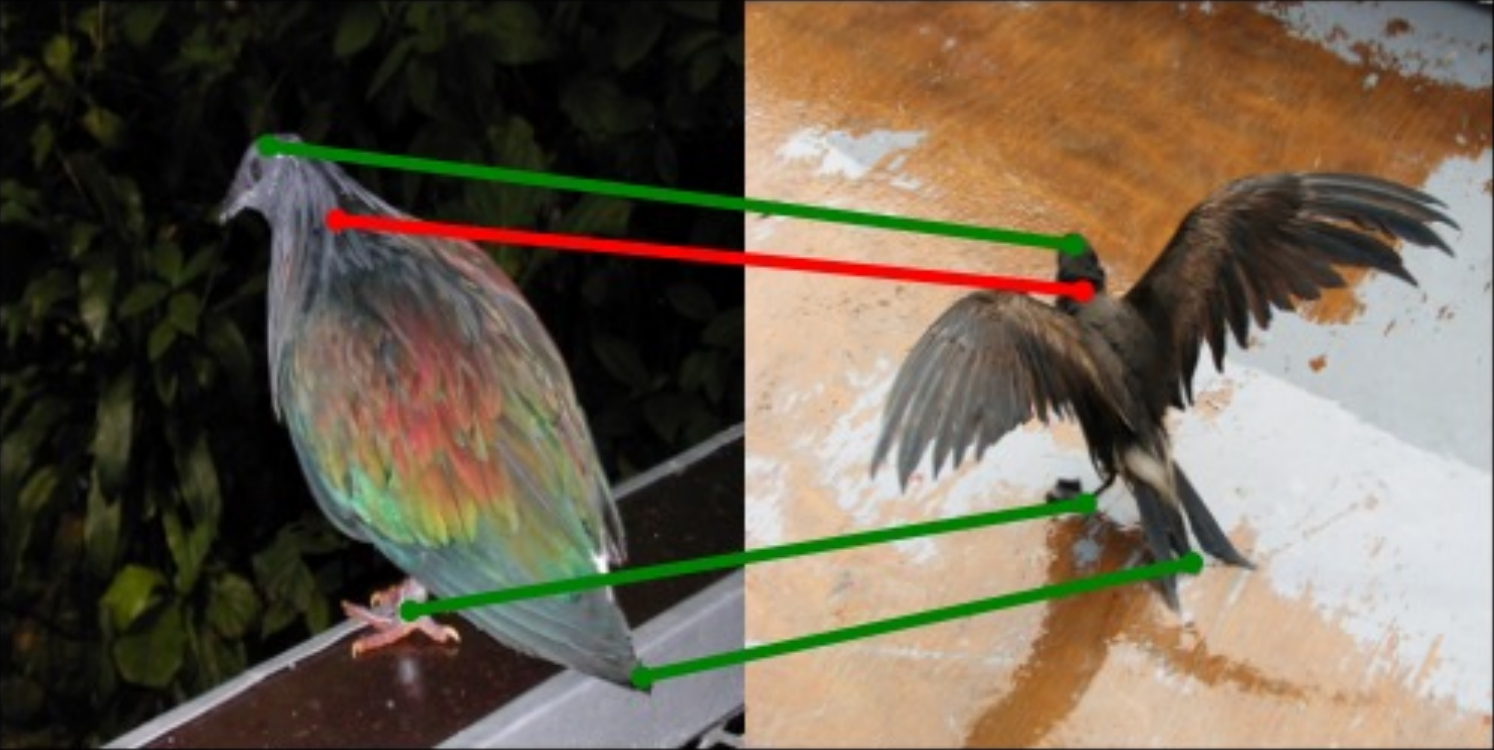}
  \end{subfigure}
  \begin{subfigure}[b]{0.19\textwidth}
    \centering
    \includegraphics[width=\textwidth]{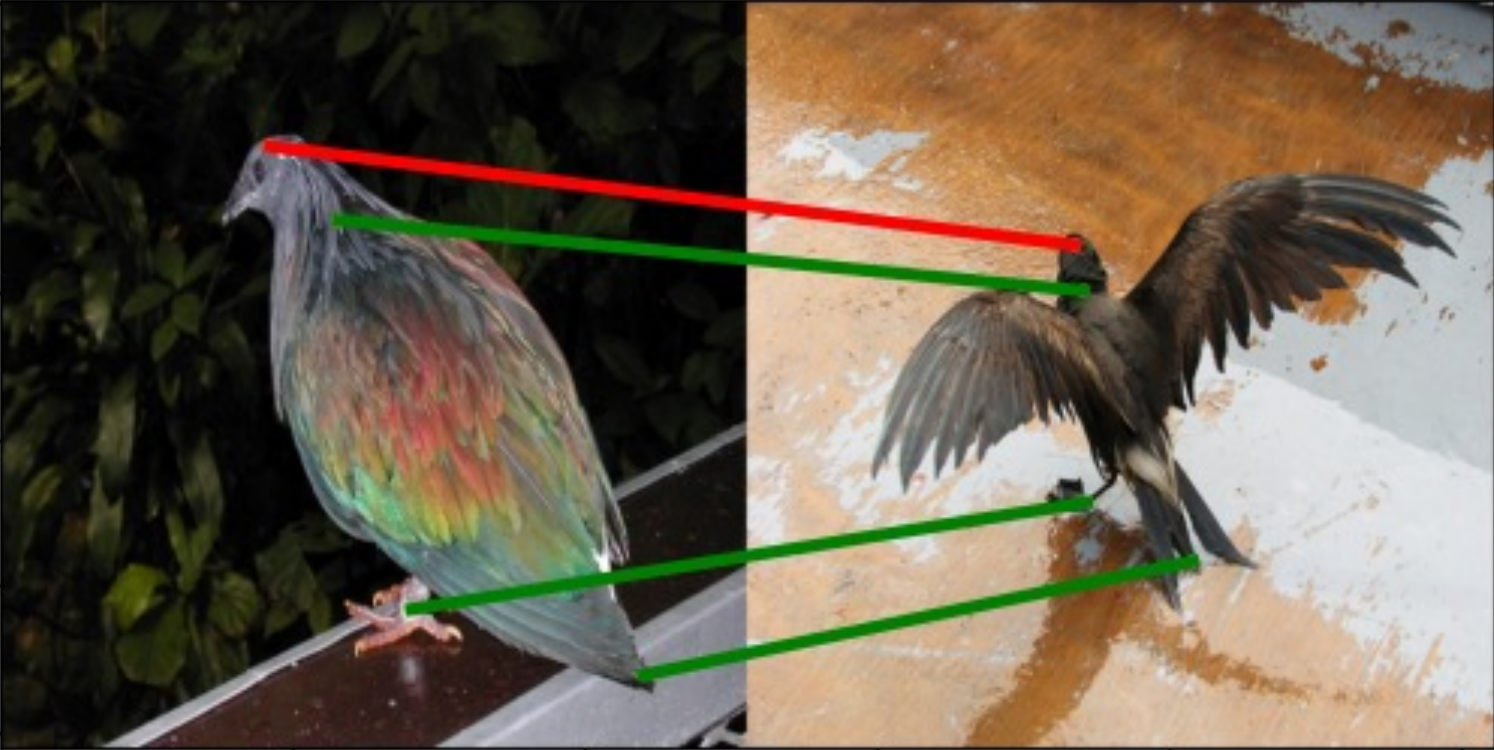}
  \end{subfigure}
    \begin{subfigure}[b]{0.19\textwidth}
    \centering
    \includegraphics[width=\textwidth]{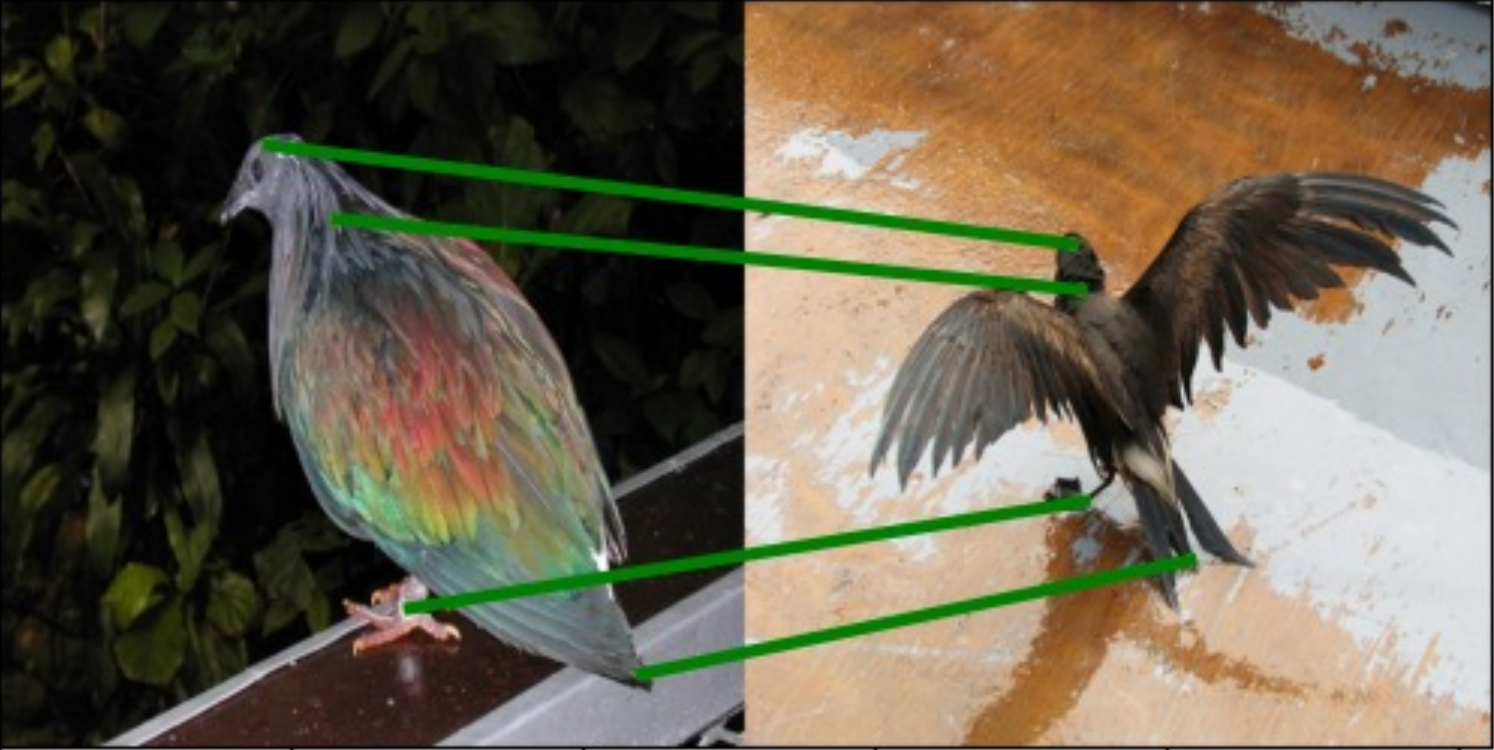}
  \end{subfigure} \\
    \begin{subfigure}[b]{0.19\textwidth}
    \centering
    \includegraphics[width=\textwidth]{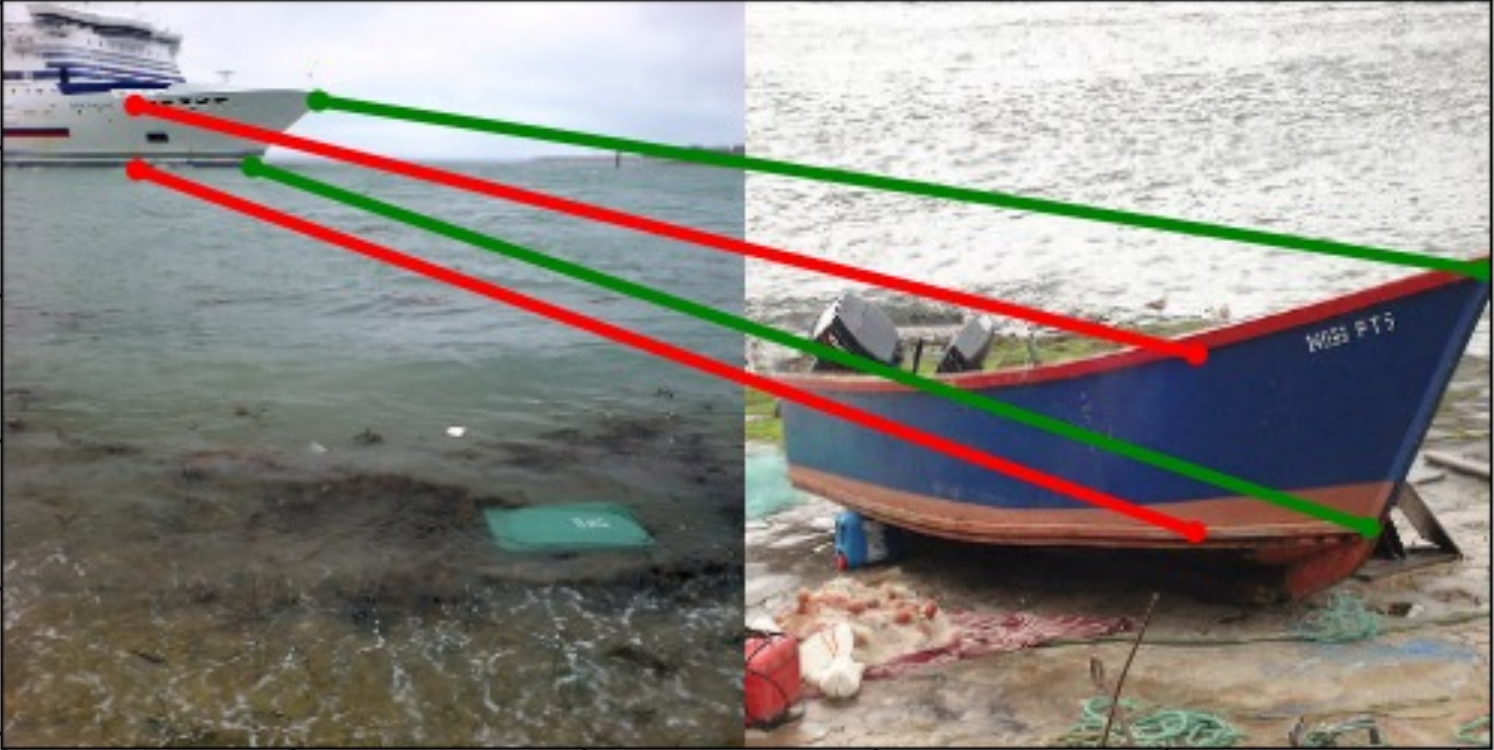}
  \end{subfigure}
  \begin{subfigure}[b]{0.19\textwidth}
    \centering
    \includegraphics[width=\textwidth]{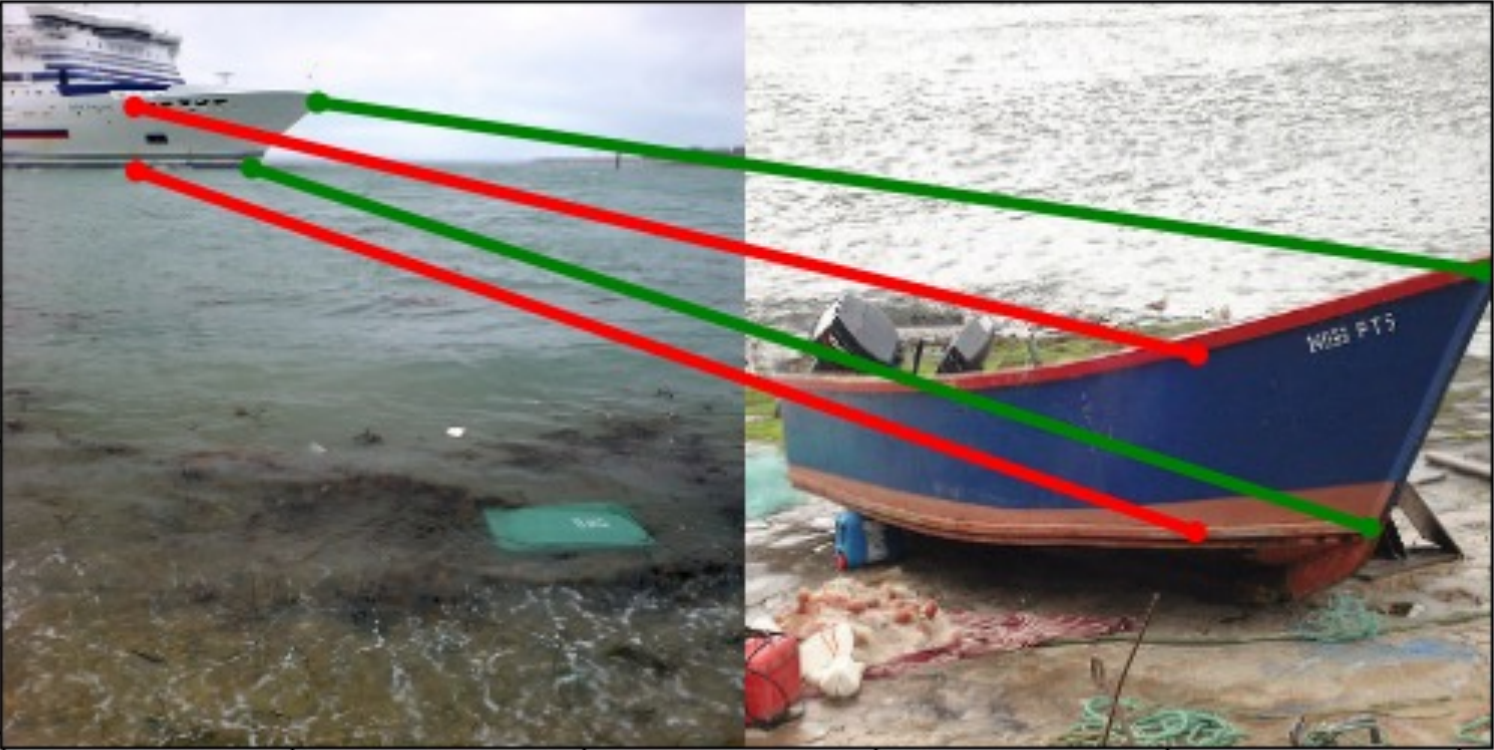}
  \end{subfigure}
  \begin{subfigure}[b]{0.19\textwidth}
    \centering
    \includegraphics[width=\textwidth]{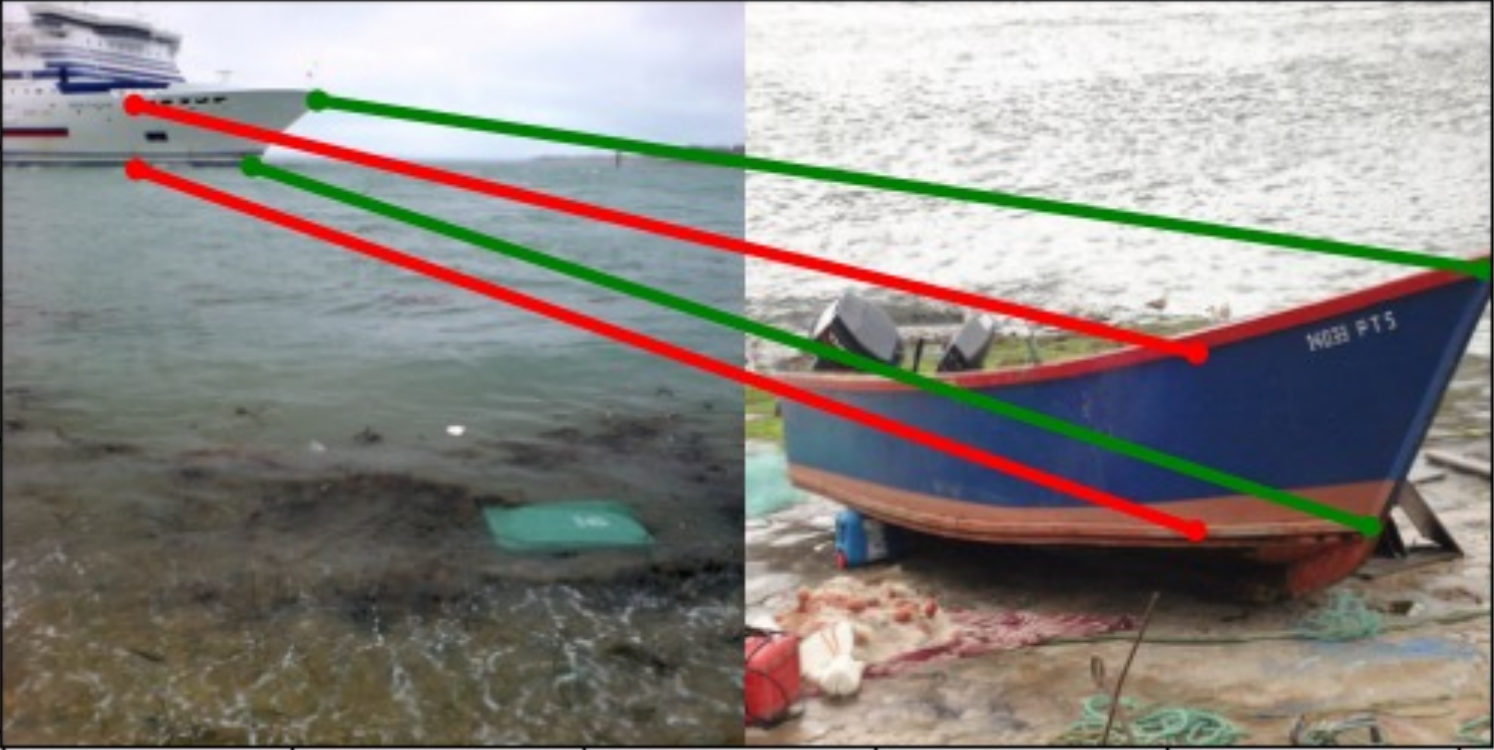}
  \end{subfigure}
  \begin{subfigure}[b]{0.19\textwidth}
    \centering
    \includegraphics[width=\textwidth]{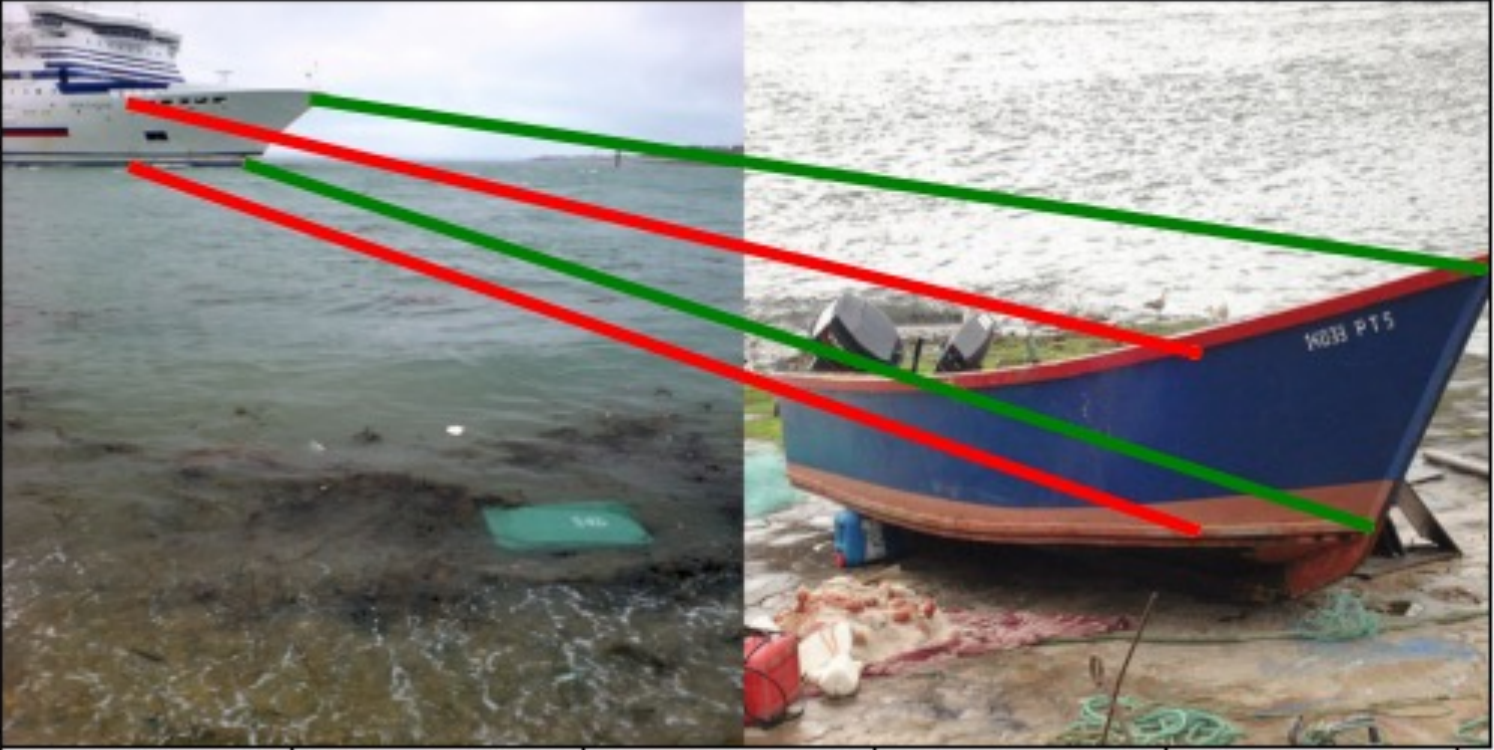}
  \end{subfigure}
    \begin{subfigure}[b]{0.19\textwidth}
    \centering
    \includegraphics[width=\textwidth]{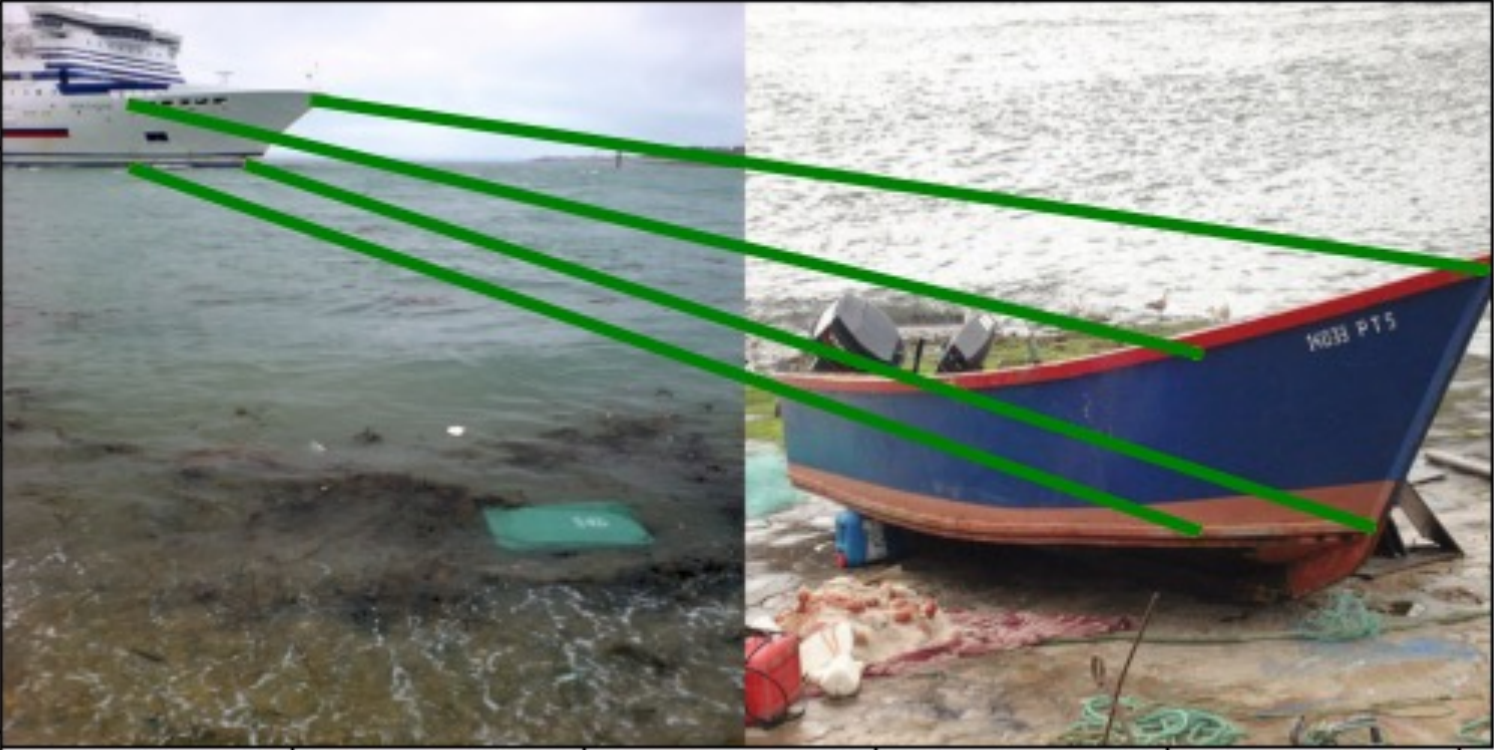}
  \end{subfigure} \\
     \begin{subfigure}[b]{0.19\textwidth}
    \centering
    \includegraphics[width=\textwidth]{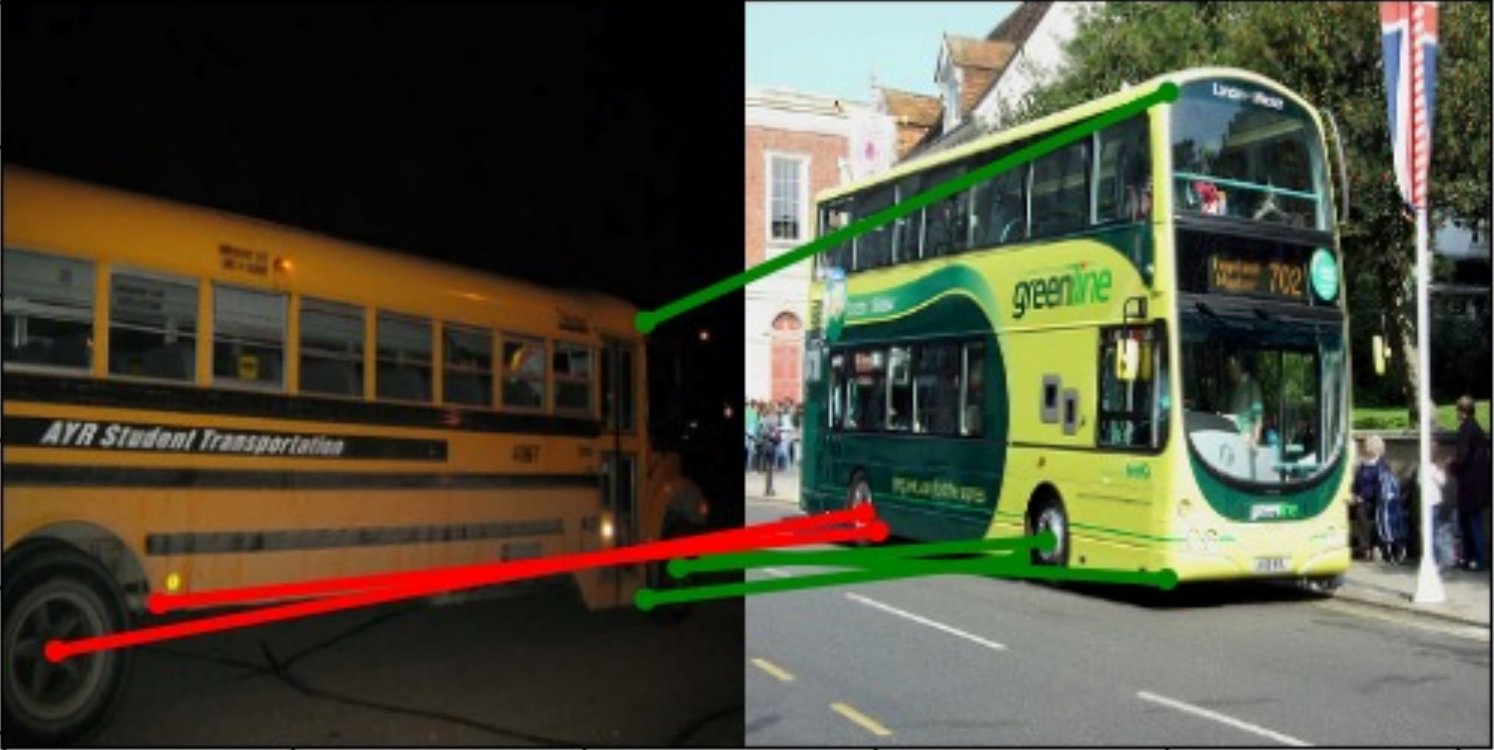}
  \end{subfigure}
  \begin{subfigure}[b]{0.19\textwidth}
    \centering
    \includegraphics[width=\textwidth]{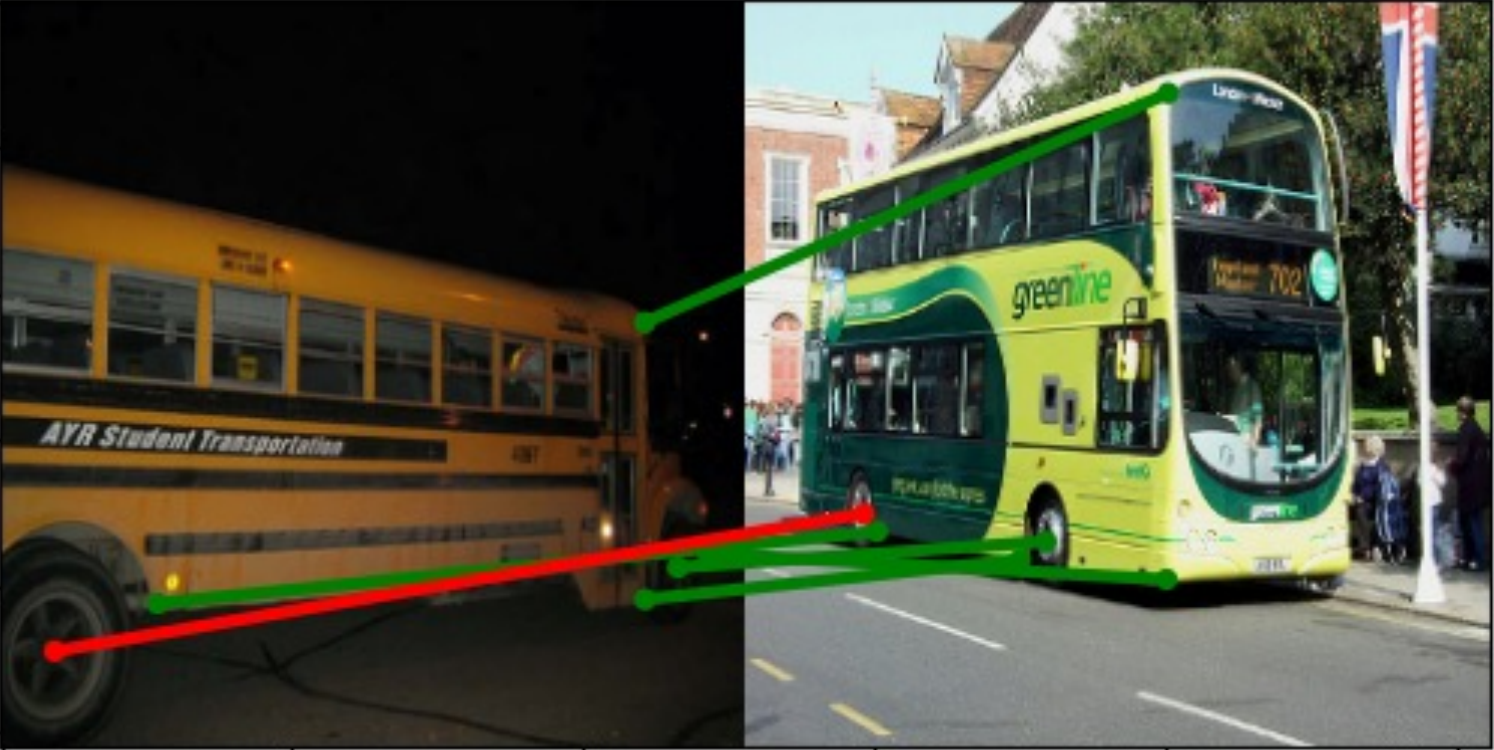}
  \end{subfigure}
  \begin{subfigure}[b]{0.19\textwidth}
    \centering
    \includegraphics[width=\textwidth]{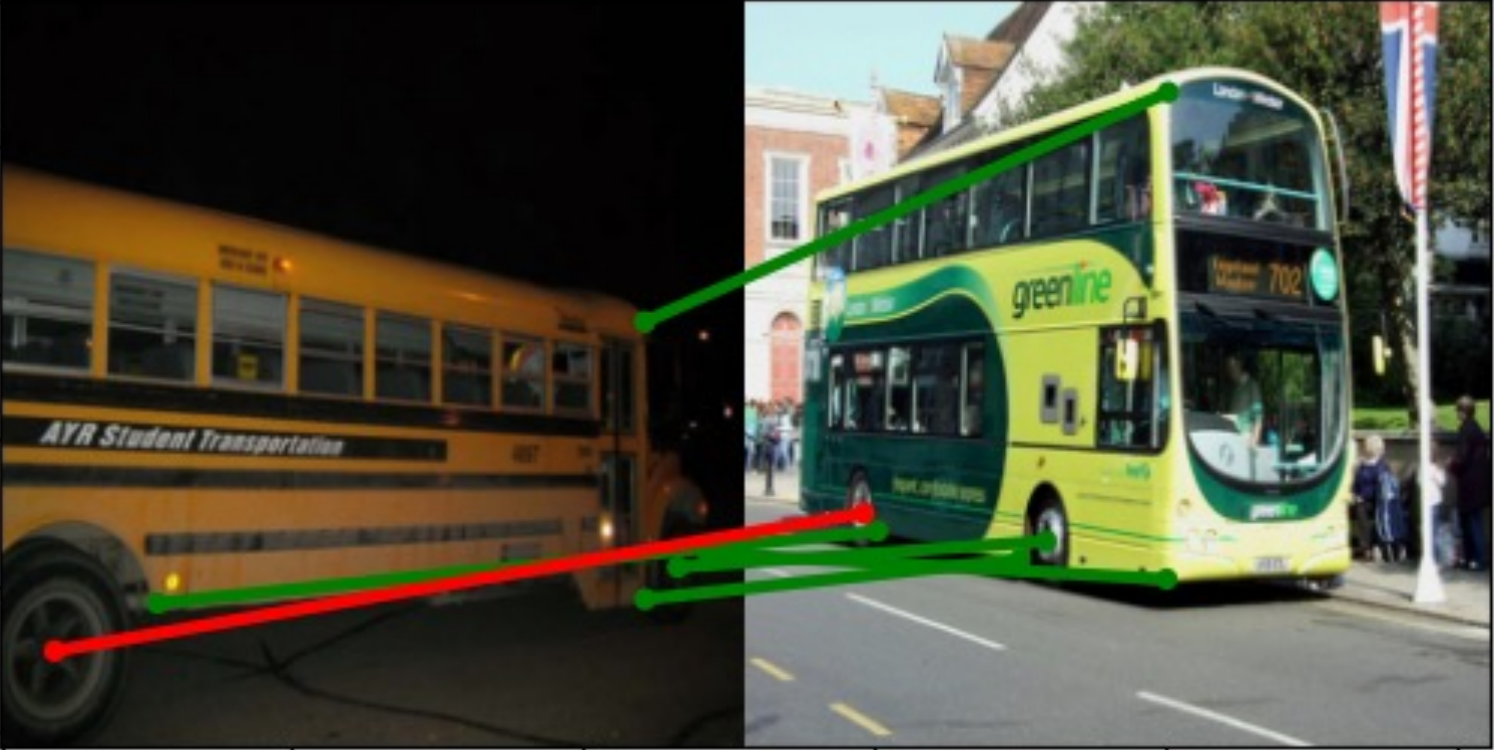}
  \end{subfigure}
  \begin{subfigure}[b]{0.19\textwidth}
    \centering
    \includegraphics[width=\textwidth]{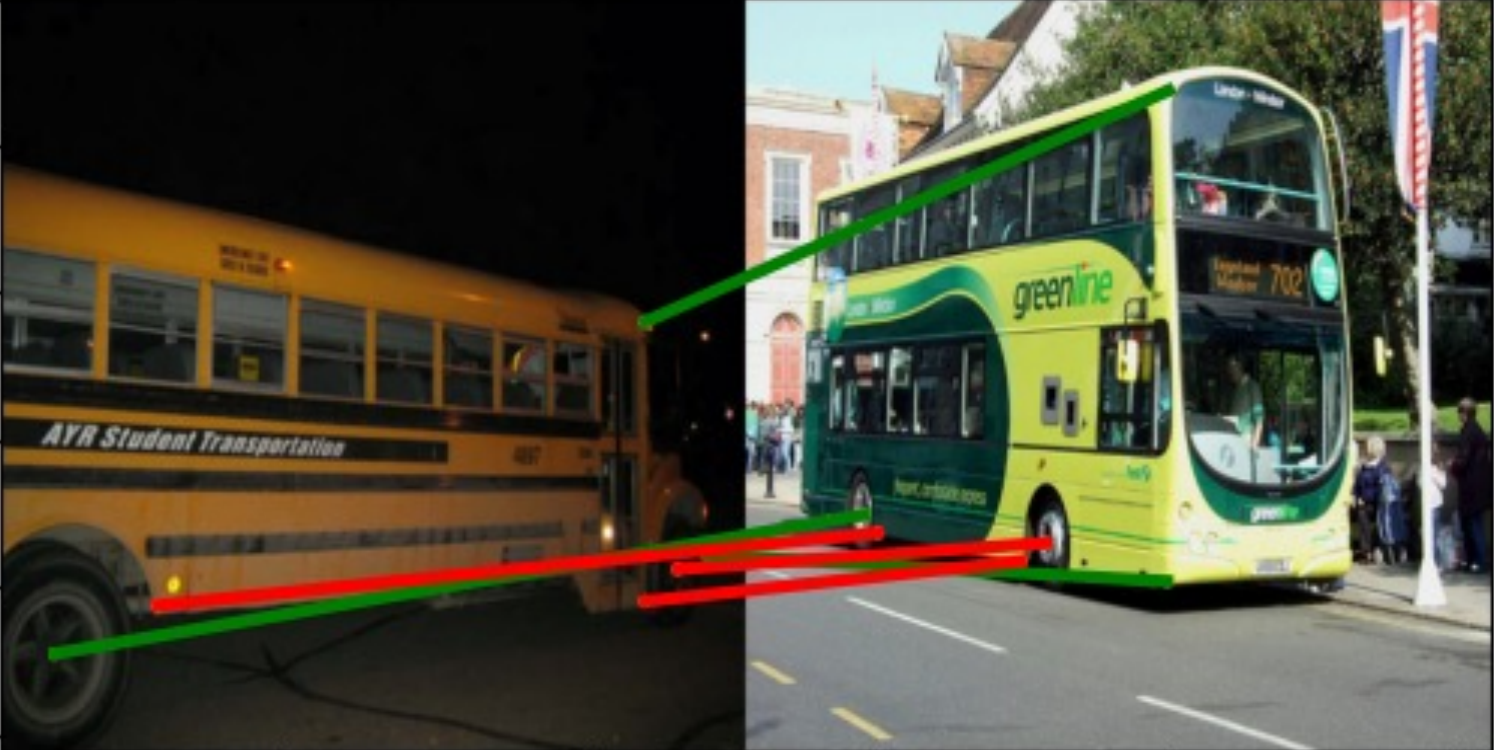}
  \end{subfigure}
    \begin{subfigure}[b]{0.19\textwidth}
    \centering
    \includegraphics[width=\textwidth]{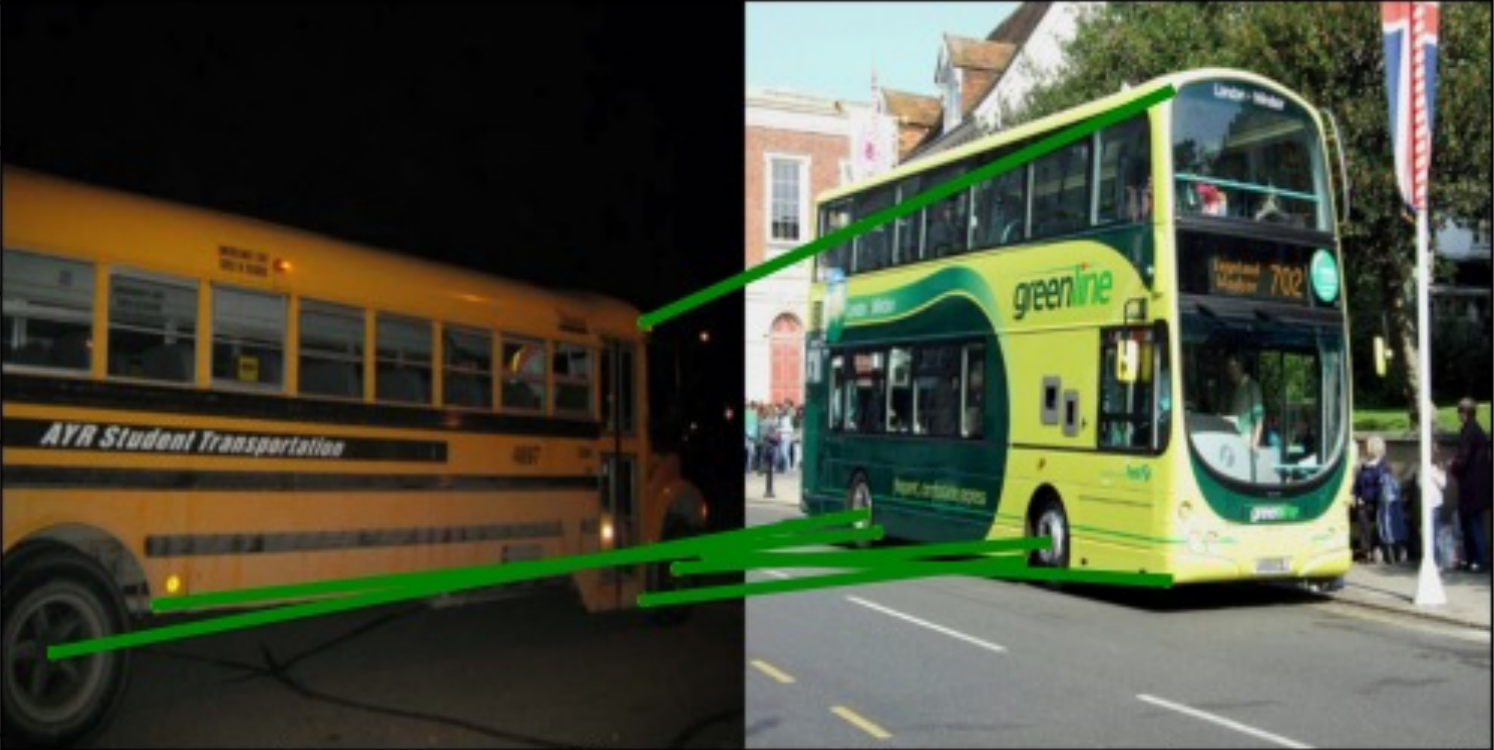}
  \end{subfigure} \\
     \begin{subfigure}[b]{0.19\textwidth}
    \centering
    \includegraphics[width=\textwidth]{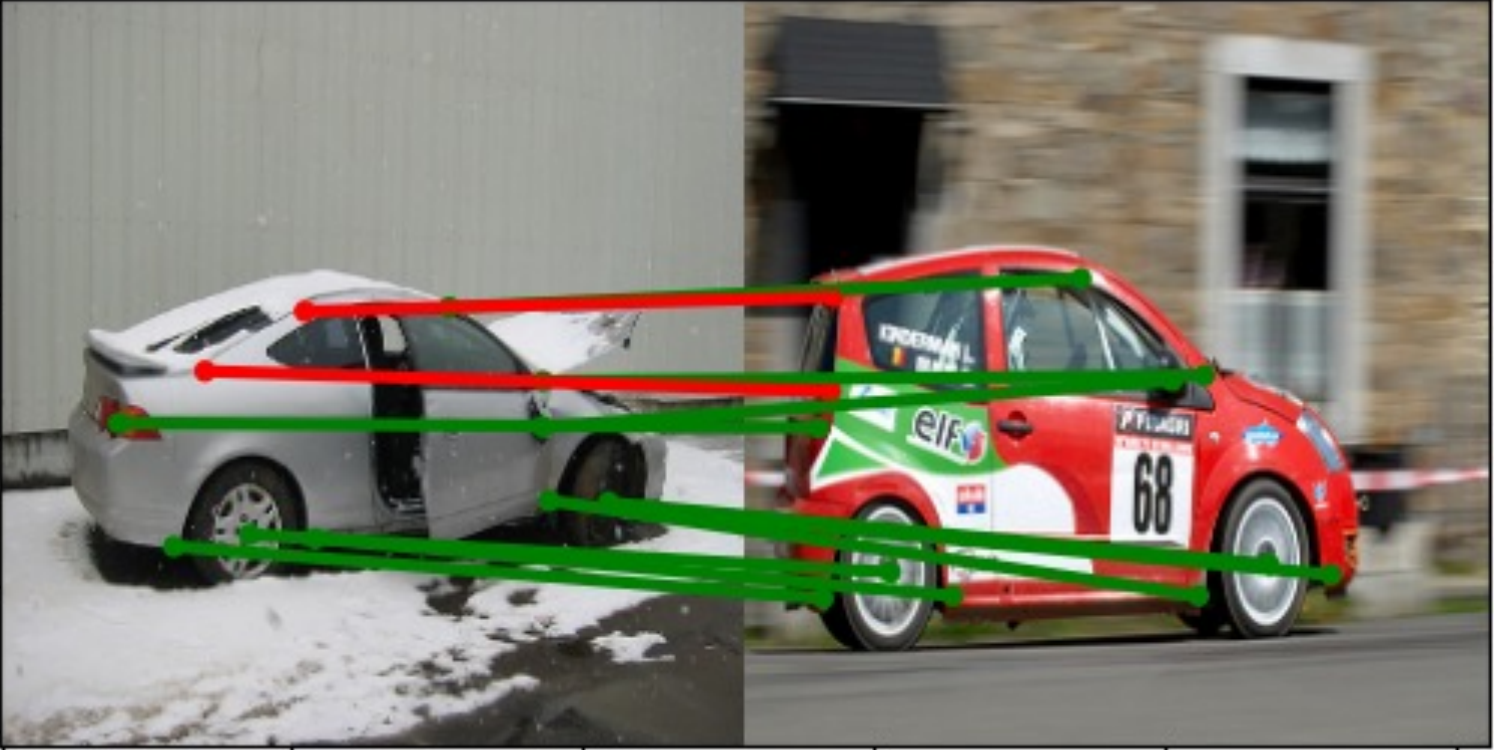}
  \end{subfigure}
  \begin{subfigure}[b]{0.19\textwidth}
    \centering
    \includegraphics[width=\textwidth]{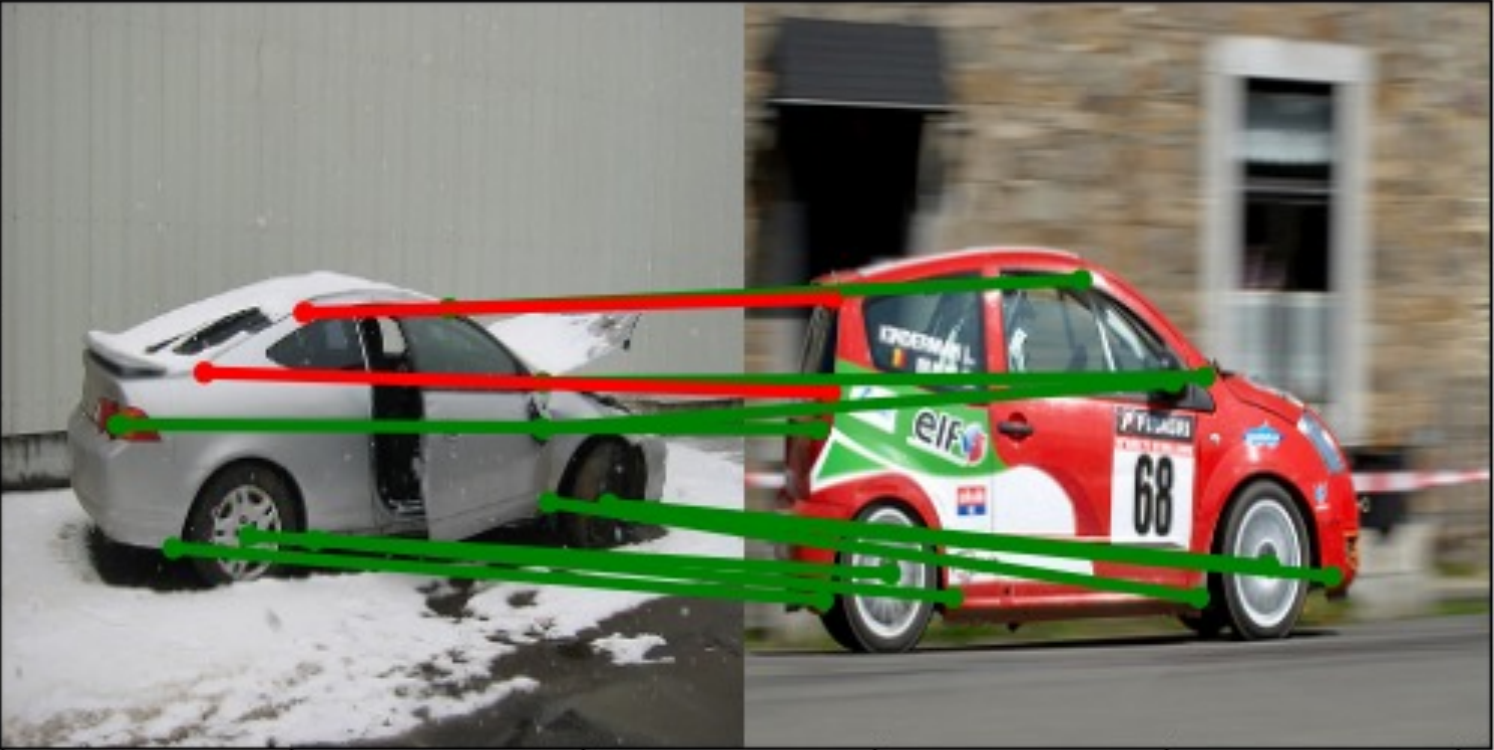}
  \end{subfigure}
  \begin{subfigure}[b]{0.19\textwidth}
    \centering
    \includegraphics[width=\textwidth]{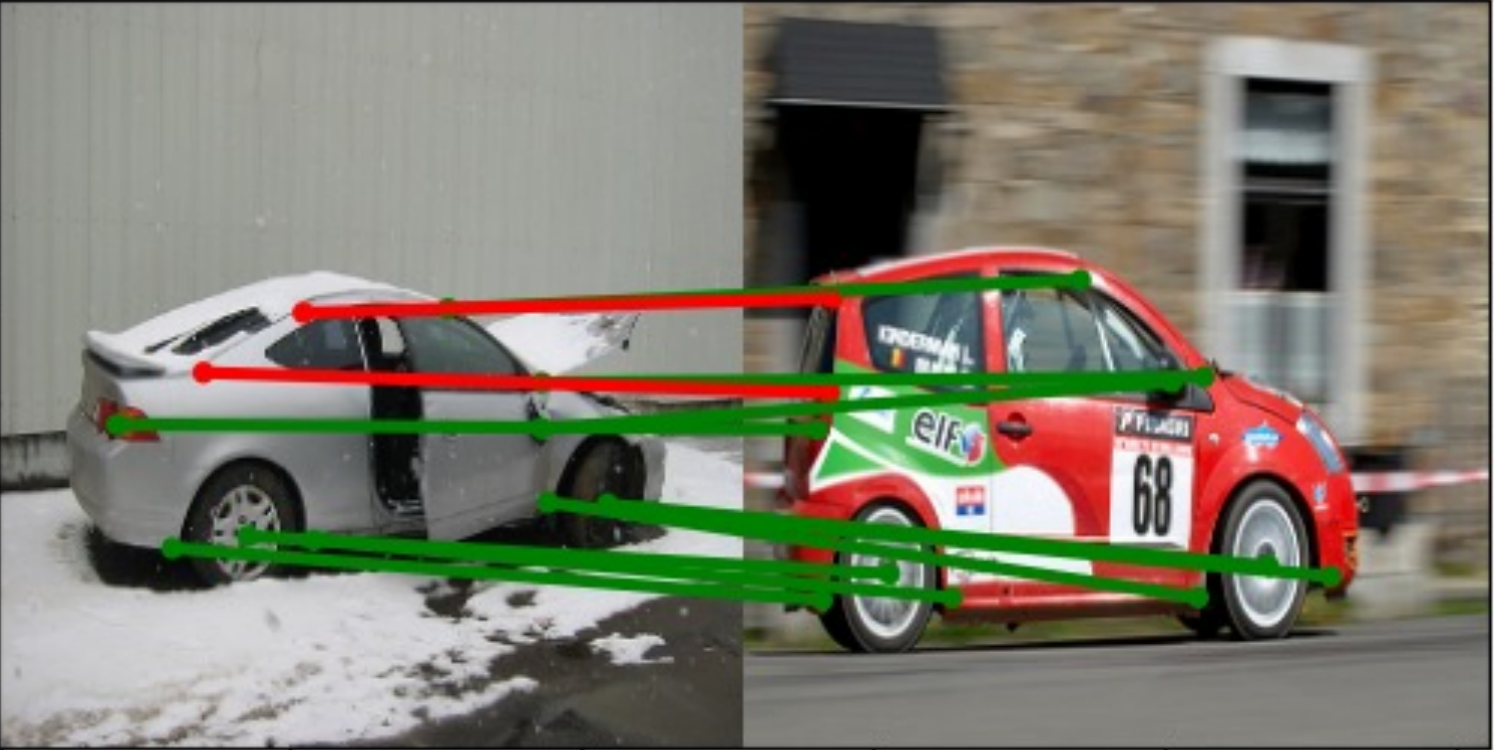}
  \end{subfigure}
  \begin{subfigure}[b]{0.19\textwidth}
    \centering
    \includegraphics[width=\textwidth]{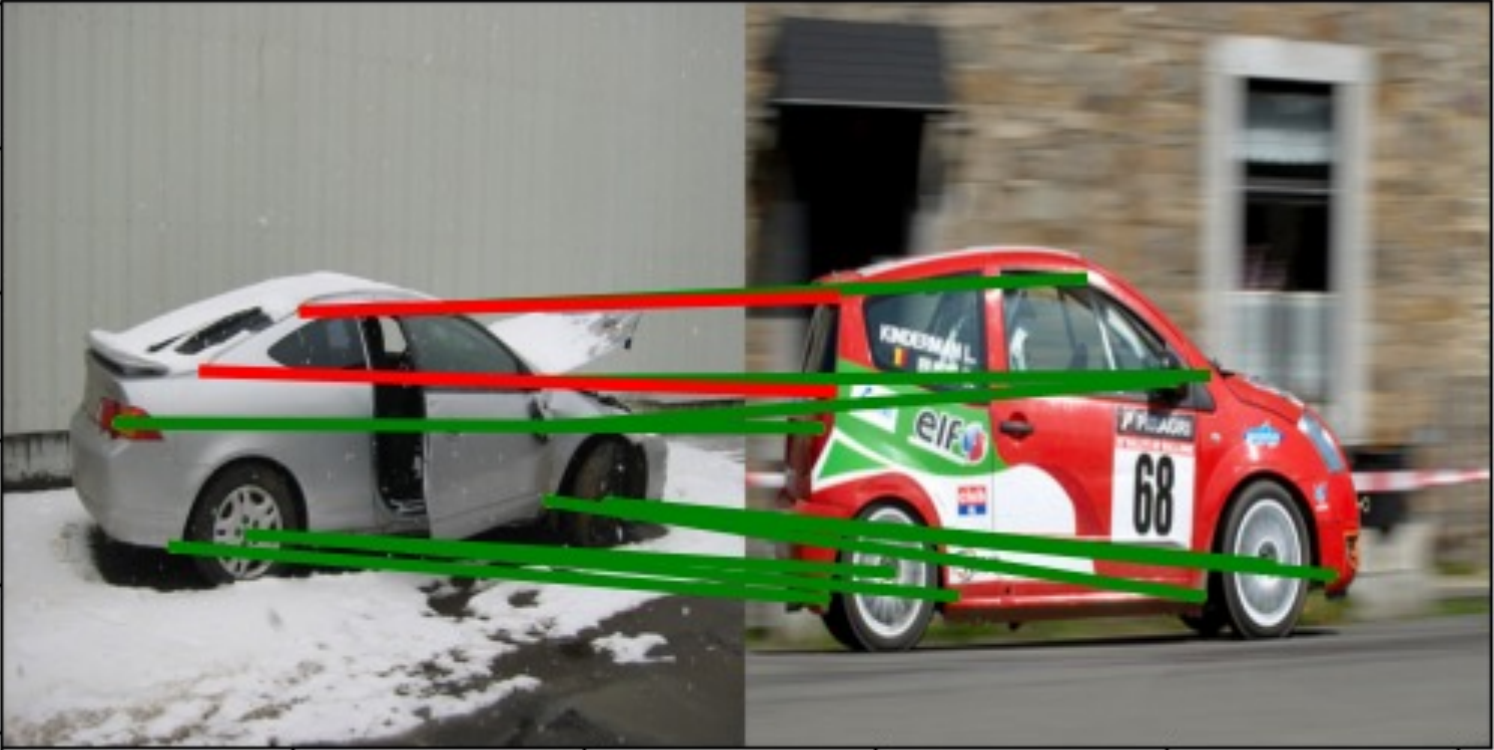}
  \end{subfigure}
    \begin{subfigure}[b]{0.19\textwidth}
    \centering
    \includegraphics[width=\textwidth]{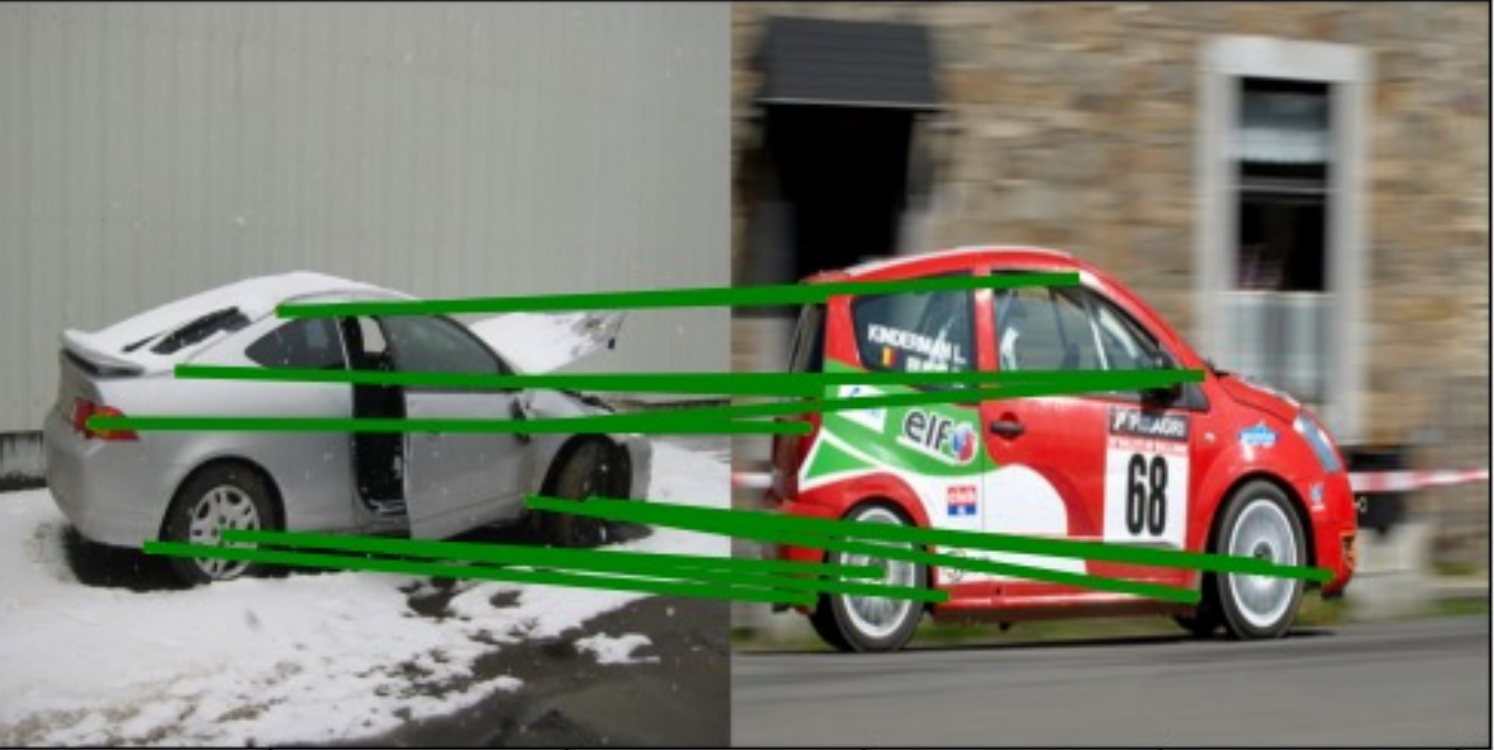}
  \end{subfigure} \\
      \begin{subfigure}[b]{0.19\textwidth}
    \centering
    \includegraphics[width=\textwidth]{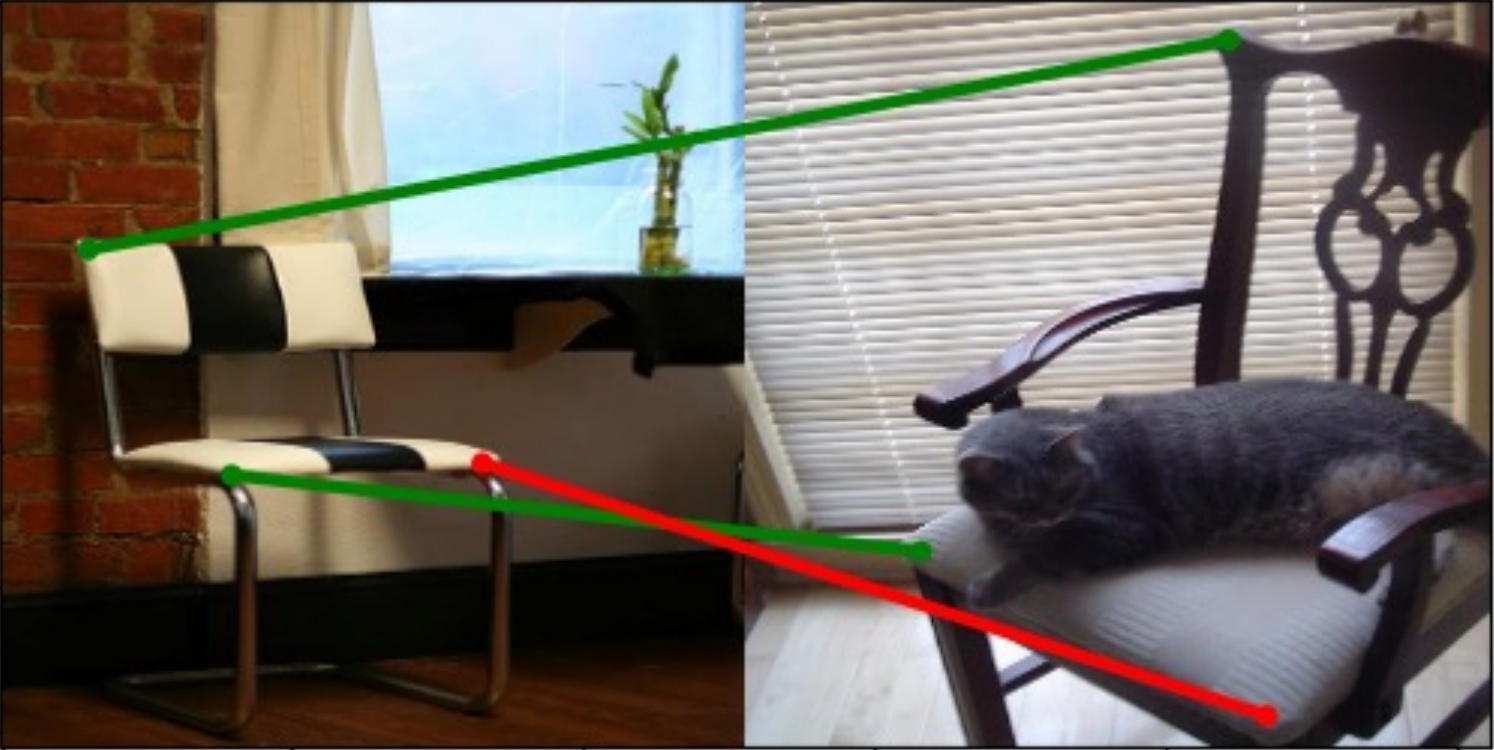}
  \end{subfigure}
  \begin{subfigure}[b]{0.19\textwidth}
    \centering
    \includegraphics[width=\textwidth]{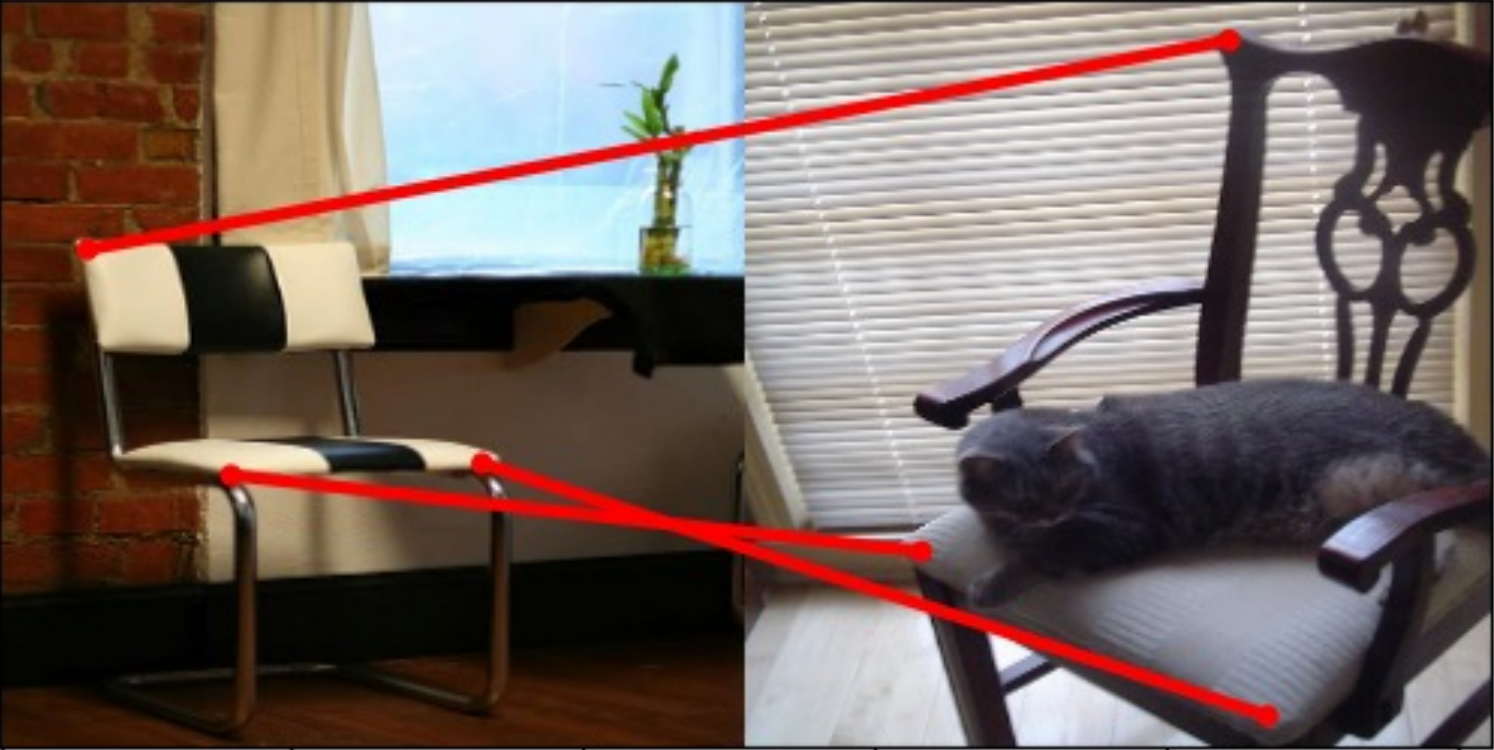}
  \end{subfigure}
  \begin{subfigure}[b]{0.19\textwidth}
    \centering
    \includegraphics[width=\textwidth]{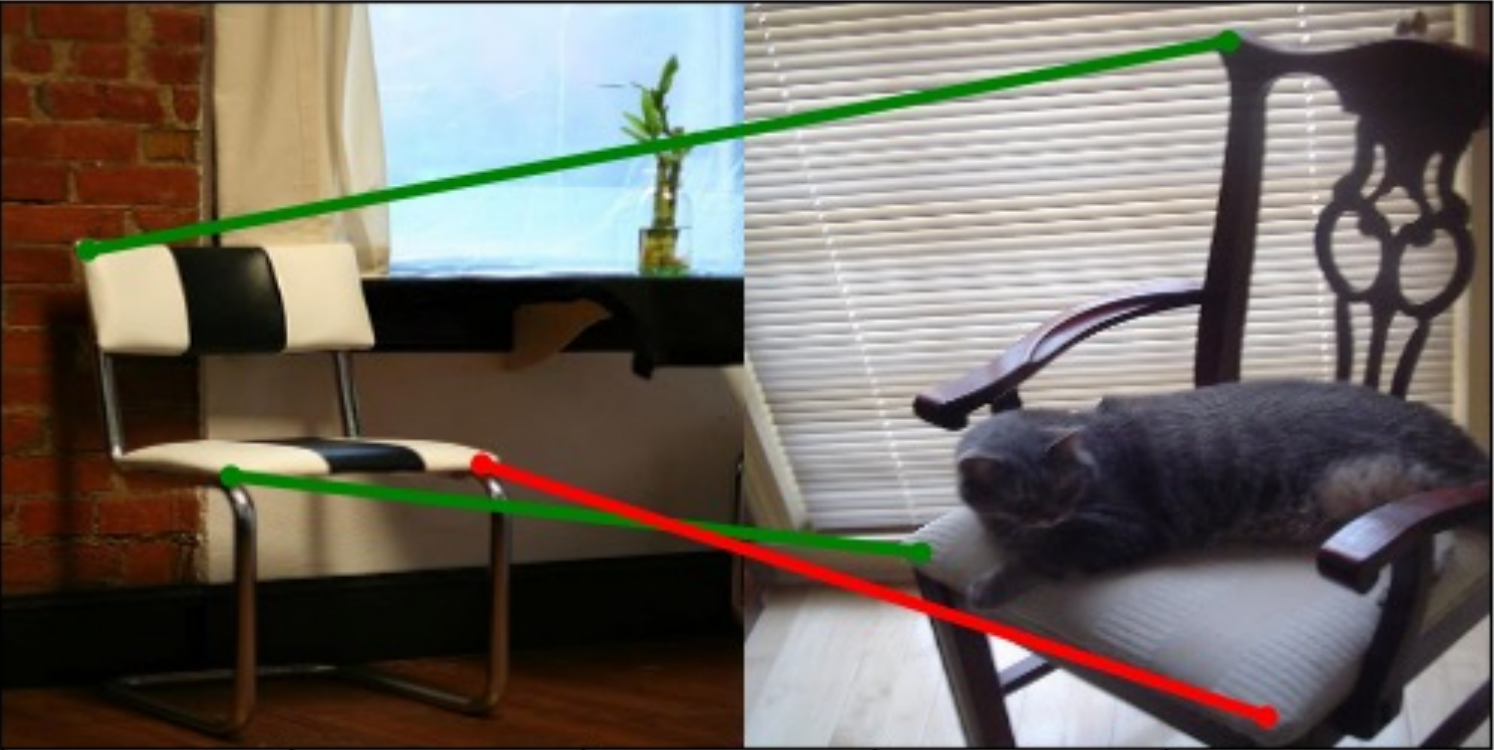}
  \end{subfigure}
  \begin{subfigure}[b]{0.19\textwidth}
    \centering
    \includegraphics[width=\textwidth]{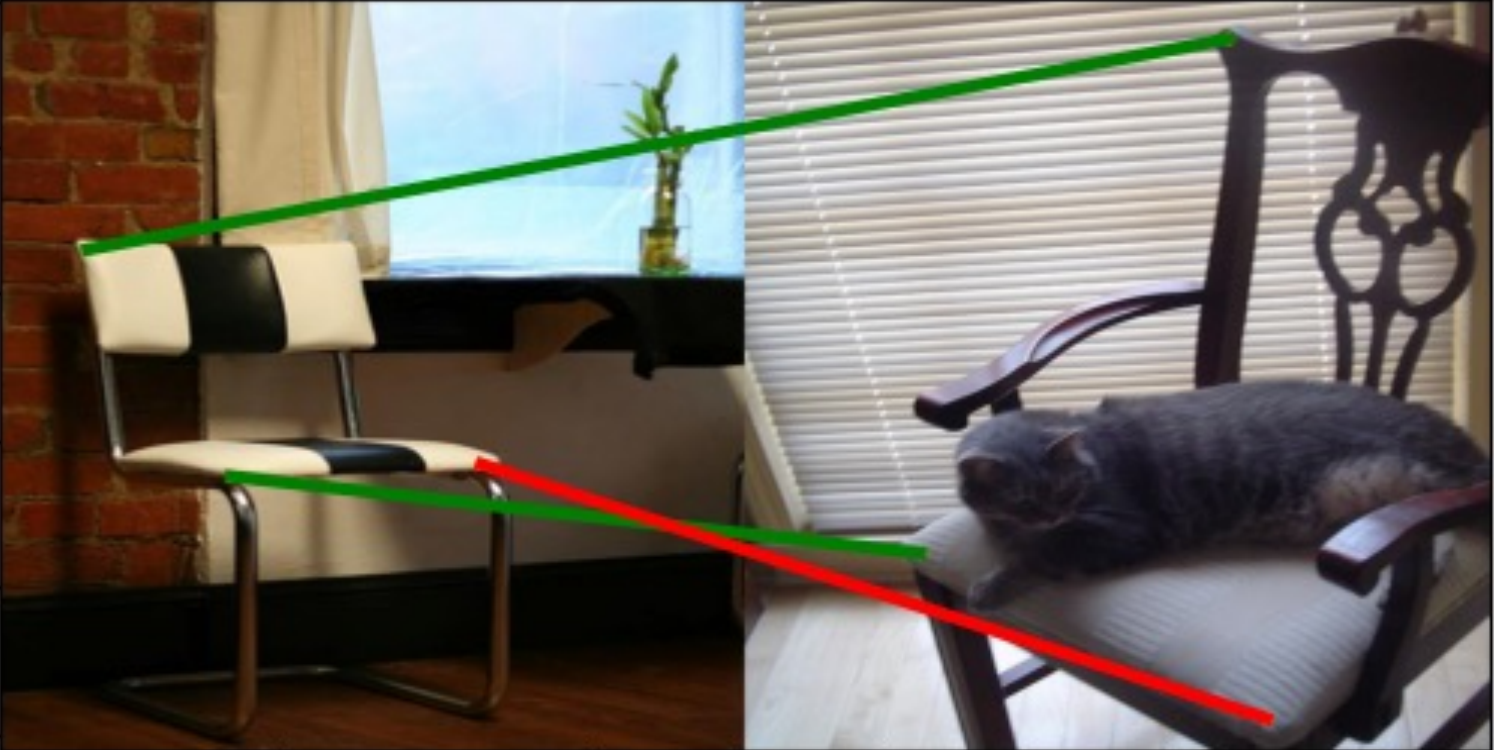}
  \end{subfigure}
    \begin{subfigure}[b]{0.19\textwidth}
    \centering
    \includegraphics[width=\textwidth]{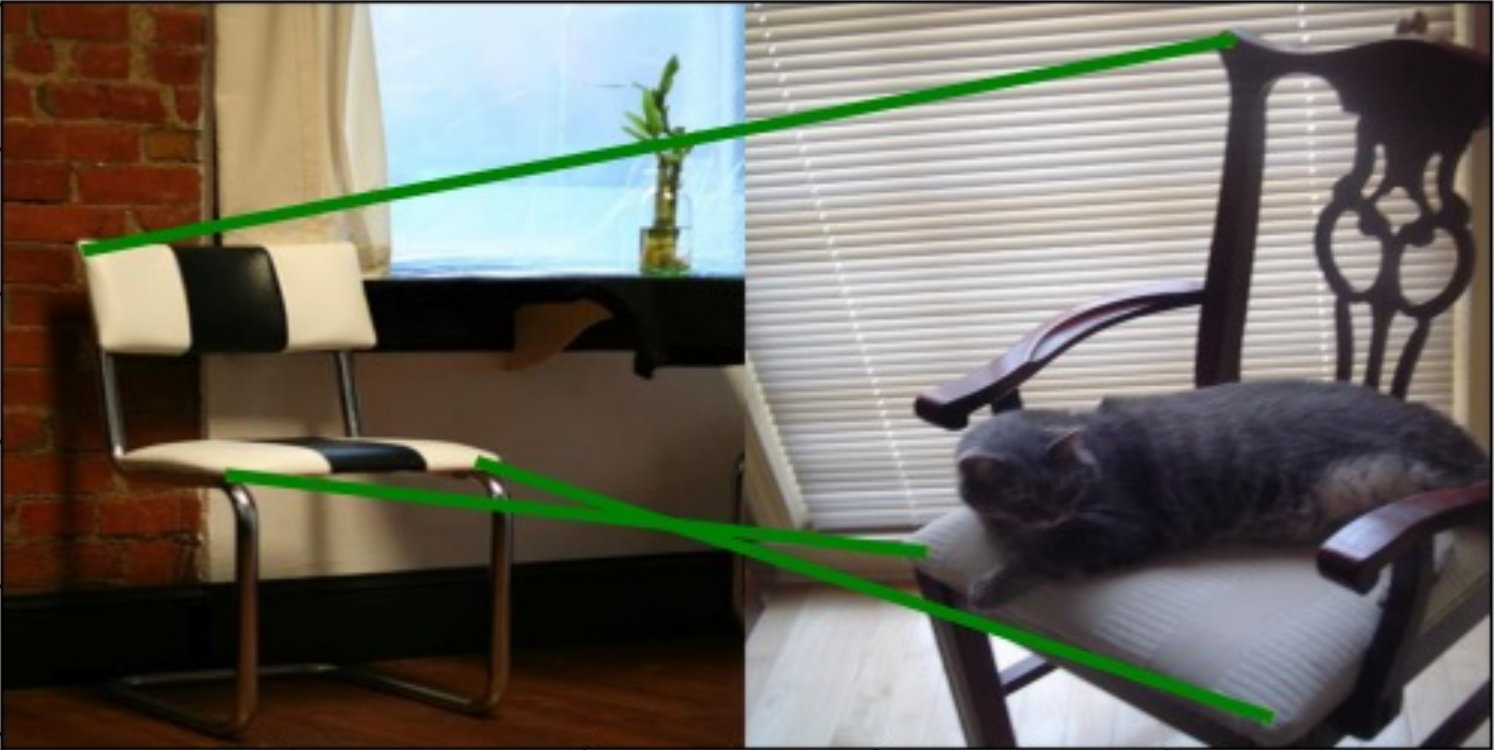}
  \end{subfigure} 
      \begin{subfigure}[b]{0.19\textwidth}
    \centering
    \includegraphics[width=\textwidth]{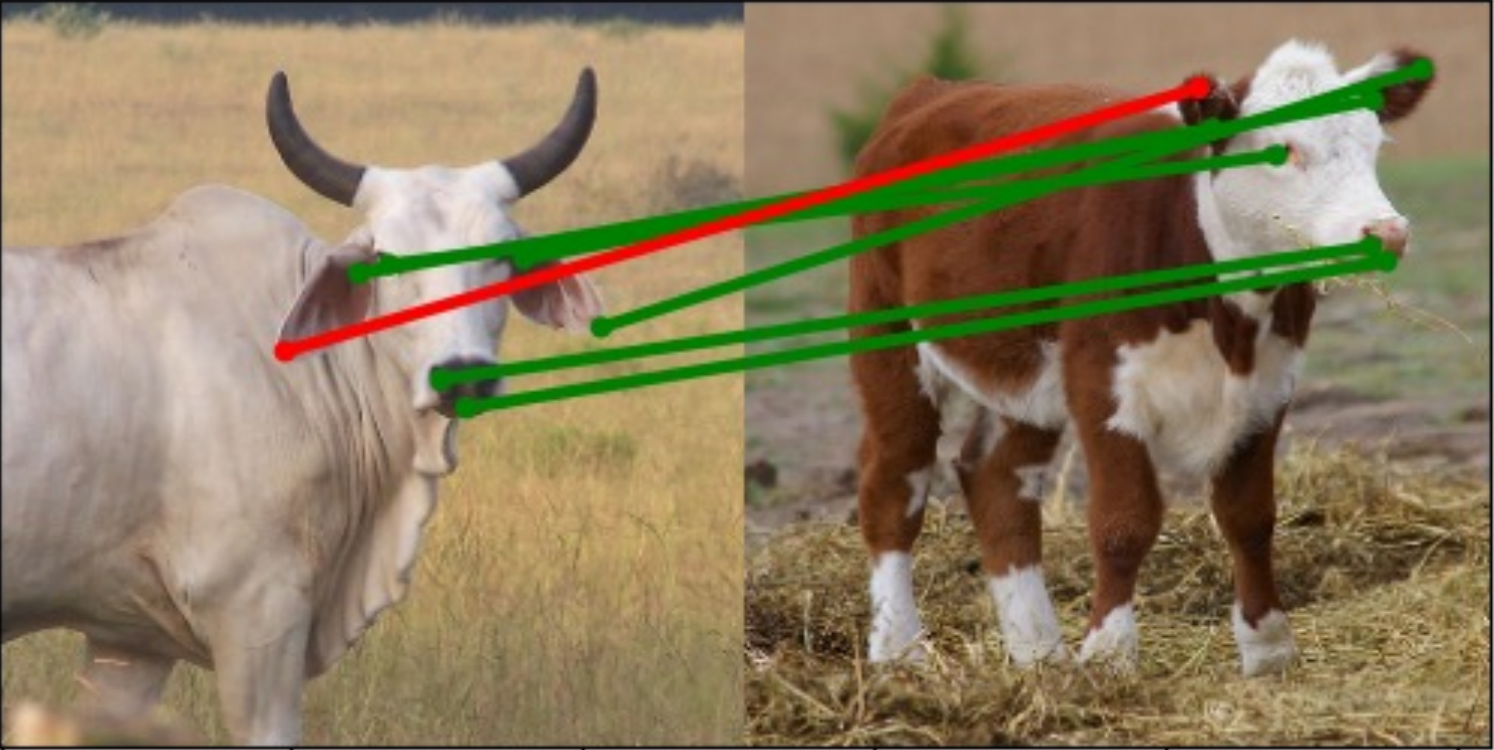}
  \end{subfigure}
  \begin{subfigure}[b]{0.19\textwidth}
    \centering
    \includegraphics[width=\textwidth]{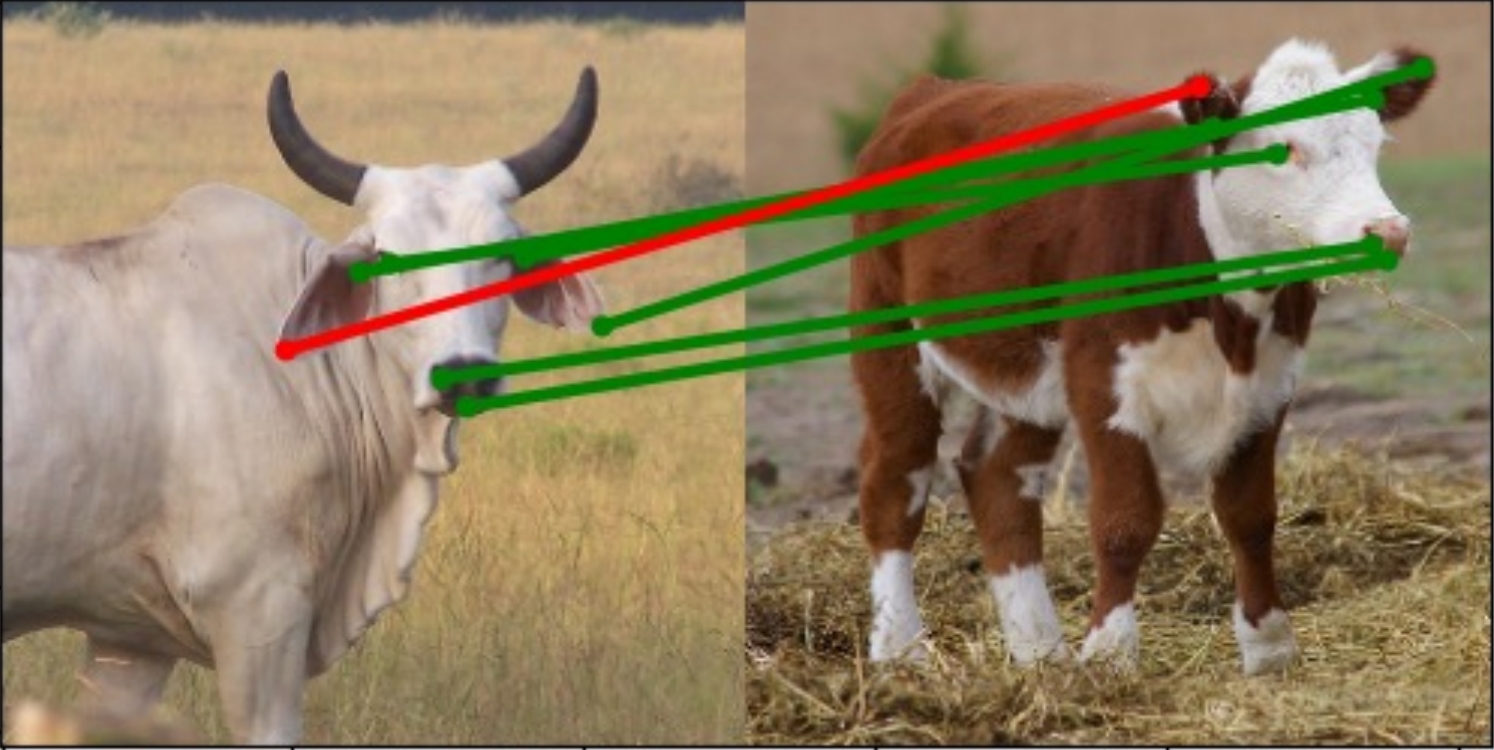}
  \end{subfigure}
  \begin{subfigure}[b]{0.19\textwidth}
    \centering
    \includegraphics[width=\textwidth]{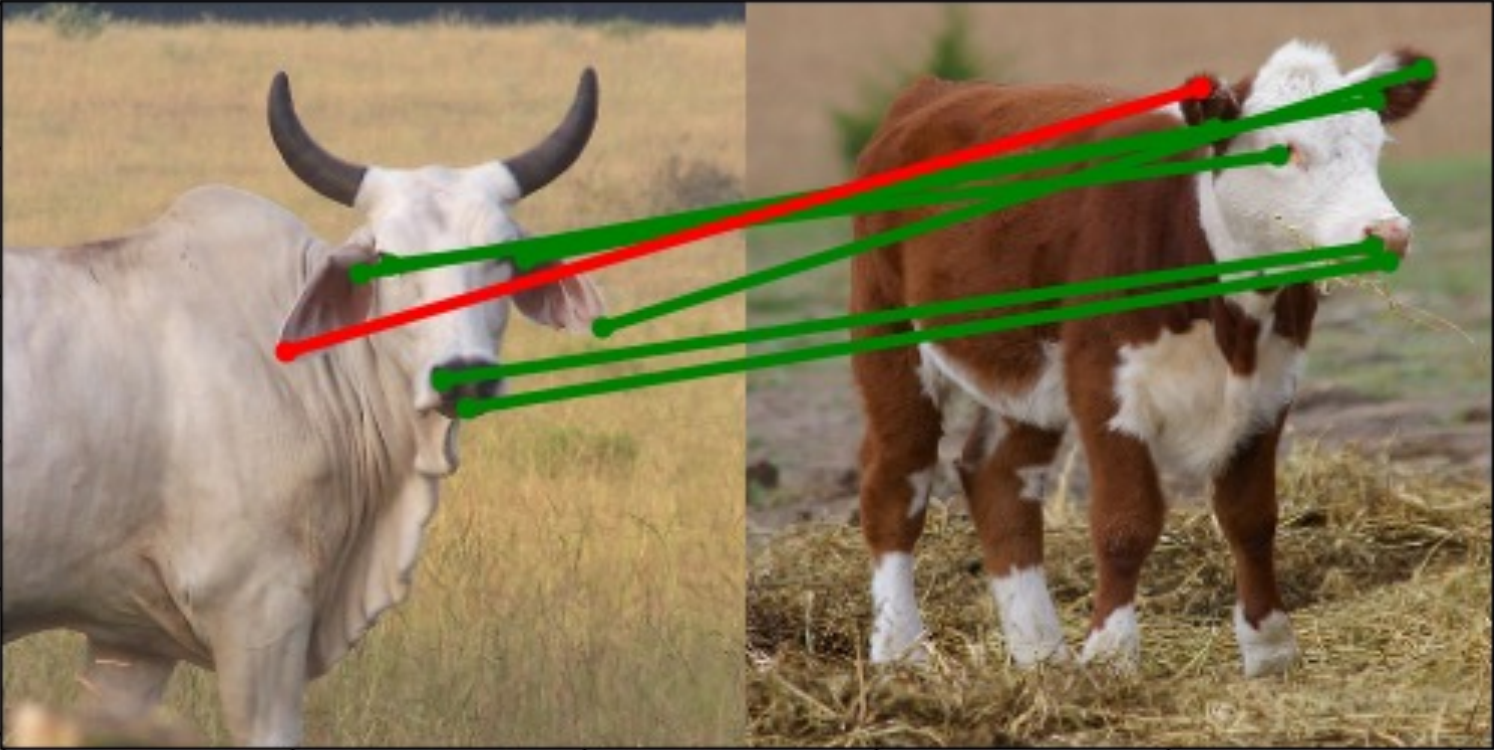}
  \end{subfigure}
  \begin{subfigure}[b]{0.19\textwidth}
    \centering
    \includegraphics[width=\textwidth]{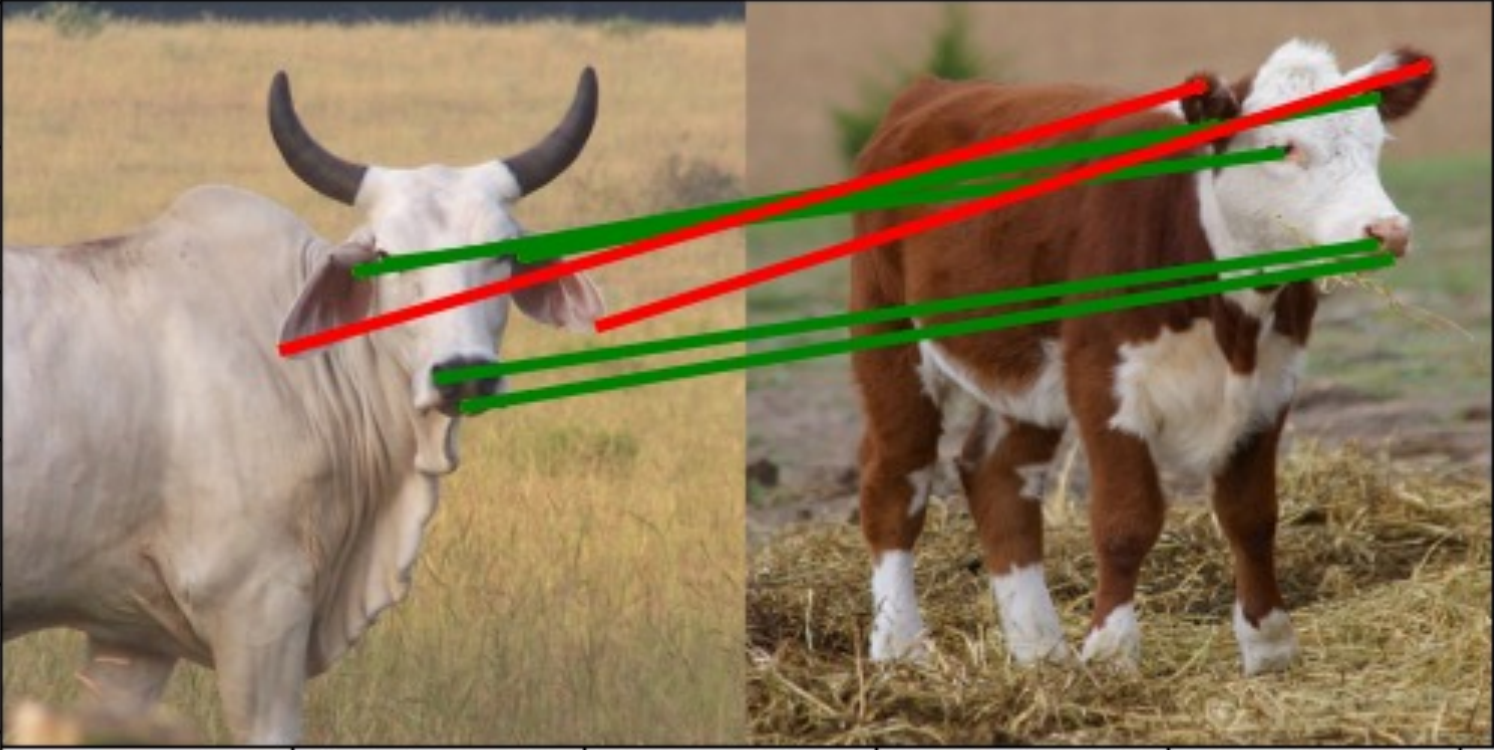}
  \end{subfigure}
    \begin{subfigure}[b]{0.19\textwidth}
    \centering
    \includegraphics[width=\textwidth]{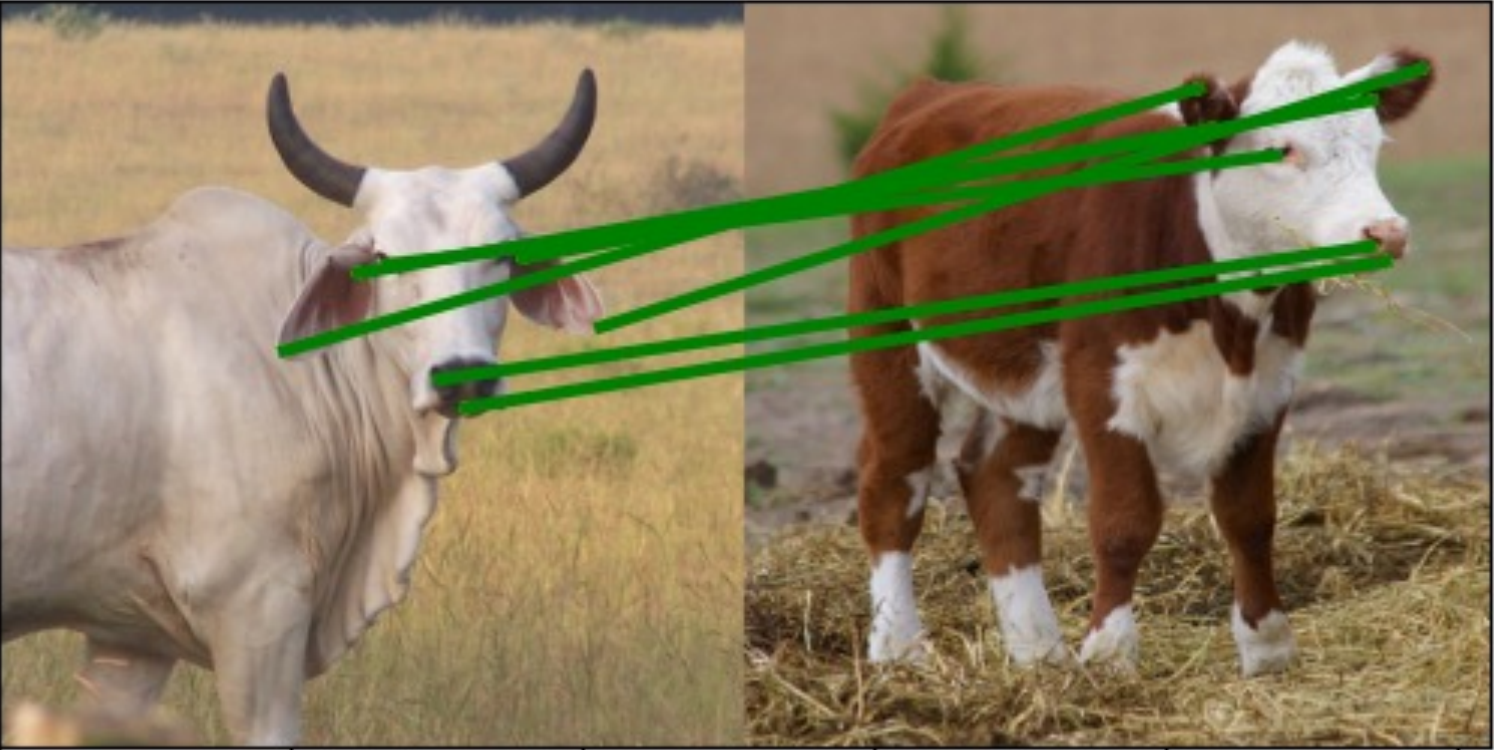}
  \end{subfigure} \\
  
  \begin{subfigure}[b]{0.19\textwidth}
    \centering
    \includegraphics[width=\textwidth]{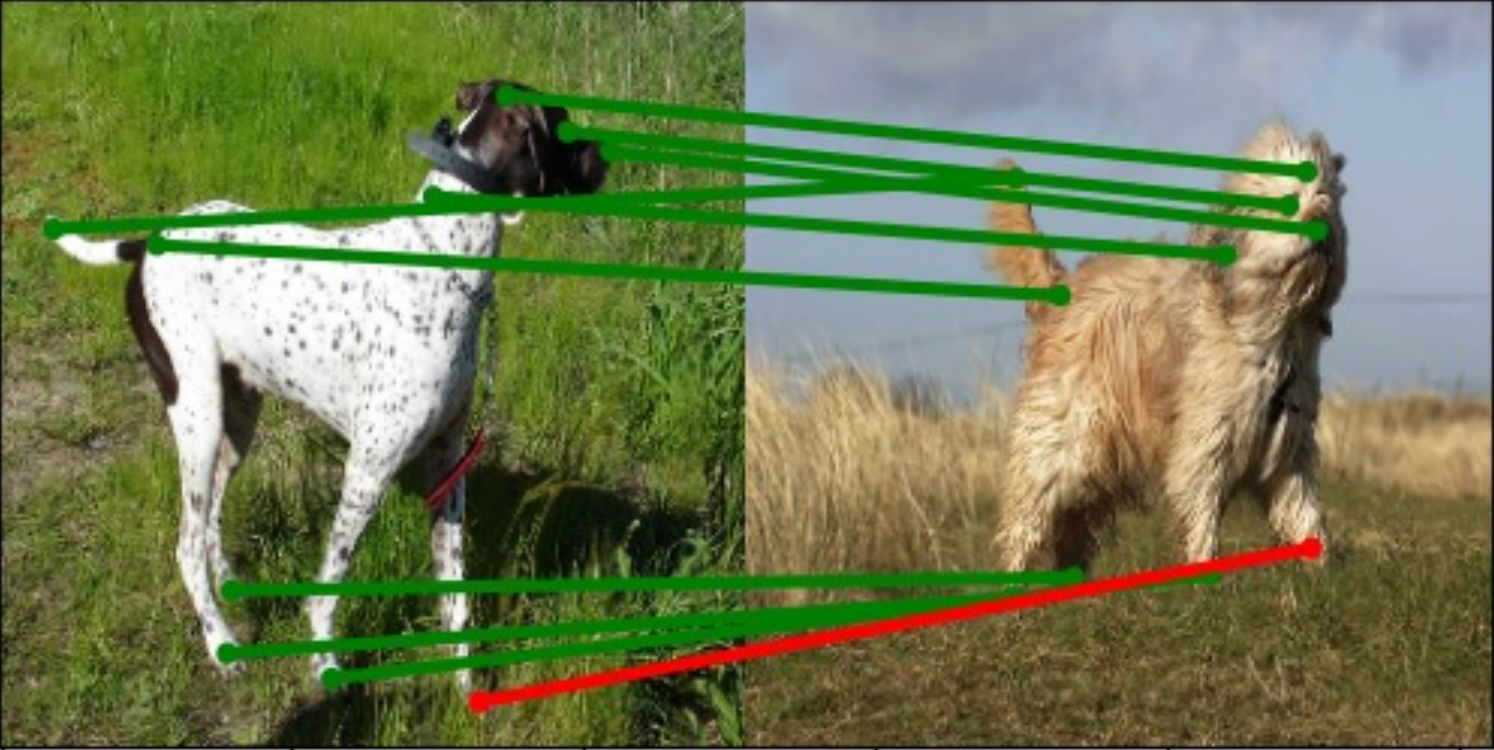}
    \caption{CATs}
  \end{subfigure}
  \begin{subfigure}[b]{0.19\textwidth}
    \centering
    \includegraphics[width=\textwidth]{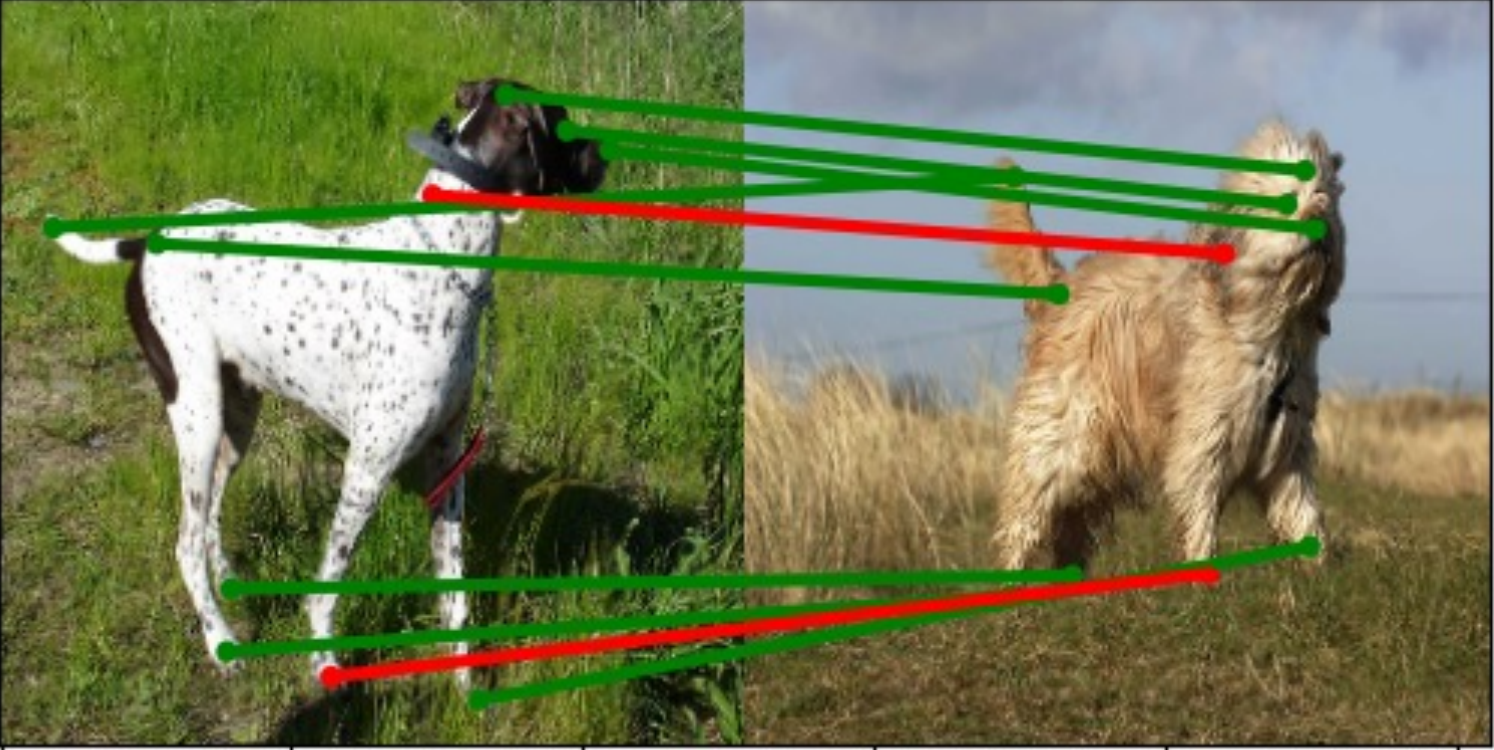}
   \caption{SemiMatch}
  \end{subfigure}
  \begin{subfigure}[b]{0.19\textwidth}
    \centering
    \includegraphics[width=\textwidth]{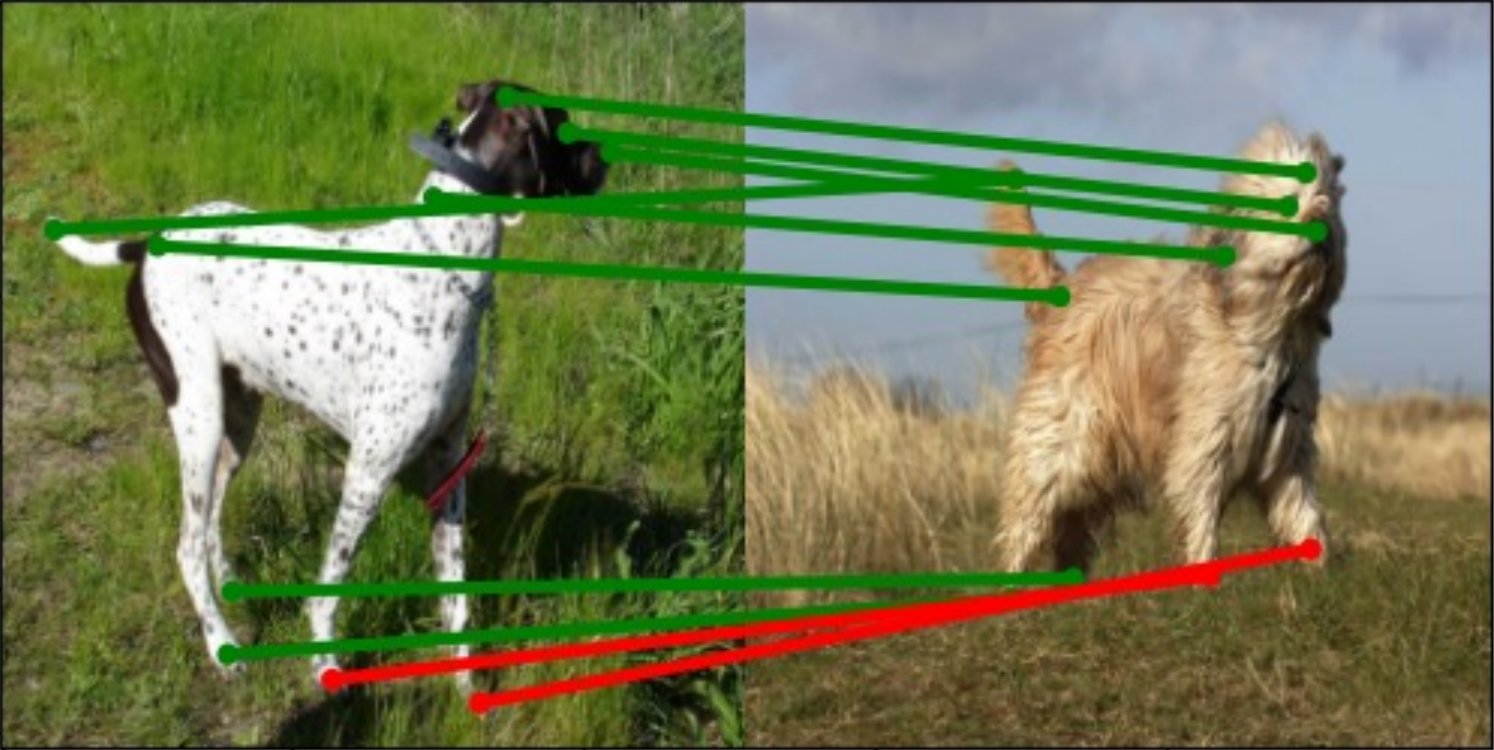}
   \caption{SCORRSAN}
  \end{subfigure}
  \begin{subfigure}[b]{0.19\textwidth}
    \centering
    \includegraphics[width=\textwidth]{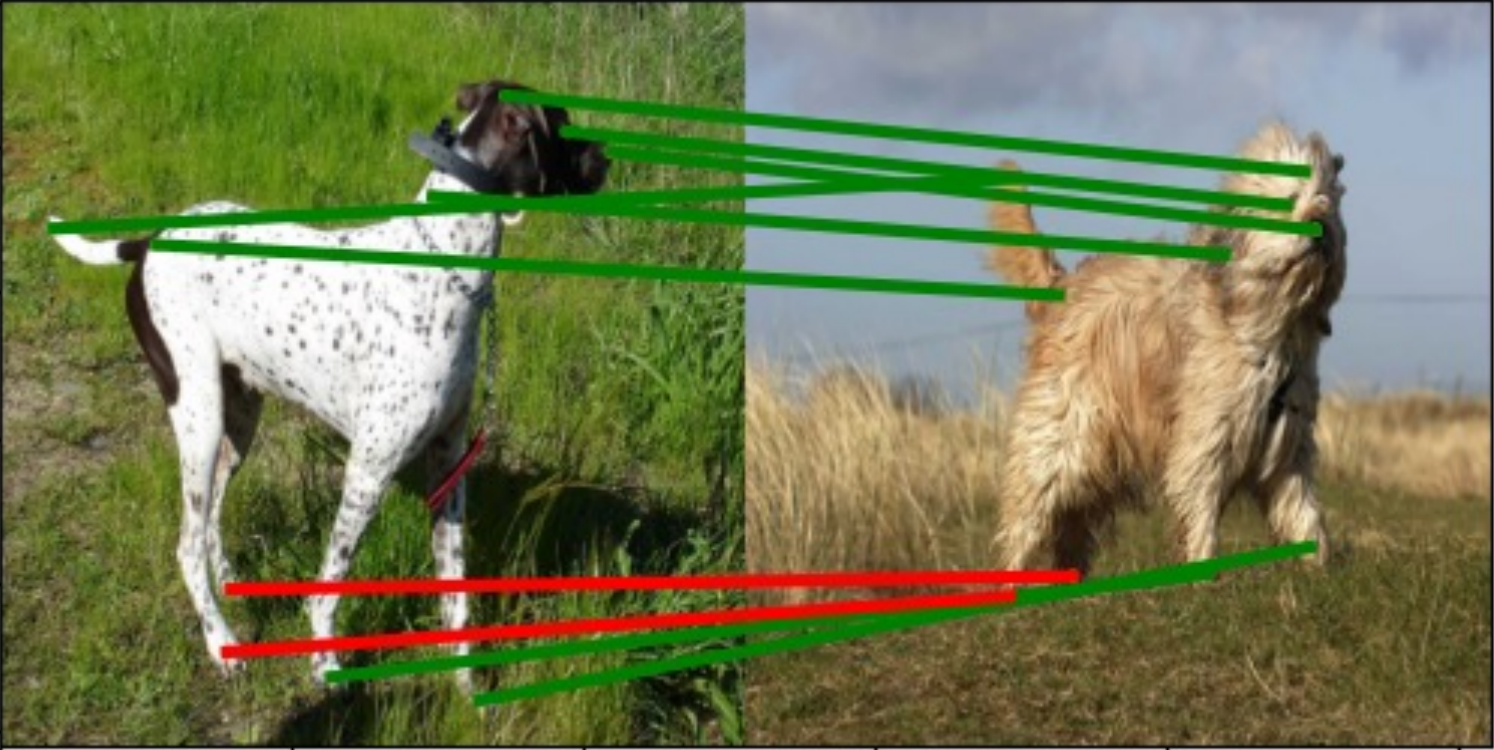}
   \caption{CATs++~}
  \end{subfigure}
    \begin{subfigure}[b]{0.19\textwidth}
    \centering
    \includegraphics[width=\textwidth]{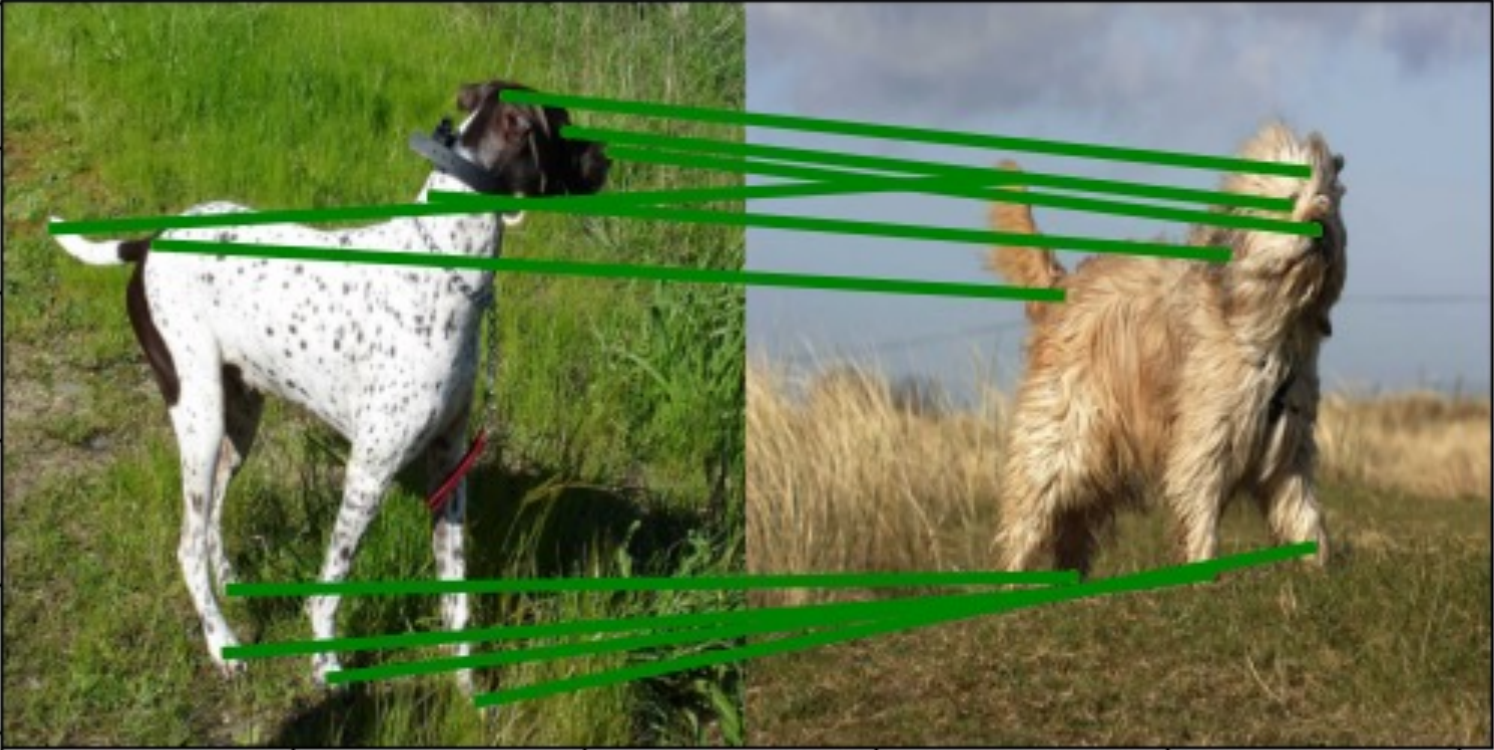}
   \caption{\ours}
  \end{subfigure} 
  \caption{\textbf{Qualitative results on SPair-71k in comparison with the competing SOTA methods.} The point-to-point matches are drawn by linking key point pairs with line segments. {\color{PineGreen} Green} and {\color{Red} red} lines denote correct and incorrect predictions with respect to the ground-truth pairs, respectively. We observe that ours outperforms the counterparts significantly across all the sample image pairs.}
  \vspace{-.5em}
  \label{fig:qual_sota}
  
\end{figure*}

\noindent\textbf{Implementation details.}
We demonstrate our proposed method with two simple baselines, CATs~\cite{cho2021semantic} and CATs++~\cite{cho2022cats++}. 
We use the best-performing model's weight trained in a supervised setting on labeled data for the initial annotator to generate labels. Only weak photometric augmentations, such as color-jitter and gray-scale, are used with a given probability of 0.2 to prevent early over-fitting. We employ stronger data augmentations to benefit generated labels more. Following the literature~\cite{cho2021semantic,cho2022cats++}, a combination of strong photometric augmentation at a frequency of 0.4 is used along with geometric augmentation~\cite{rocco2017convolutional,lee2019sfnet,truong2020glu}. 
The confidence threshold $\tau$ for the generated correspondences is commonly set as 0.7 for all image pairs in training datasets.
We pick the best trained model as a new machine annotator for successive iterative training. %

\subsection{Comparison on Benchmarks}
We evaluate our method in comparison with the SOTA methods~\cite{zhao2021multi,min2021convolutional,cho2021semantic,kim2022transformatcher,cho2022cats++,kim2024efficient} trained based on the supervised protocol with existing keypoint annotations. We also compare with similar methods~\cite{kim2022semimatch,huang2022learning}, using both generated labels and GT labels and a method without supervision~\cite{tang2023emergent}

\noindent\textbf{On SPair-71k.}
Tab.~\ref{tab:spair} shows PCKs ($\alpha_{\text{bbox}}{=}0.1$) on all 18 object classes, including the overall mean PCK. Our overall averaged PCK=\textbf{62.0}\% significantly outperforms the current state-of-the-art methods. We achieve {+2.2}\% of PCK improvements over the baseline~\cite{cho2022cats++}. It demonstrates that the matching networks, especially with correlation enhancement architecture, have been under-trained with sparse and limited keypoint supervision. Furthermore, our consistently superior performance in both per-class and average PCK compared to state-of-the-art methods across various regimes~\cite{cho2022cats++,kim2024efficient,kim2022semimatch,huang2022learning,tang2023emergent} indicate improved generalizability, which facilitates handling large intra-class variation and deformation between instances within the same object class.

Additionally, as shown in Fig.~\ref{fig:qual_sota}, we visualize the sampled example pairs with the predicted matches for \ours \ and the competing methods showing the best performance in both the supervised regime, such as CATs~\cite{cho2021semantic}, CATs++~\cite{cho2022cats++} and the regime similar to ours, such as SemiMatch~\cite{kim2022semimatch}, SCORRSAN~\cite{huang2022learning}, using both supervision from generated labels and GT keypoint labels. This demonstrates that our method estimates correspondences between image pairs more accurately than others, even under a large discrepancy in viewpoint and scale.

\begin{table}[t]
\caption{\textbf{Comparison with state-of-the-art methods on PF-PASCAL and PF-WILLOW.}
Numbers in bold denote the best, and underlined ones are the second best. \ours outperforms the competing methods again, like in Spair-71k.}
\label{tab:pf}
    \centering
    \tabcolsep=.5em
    \begin{tabular}{l|ccc|ccc}
    \toprule 
    \multirow{3}{*}{Methods} & 
    \multicolumn{3}{c|}{\textbf{PF-PASCAL}} & 
    \multicolumn{3}{c}{\textbf{PF-WILLOW}} 
     \\
     &
    \multicolumn{3}{c|}{$\alpha_\text{img}$} &
    \multicolumn{3}{c}{$\alpha_\text{bbox}$} 
    \\
    
    &  
    $0.05$ &  $0.1$ &  $0.15$ &  
    $0.05$ &  $0.1$ &  $0.15$ 
    \\ \midrule
    HPF%
    &
    60.1 & 84.8 & 92.7 & 
    45.9 &74.4 &85.6
    \\
    DHPF%
    &
    75.7 & 90.7 & 95.0 &
    49.5 & 77.6 & 89.1 
    \\
    MMNet%
    &
    77.6 & 89.1 & 94.3 &
    - & - & - 
    \\
    CHM%
    &
    80.1 & 91.6 & - &
     52.7 & {79.4} & -
    \\
    
     CATs%
     &
    75.4 &92.6& 96.4 &
    50.3& 79.2& {90.3}
    \\
TransforMatcher%
    & 
    - &{80.8}&91.8 &
    -&{76.0}&{-}
    \\
    CATs++%
    &
    \textbf{84.9} &\underline{93.8} & \textbf{96.8}&
    56.7 & 81.2 &-
    \\
    PMNC%
    &
    \underline{82.4} & 90.6 & - &
    - &-& - \\
    SemiMatch%
    &
    80.1 &93.5 &96.6 &
    54.0 &\underline{82.1}& \underline{92.1} 
    \\
    SCORRSAN%
    & 
    81.4 &{92.9}&96.1 &
    54.1&{80.0}&{89.8}
    \\
    DIFT$_{sd}$ 
    & 
    - &{-}&- &
    \underline{58.1}&{81.2}&{-}
    \\HCCNet%
    &
    {-} & 92.4 & - &
    - &74.5& -
    \\
    \midrule
    
    \rowcolor{gray!10} 
    \ours (ours) &
    \textbf{84.9} & \textbf{94.3} &  \underline{96.7} &
    \textbf{59.6}& \textbf{83.6 }& \textbf{92.9}
    \\
    \bottomrule
    \end{tabular}
\vspace{-.5em}
\end{table}
\begin{table}[t]
\centering
\caption{\textbf{PCK comparison among training methods.} For a fair comparison, we use the fixed baseline CATs++~\protect\cite{cho2022cats++} for all semi-supervised training methods. While all lead to performance improvements, ours enjoy the most significant enhancement, which highlights the need to prioritize the data-hungry matter.
}
\label{tab:baseline_comparison}
\tabcolsep=1.5em
    \begin{tabular}{r|l|l}
    \toprule 
    & Methods & PCK \\
    \midrule
(a) & Baseline & 59.8 \\
\midrule
(b) & (a) + CNNGeoU~\cite{laskar2018semi} & 60.1 {\scriptsize (+0.3)}\\
(c) & (a) + PWarpC~\cite{truong2022probabilistic} & 60.5 {\scriptsize (+0.7)}\\
(d) & (a) + SCORRSAN~\cite{huang2022learning} &  61.0 {\scriptsize (+1.2)}\\
 \rowcolor{gray!10} 
(e) & (a) + \ours (ours) & \textbf{62.0} {\scriptsize (+2.2)}\\
\bottomrule
\end{tabular}
\vspace{-.5em}
\end{table}

\noindent\textbf{On PF-PASCAL and PF-WILLOW.}
Tab.~\ref{tab:pf} summarizes our results on the PF-PASCAL and PF-WILLOW datasets compared with the other competing methods trained on PF-PASCAL from each initialized model (\ie, usually pre-trained on ImageNet~\cite{russakovsky2015imagenet}). We also fine-tune our model, pre-trained on SPair-71k with the unlabeled data from PASCAL VOC 2012, on the PF-PASCAL dataset to evaluate the generalization capability of our model on different datasets. \ours records the new state-of-the-art PCK value {94.3} that beats the previous state-of-the-art value of 93.8, which is almost saturated, on PF-PASCAL.

On PF-WILLOW, \ours \ outperform the baseline~\cite{cho2022cats++} by {2.9}\% / {2.4}\% ($\alpha{=}0.05/0.1$), surpassing other competing methods across different regimes. Note that our method not only achieves higher PCKs than the competing methods on PF-PASCAL but also outperforms them by a more significant margin on PF-WILLOW. This signifies the generalization capability of our method and discloses that ours learns a general representation, which can be applied to various datasets, different from the baselines usually overfitted on a specific dataset. 

\noindent\textbf{Controlled experiments for learning methods.}
We conduct controlled comparisons between our method and existing semi-supervised methods~\cite{laskar2018semi,truong2021warp,huang2022learning}. All the methods are trained with the fixed baseline CATs++~\cite{cho2021semantic} for a fair comparison to evaluate the methods' uniqueness in improving each method's performance without any potential influence from the model difference.
We use SPair-71k~\cite{min2019spair}, which contains fixed, sparsely-annotated pairs, for a comprehensive comparison. We strive to report the best results for each method via parameter searches. 

Tab.~\ref{tab:baseline_comparison} first shows \ours \ outperforms all the competitors. 
Specifically, (a) and (b), using a cycle consistency, and (c), using generated labels as sources of unsupervised loss signal, show limited performance improvement compared to the baseline supervised learner (a). This is attributed to their narrow focus on augmenting labels within a limited amount of labeled data. Unlike them, our method (d) focuses on densifying generated labels even with \textit{unlabeled} data as well as labeled data, thereby highly boosting performance. 

\begin{table*}[t]
\caption{\textbf{Robust evaluation on SPair-C}. We report the PCK numbers of our model and the baseline evaluated on clean (SPair-71k) and the newly introduced corrupted dataset (Spair-C), using the pre-trained weights of our model and the baseline model~\protect\cite{cho2021semantic} 
 provided by official code on SPair-71k.%
We report detailed PCKs for all the corruptions and further report the averaged PCKs to facilitate comparison. Numbers in bold indicate the best performance, and underlined ones are the second best.}
\label{tab:rob_PCK}
\centering
\resizebox{1\linewidth}{!}{
\tabcolsep=0.2em
\begin{tabular}{l|c|cccc|cc|cccc|ccccc|c|c}
\toprule

\multirow{2}{*}{Methods} &
\multirow{2}{*}{\textbf{Sev.}} &
\multicolumn{4}{c|}{\textbf{Noise}} &
\multicolumn{2}{c|}{\textbf{Blur}} &
\multicolumn{4}{c|}{\textbf{Weather}} &
\multicolumn{5}{c|}{\textbf{Digital}} &
\multirow{2}{*}{\textbf{Corrup.}} &
\multirow{2}{*}{\textbf{Clean}} 
\\
&  &  
\scriptsize{Gauss.} & \scriptsize{Shot} & \scriptsize{Impulse} & \scriptsize{Speckle} &
\scriptsize{Defocus} & \scriptsize{Gaussian} &
\scriptsize{Snow} & \scriptsize{Frost} & \scriptsize{Fog} & \scriptsize{Spatter} &
\scriptsize{Bright} & \scriptsize{Contrast} & \scriptsize{Saturate} & \scriptsize{Pixel} & \scriptsize{JPEG} &  &
\\
\midrule

\multirow{6}{*}{ Baseline} &

1 &
45.0 & 44.8 & 42.0 & 45.9&
41.0 & 45.5 &
39.3& 42.3& 43.7& 48.3 &
48.8 & 46.3& 48.9 & 47.7& 46.2 &
45.0 & \multirow{6}{*}{49.9} 
\\

& 
2 &
40.7 & 41.1  & 38.1 & 44.3&
35.9 & 38.1 &
28.8& 34.6& 41.9& 42.7 &
48.1 &44.4 & 48.1 & 46.8& 44.5 &
41.2 & 
\\
& 
3 &
35.3 & 35.5 & 35.2 & 37.9&
27.7 & 29.8 &
28.8 & 29.1& 38.3& 37.5 &
47.3& 40.9& 48.4 &43.3 & 43.8 &
37.3 &
\\
& 
4 &
27.9 & 27.8 & 27.9 & 34.1&
21.6 & 23.6 &
23.6 & 27.9& 35.7& 35.1 &
46.1 &31.3 & 46.5 &36.5 & 40.4 &
32.4 &
\\
& 
5 &
19.7 & 22.7 & 21.8 & 29.4&
17.1 & 16.1 &
23.9 &25.4 & 27.6& 29.0 &
44.1 &20.5 &43.9 &33.6& 37.0 &
27.5 &
\\
\cline{2-18}
&
avg. &
33.7 & 34.4 & 33.0 & 38.3 &
28.7 & 30.6 &
28.9 & 31.9 & 37.4 & 38.5 &
46.9 & 36.7 & 47.2 & 41.6 &42.4 &
36.7 &
\\
\hline

\midrule
\multirow{6}{*}{\ours} &

1 &
47.9 & 48.5 & 45.7 & 49.5 &
46.9 & 50.4 &
44.2& 47.0 & 49.3& 51.0 &
52.3& 50.8& 52.1 &  50.7& 49.8 &
49.1 & \multirow{6}{*}{53.0} 

\\

& 
2 &
44.7 & 45.2 & 42.4 & 47.8 &
43.4 & 44.9 &
35.7 & 40.6& 47.4& 46.0 &
51.7 &49.6 & 51.6 & 50.5 & 48.8 &
46.0 &
\\
& 
3 &
39.5& 40.2 & 39.5 & 41.9 &
36.5 & 39.2 &
35.3& 34.6& 44.9& 42.1 &
51.5 &46.8 & 51.7 & 45.5& 47.0 &
42.4 &
\\
& 
4 &
31.4& 30.9 & 30.8 & 38.3 &
29.2 & 31.6 &
30.1 &33.2 & 43.5& 39.1 & 
50.3& 37.5& 50.1 & 38.0& 43.5 &
37.2 &
\\
& 
5 &
20.4 & 24.3 & 22.5 & 32.6 &
22.6 & 19.3 &
31.1 & 29.5& 36.2& 32.6 &
48.2 &17.4 & 39.2& 34.9& 39.2 &
30.0 & 
\\
\cline{2-18}
&
avg. &
\textbf{36.8} & \textbf{37.8} & \textbf{36.2} & \textbf{42.0} &
\textbf{35.7} & \textbf{37.1}&
\textbf{35.3} & \textbf{37.0} & \textbf{44.3}& \textbf{42.2} &
\textbf{50.8} & \textbf{40.4} & \textbf{48.9}& \textbf{43.9} & \textbf{46.1} &
\textbf{40.9} &
\\
\hline

\bottomrule
\end{tabular}
}
\end{table*}
\subsection{Analyzing Our Method}

\noindent\textbf{Robustness evaluation.} Here, we construct a new benchmark for semantic correspondence estimation robustness, named \textit{SPair-C}, following the regime~\cite{hendrycks2019benchmarking}. We highlight that \textit{SPair-C} is the first dataset with corruption and noise for dense correspondence learning and will be useful for future evaluations. We verify our method's robustness by evaluating whether the model can robustly predict correspondences on corrupted images. Additional details for the dataset are provided in the supplementary material.

Tab.~\ref{tab:rob_PCK} shows the overall PCK values throughout 15 corruptions in SPair-C for \ours with CATs versus CATs. We observe that  \ours consistently outperforms the baseline~\cite{cho2021semantic} in terms of the average PCK values across different severities of a single corruption (located in the column of the table) and across different types of corruption with the same severity (shown in the row of the table). Moreover, the average PCK for all 75 corruptions of \ours is 40.9, which represents a 4.2 improvement over the baseline, surpassing the gap of 3.1 observed in the clean SPair dataset. This improvement is presumably attributed to novel labeled keypoints mined by our method, which allowed the model to extract more robust features, even in the presence of corrupted pixels. As a result, the model's robustness to typical corruptions, commonly found in natural images, further reinforces our generalization capability effectively.  

\begin{figure}[t]

\begin{center}
    \centering  
\includegraphics[width=0.75\columnwidth]{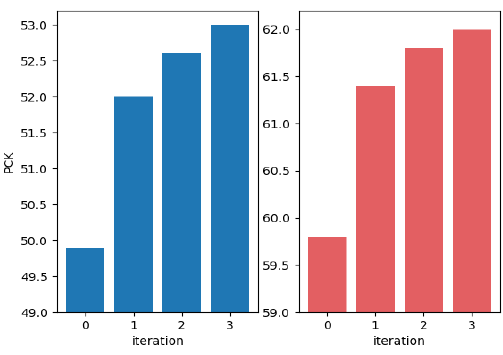}\qquad \hfill
\end{center}
    \vspace{-1.5em}
  \caption{\textbf{PCK at each iteration in iterative training.} We report PCK values at each iteration to show the effectiveness of our training framework. We use identical architecture for the teacher and student and set the iterative training interval to 50 epochs for simplicity. The left and right figures are the results of \ours trained upon CATs and CATs++ backbones, respectively. This indicates that the baseline models were, in fact, undertrained and possess the capacity for further training, highlighting the data-hungry problem.} 
  \label{fig:iter}
\end{figure}

\noindent\textbf{Impact of iterative training.} 
Fig.~\ref{fig:iter} shows the improved PCK values as iterative training progresses, which empirically proves the effectiveness of our iterative training framework. 
We verify the effectiveness using two baseline models, including CATs and CATs++. As observed in Fig.~\ref{fig:iter}, performance consistently improves with each iteration, and the difference between the first and the last iterations is nearly 3\% under the same training hyperparameters. This suggests that improved annotations from stronger models on challenging images by more intense data augmentations can effectively boost performance. This exposes the data-hungry issue in earlier baselines, suggesting they were likely undertrained.

\noindent\textbf{Training time efficiency.}
A pertinent question regarding our method is the extent of overall training time required by our iterative learning protocol. Fortunately, due to the efficiency of our approach to handling the expanded data, this is not a significant concern.
Our method's multi-iteration training with more data may increase the training time, but our rapid convergence (8 for ours vs. 42 epochs for \cite{huang2022learning}) offsets this, maintaining time efficiency despite repeated iterations.
 
\noindent\textbf{PCK analysis by variation factors.}
The averaged PCK is insufficient to demonstrate the performance of the matching models precisely because it is evaluated without considering variation factors. Therefore, the desirable model should show an even PCK performance among the diverse difficulty levels under various factors. To confirm the accurate performance comparison, we conduct the PCK analysis based on the variation factors and difficulty levels on SPair-71k in Tab.~\ref{tab:PCK_difficulty_variation}. 
In experiments, our models show robustness by maintaining higher PCK values consistently across different difficulty levels for various factors compared to the other models. This proves that a large amount of unlabeled data used by our method can cover a wide range of data distribution, including diverse difficulty levels and scene variations.

\begin{table*}[t]
\centering
\caption{\textbf{PCK analysis of state-of-the-art methods on SPair-71k.} All methods commonly show lower PCK than average PCK as difficulty levels of labeled data become more difficult, but \ours shows the best at each level in all but two.}
\label{tab:PCK_difficulty_variation}
\resizebox{1\linewidth}{!}{
\small
\begin{tabular}{l|c| ccc| ccc| cccc| cccc}
\toprule 

\multirow{2}{*}{Methods} & 
\multirow{2}{*}{\textbf{All}} & 
\multicolumn{3}{c|}{\textbf{View-point}} & 
\multicolumn{3}{c|}{\textbf{Scale}} &
\multicolumn{4}{c|}{\textbf{Truncation}} &
\multicolumn{4}{c}{\textbf{Occlusion}} 
 \\
&  & 
\scriptsize{Easy} & \scriptsize{Medi} & \scriptsize{Hard} &
\scriptsize{Easy} & \scriptsize{Medi} & \scriptsize{Hard} &
\scriptsize{None} & \scriptsize{Src} & \scriptsize{Tgt} & \scriptsize{Both} &
\scriptsize{None} & \scriptsize{Src} & \scriptsize{Tgt} & \scriptsize{Both}
\\
\midrule
NC-NET~\cite{rocco2020ncnet} &
26.4 &
34.0 & 18.6 & 12.8 & 
31.7 & 23.8 & 14.2 &
29.1 & 22.9 & 23.4  &21.0 &
29.0 & 21.1 & 21.8 & 19.6 
\\
HPF~\cite{min2019hyperpixel} &
28.2 & 
35.6 & 20.3 & 15.5 & 
33.0 & 26.1 & 15.8 & 
31.0 & 24.6 & 24.0 & 23.7 &
30.8 & 23.5 & 22.8 & 21.8 
 
\\
CATs~\cite{cho2021semantic}&
49.9 & 
54.6 & 44.5 & 43.6 & 
54.5 & 49.2 & 36.3 &
53.7 & 42.0 & 49.4 & 40.6 &
52.7 & 44.5 & 44.9 & 42.0
\\
CATs++~\cite{cho2022cats++}&
{59.8} & 
{63.5} & {55.9} & {53.0} & 
{62.8} & {59.6} & {50.2} &
{61.8} & \textbf{55.2} & {56.3} & {54.8} &
{63.6} & {52.4} & \textbf{57.9} & {50.8}
\\

SemiMatch~\cite{kim2022semimatch} & 
50.8 & 
54.8 & 44.1 & 46.2 & 
55.3 & 50.2 & 36.6 &
54.2 & 43.2 & 50.0 & 42.7 &
53.6 & 45.1 & 44.9 & 43.3
\\
SCORRSAN~\cite{huang2022learning} & 
55.3 & 
59.2 & 51.2 & 48.9 & 
58.7 & 55.0 & 45.0 &
59.2 & 46.1 & 55.0 & 46.9 &
57.8 & 50.2 & 50.7 & 48.7
\\
\midrule
\rowcolor{gray!10} 
\rowcolor{gray!10} 
\ours (ours) &
\textbf{62.0} &
\textbf{66.0} & \textbf{57.9} & \textbf{55.0} & 
\textbf{64.7} & \textbf{61.4} & \textbf{54.4} &
\textbf{65.5} & {54.5} & \textbf{60.9} & \textbf{55.1} &
\textbf{64.3} & \textbf{57.3} & {57.6} & \textbf{56.2}
\\
\bottomrule
\end{tabular}
} 
\end{table*}

\begin{table}[t]
    \centering
    \caption{\textbf{Ablation study with the components}. We perform a factor analysis of the elements used for training \ours-CATs. We compute PCKs on SPair-71k for each component. We observe all the components contribute to the PCK improvements.}
    \label{tab:ab}
    \small
    \tabcolsep=1.5em
   \begin{tabular}{r|lcc}
        \toprule
         & \multicolumn{1}{c}{\multirow{2}{*}{Components}} &\multicolumn{2}{c}{$\alpha_\text{bbox}$} 
         \\
        & & $0.05$ &  $0.10$
         \\
        \midrule

        (a) & \ours (ours)  &\textbf{29.6} & \textbf{53.0}\\
        \midrule 
        (b) & (a) - Iterative training  &28.8 {\scriptsize  (-0.8)} & 52.0 {\scriptsize  (-1.0)}\\
        (c) & (b) - Data noise &28.5 {\scriptsize  (-0.3)} & 51.6 {\scriptsize  (-0.4)}\\
        (d) & (c) - Unpaired data  &26.9 {\scriptsize  (-1.6)} & 49.9 {\scriptsize  (-1.7)}\\

        \bottomrule
    \end{tabular}%
\end{table}

\noindent\textbf{Ablation study.} 
We comprehensively analyze each component in our method in Tab.~\ref{tab:ab}. For a fair comparison, we train all the variants on SPair-71k under the same experimental setting. 
The ablation results show the impacts of each component consisting of \ours.
Compared to (a) \ours, (b) demonstrates the impact of iterative training by running the same epochs to show the impact (\ie,  (b) trained at once, and (a) trained for a total of three iterative training divided by 50 epochs). 
(c) shows that data noise also contributes to the performance of \ours.
(d) shows the benefits that novel generated labels are densified at pixel-level and image-level, respectively, by showing a large margin of 1.6, compared to (c).
The result demonstrates that the data-hungry issue in the semantic correspondence task is the most crucial in performance degradation since their performance gains are more significant than others.

\section{Related Work}

\noindent\textbf{Semantic correspondence learning.} 
Recent methods for semantic correspondence~\cite{liu2020semantic,li2020correspondence,min2021convolutional,li2021probabilistic,zhao2021multi,min2020learning,cho2021semantic,kim2022transformatcher,cho2022cats++,kim2024efficient} inevitably train complicated matching networks to maximize performance in a supervised manner with limited qualified dataset~\cite{ham2017proposal,min2019spair}, which leads to high computational demands and poor generalization capability across datasets.  
 
Some unsupervised strategies~\cite{laskar2018semi,truong2022probabilistic,kim2022semimatch,huang2022learning} extend their unsupervised loss to the supervised regime and significantly improve the performance of the previous supervised approaches. This shows that the performance of the existing supervised model was not fully learned due to a lack of data. Specifically, the methods~\cite{laskar2018semi,truong2022probabilistic} use a cycle consistency for unsupervised loss signal, and the others~\cite{kim2022semimatch,huang2022learning} utilize pseudo-labels, generated by the model's prediction between real images, combined with confidence measures to guarantee the quality of pseudo-labels.
Recently, methods~\cite{luo2024diffusion,zhang2024tale,hedlin2024unsupervised} have emerged that implement unsupervised semantic correspondence by tapping into the inherent knowledge embedded in pre-trained diffusion models trained on large text-image datasets to facilitate semantic correspondence. Leveraging the knowledge in pre-trained models for unseen data may share a similar spirit, but our framework does not require billion-scale data for generative modeling like those. Our method employs a lighter pre-trained model as a guidance labeler to address the scarcity of both image and point pairs by utilizing massive unlabeled data through the pre-trained model.

\noindent\textbf{Semi-supervised learning methods for semantic correspondence learning.} 
Previous literature~\cite{tarvainen2017mean,xie2020self,pham2021meta,li2022rethinking} were proposed to utilize a teacher-student structure mainly for semi-supervised learning. Those methods are distinguished from the earlier methods pseudo-labeling~\cite{lee2013pseudo,shi2018transductive} and consistency regularization methods~\cite{rasmus2015semi} due to their use of data and the resulting performance achieved. A teacher model, generally trained on a small set of labeled data, generates pseudo-labels on a larger unlabeled data to guide the student model, and then the student is jointly trained on a combination of labeled and pseudo-labeled images. 
Recent studies have applied the teacher-student framework for pixel-level semi-supervised learning, specifically for the semantic correspondence task~\cite{li2021probabilistic,huang2022learning}. 
They employ a teacher model to generate additional pseudo-labels using knowledge from keypoint periphery~\cite{huang2022learning} or hypotheses~\cite{li2021probabilistic} across labeled image pairs. On the other hand, our method labels overlooked unlabeled data using a machine annotator, continually repeating the process by assigning the learned student back to the teacher.

\section{Conclusion}
In this paper, we have proposed a simple baseline that leverages unpaired images for semantic correspondence learning. 
Instead of using a sizeable complicated model with strong data augmentations to augment paired images, we have aimed to break the stereotype of using given labeled image pairs by expanding the training pairs with machine-annotated unpaired images. Only with a machine-annotation-based framework for labeling the unpaired images, our method could beat the state-of-the-art models on SPair-71k, PF-PASCAL, and PF-WILLOW by large margins. Additionally, our approach could continuously improve performance by repeating the training process with increasingly challenging image pairs after each step. It also turns out that a resultant model has become more robust to corrupted images.

\noindent\textbf{Limitations.}
Going beyond the scale of Spair-71k and tackling more challenging datasets unrelated to the semantic correspondence task would reveal a more generalized impact of our work. Furthermore, an exciting direction can be utilizing recently proposed Transformer-based architectures to deal with unpaired data with expanded data. 

\bibliographystyle{splncs04}
\bibliography{main}

\begin{thebibliography}{10}
\providecommand{\url}[1]{\texttt{#1}}
\providecommand{\urlprefix}{URL }
\providecommand{\doi}[1]{https://doi.org/#1}

\bibitem{bristow2015dense}
Bristow, H., Valmadre, J., Lucey, S.: Dense semantic correspondence where every pixel is a classifier. In: Proceedings of the IEEE International Conference on Computer Vision. pp. 4024--4031 (2015)

\bibitem{cho2021semantic}
Cho, S., Hong, S., Jeon, S., Lee, Y., Sohn, K., Kim, S.: Semantic correspondence with transformers. arXiv preprint arXiv:2106.02520  (2021)

\bibitem{cho2022cats++}
Cho, S., Hong, S., Kim, S.: Cats++: Boosting cost aggregation with convolutions and transformers. arXiv preprint arXiv:2202.06817  (2022)

\bibitem{coates2011analysis}
Coates, A., Ng, A., Lee, H.: An analysis of single-layer networks in unsupervised feature learning. In: AISTATS (2011)

\bibitem{everingham2015pascal}
Everingham, M., Eslami, S.A., Van~Gool, L., Williams, C.K., Winn, J., Zisserman, A.: The pascal visual object classes challenge: A retrospective. International journal of computer vision  \textbf{111}(1),  98--136 (2015)

\bibitem{ham2017proposal}
Ham, B., Cho, M., Schmid, C., Ponce, J.: Proposal flow: Semantic correspondences from object proposals. IEEE Transactions on Pattern Analysis and Machine Intelligence  \textbf{40}(7),  1711--1725 (2017)

\bibitem{han2017scnet}
Han, K., Rezende, R.S., Ham, B., Wong, K.Y.K., Cho, M., Schmid, C., Ponce, J.: Scnet: Learning semantic correspondence. In: Proceedings of the IEEE International Conference on Computer Vision. pp. 1831--1840 (2017)

\bibitem{hedlin2024unsupervised}
Hedlin, E., Sharma, G., Mahajan, S., Isack, H., Kar, A., Tagliasacchi, A., Yi, K.M.: Unsupervised semantic correspondence using stable diffusion. Advances in Neural Information Processing Systems  \textbf{36} (2024)

\bibitem{hendrycks2019benchmarking}
Hendrycks, D., Dietterich, T.: Benchmarking neural network robustness to common corruptions and perturbations. arXiv preprint arXiv:1903.12261  (2019)

\bibitem{hosni2012fast}
Hosni, A., Rhemann, C., Bleyer, M., Rother, C., Gelautz, M.: Fast cost-volume filtering for visual correspondence and beyond. IEEE Transactions on Pattern Analysis and Machine Intelligence  \textbf{35}(2),  504--511 (2012)

\bibitem{huang2019dynamic}
Huang, S., Wang, Q., Zhang, S., Yan, S., He, X.: Dynamic context correspondence network for semantic alignment. In: Proceedings of the IEEE International Conference on Computer Vision. pp. 2010--2019 (2019)

\bibitem{huang2022learning}
Huang, S., Yang, L., He, B., Zhang, S., He, X., Shrivastava, A.: Learning semantic correspondence with sparse annotations. In: Computer Vision--ECCV 2022: 17th European Conference, Tel Aviv, Israel, October 23--27, 2022, Proceedings, Part XIV. pp. 267--284. Springer (2022)

\bibitem{hui2018liteflownet}
Hui, T.W., Tang, X., Loy, C.C.: Liteflownet: A lightweight convolutional neural network for optical flow estimation. In: Proceedings of the IEEE Conference on Computer Vision and Pattern Recognition. pp. 8981--8989 (2018)

\bibitem{hur2015generalized}
Hur, J., Lim, H., Park, C., Chul~Ahn, S.: Generalized deformable spatial pyramid: Geometry-preserving dense correspondence estimation. In: Proceedings of the IEEE Conference on Computer Vision and Pattern Recognition. pp. 1392--1400 (2015)

\bibitem{ilg2017flownet}
Ilg, E., Mayer, N., Saikia, T., Keuper, M., Dosovitskiy, A., Brox, T.: Flownet 2.0: Evolution of optical flow estimation with deep networks. In: Proceedings of the IEEE conference on computer vision and pattern recognition. pp. 2462--2470 (2017)

\bibitem{kim2022semimatch}
Kim, J., Ryoo, K., Seo, J., Lee, G., Kim, D., Cho, H., Kim, S.: Semi-supervised learning of semantic correspondence with pseudo-labels. In: Proceedings of the IEEE/CVF Conference on Computer Vision and Pattern Recognition. pp. 19699--19709 (2022)

\bibitem{kim2019semantic}
Kim, S., Min, D., Jeong, S., Kim, S., Jeon, S., Sohn, K.: Semantic attribute matching networks. In: Proceedings of the IEEE Conference on Computer Vision and Pattern Recognition. pp. 12339--12348 (2019)

\bibitem{kim2022transformatcher}
Kim, S., Min, J., Cho, M.: Transformatcher: Match-to-match attention for semantic correspondence. In: Proceedings of the IEEE/CVF Conference on Computer Vision and Pattern Recognition. pp. 8697--8707 (2022)

\bibitem{kim2024efficient}
Kim, S., Min, J., Cho, M.: Efficient semantic matching with hypercolumn correlation. In: Proceedings of the IEEE/CVF Winter Conference on Applications of Computer Vision. pp. 139--148 (2024)

\bibitem{kokkinos2021learning}
Kokkinos, F., Kokkinos, I.: Learning monocular 3d reconstruction of articulated categories from motion. In: Proceedings of the IEEE/CVF Conference on Computer Vision and Pattern Recognition. pp. 1737--1746 (2021)

\bibitem{laskar2018semi}
Laskar, Z., Kannala, J.: Semi-supervised semantic matching. In: Proceedings of the European Conference on Computer Vision (ECCV) Workshops. pp.~0--0 (2018)

\bibitem{lee2013pseudo}
Lee, D.H., et~al.: Pseudo-label: The simple and efficient semi-supervised learning method for deep neural networks. In: Workshop on challenges in representation learning, ICML. p.~896 (2013)

\bibitem{lee2021patchmatch}
Lee, J.Y., DeGol, J., Fragoso, V., Sinha, S.N.: Patchmatch-based neighborhood consensus for semantic correspondence. In: Proceedings of the IEEE Conference on Computer Vision and Pattern Recognition. pp. 13153--13163 (2021)

\bibitem{lee2019sfnet}
Lee, J., Kim, D., Ponce, J., Ham, B.: Sfnet: Learning object-aware semantic correspondence. In: Proceedings of the IEEE Conference on Computer Vision and Pattern Recognition. pp. 2278--2287 (2019)

\bibitem{lee2020reference}
Lee, J., Kim, E., Lee, Y., Kim, D., Chang, J., Choo, J.: Reference-based sketch image colorization using augmented-self reference and dense semantic correspondence. In: Proceedings of the IEEE Conference on Computer Vision and Pattern Recognition. pp. 5801--5810 (2020)

\bibitem{li2022rethinking}
Li, H., Wu, Z., Shrivastava, A., Davis, L.S.: Rethinking pseudo labels for semi-supervised object detection. In: Proceedings of the AAAI Conference on Artificial Intelligence. pp. 1314--1322 (2022)

\bibitem{li2020correspondence}
Li, S., Han, K., Costain, T.W., Howard-Jenkins, H., Prisacariu, V.: Correspondence networks with adaptive neighbourhood consensus. In: Proceedings of the IEEE Conference on Computer Vision and Pattern Recognition. pp. 10196--10205 (2020)

\bibitem{li2021probabilistic}
Li, X., Fan, D.P., Yang, F., Luo, A., Cheng, H., Liu, Z.: Probabilistic model distillation for semantic correspondence. In: Proceedings of the IEEE Conference on Computer Vision and Pattern Recognition. pp. 7505--7514 (2021)

\bibitem{li2020online}
Li, X., Liu, S., De~Mello, S., Kim, K., Wang, X., Yang, M.H., Kautz, J.: Online adaptation for consistent mesh reconstruction in the wild. Advances in Neural Information Processing Systems  \textbf{33},  15009--15019 (2020)

\bibitem{liu2010sift}
Liu, C., Yuen, J., Torralba, A.: Sift flow: Dense correspondence across scenes and its applications. IEEE Transactions on Pattern Analysis and Machine Intelligence  \textbf{33}(5),  978--994 (2010)

\bibitem{liu2020semantic}
Liu, Y., Zhu, L., Yamada, M., Yang, Y.: Semantic correspondence as an optimal transport problem. In: Proceedings of the IEEE Conference on Computer Vision and Pattern Recognition. pp. 4463--4472 (2020)

\bibitem{luo2024diffusion}
Luo, G., Dunlap, L., Park, D.H., Holynski, A., Darrell, T.: Diffusion hyperfeatures: Searching through time and space for semantic correspondence. Advances in Neural Information Processing Systems  \textbf{36} (2024)

\bibitem{melekhov2019dgc}
Melekhov, I., Tiulpin, A., Sattler, T., Pollefeys, M., Rahtu, E., Kannala, J.: Dgc-net: Dense geometric correspondence network. In: IEEE Winter Conference on Applications of Computer Vision. pp. 1034--1042. IEEE (2019)

\bibitem{min2021convolutional}
Min, J., Cho, M.: Convolutional hough matching networks. In: Proceedings of the IEEE Conference on Computer Vision and Pattern Recognition. pp. 2940--2950 (2021)

\bibitem{min2021hypercorrelation}
Min, J., Kang, D., Cho, M.: Hypercorrelation squeeze for few-shot segmentation. In: Proceedings of the IEEE/CVF International Conference on Computer Vision. pp. 6941--6952 (2021)

\bibitem{min2019hyperpixel}
Min, J., Lee, J., Ponce, J., Cho, M.: Hyperpixel flow: Semantic correspondence with multi-layer neural features. In: Proceedings of the IEEE International Conference on Computer Vision. pp. 3395--3404 (2019)

\bibitem{min2019spair}
Min, J., Lee, J., Ponce, J., Cho, M.: Spair-71k: A large-scale benchmark for semantic correspondence. arXiv preprint arXiv:1908.10543  (2019)

\bibitem{min2020learning}
Min, J., Lee, J., Ponce, J., Cho, M.: Learning to compose hypercolumns for visual correspondence. In: Computer Vision--ECCV 2020: 16th European Conference, Glasgow, UK, August 23--28, 2020, Proceedings, Part XV 16. pp. 346--363. Springer (2020)

\bibitem{pham2021meta}
Pham, H., Dai, Z., Xie, Q., Le, Q.V.: Meta pseudo labels. In: Proceedings of the IEEE/CVF Conference on Computer Vision and Pattern Recognition (2021)

\bibitem{rasmus2015semi}
Rasmus, A., Berglund, M., Honkala, M., Valpola, H., Raiko, T.: Semi-supervised learning with ladder networks. Advances in Neural Information Processing Systems  \textbf{28} (2015)

\bibitem{rizve2021defense}
Rizve, M.N., Duarte, K., Rawat, Y.S., Shah, M.: In defense of pseudo-labeling: An uncertainty-aware pseudo-label selection framework for semi-supervised learning. arXiv:2101.06329  (2021)

\bibitem{rocco2017convolutional}
Rocco, I., Arandjelovic, R., Sivic, J.: Convolutional neural network architecture for geometric matching. In: Proceedings of the IEEE Conference on Computer Vision and Pattern Recognition. pp. 6148--6157 (2017)

\bibitem{rocco2020ncnet}
Rocco, I., Cimpoi, M., Arandjelovic, R., Torii, A., Pajdla, T., Sivic, J.: Ncnet: Neighbourhood consensus networks for estimating image correspondences. IEEE Transactions on Pattern Analysis and Machine Intelligence  (2020)

\bibitem{russakovsky2015imagenet}
Russakovsky, O., Deng, J., Su, H., Krause, J., Satheesh, S., Ma, S., Huang, Z., Karpathy, A., Khosla, A., Bernstein, M., et~al.: Imagenet large scale visual recognition challenge. IJCV  (2015)

\bibitem{shi2018transductive}
Shi, W., Gong, Y., Ding, C., Tao, Z.M., Zheng, N.: Transductive semi-supervised deep learning using min-max features. In: Proceedings of the European Conference on Computer Vision. pp. 299--315 (2018)

\bibitem{sohn2020fixmatch}
Sohn, K., Berthelot, D., Carlini, N., Zhang, Z., Zhang, H., Raffel, C.A., Cubuk, E.D., Kurakin, A., Li, C.L.: Fixmatch: Simplifying semi-supervised learning with consistency and confidence. In: Advances in Neural Information Processing Systems (2020)

\bibitem{sun2018pwc}
Sun, D., Yang, X., Liu, M.Y., Kautz, J.: Pwc-net: Cnns for optical flow using pyramid, warping, and cost volume. In: Proceedings of the IEEE Conference on Computer Vision and Pattern Recognition. pp. 8934--8943 (2018)

\bibitem{tang2023emergent}
Tang, L., Jia, M., Wang, Q., Phoo, C.P., Hariharan, B.: Emergent correspondence from image diffusion. arXiv preprint arXiv:2306.03881  (2023)

\bibitem{tarvainen2017mean}
Tarvainen, A., Valpola, H.: Mean teachers are better role models: Weight-averaged consistency targets improve semi-supervised deep learning results. In: Advances in Neural Information Processing Systems (2017)

\bibitem{truong2020glu}
Truong, P., Danelljan, M., Timofte, R.: Glu-net: Global-local universal network for dense flow and correspondences. In: Proceedings of the IEEE Conference on Computer Vision and Pattern Recognition. pp. 6258--6268 (2020)

\bibitem{truong2021warp}
Truong, P., Danelljan, M., Yu, F., Van~Gool, L.: Warp consistency for unsupervised learning of dense correspondences. In: Proceedings of the IEEE International Conference on Computer Vision. pp. 10346--10356 (2021)

\bibitem{truong2022probabilistic}
Truong, P., Danelljan, M., Yu, F., Van~Gool, L.: Probabilistic warp consistency for weakly-supervised semantic correspondences. arXiv preprint arXiv:2203.04279  (2022)

\bibitem{xie2021few}
Xie, G.S., Xiong, H., Liu, J., Yao, Y., Shao, L.: Few-shot semantic segmentation with cyclic memory network. In: Proceedings of the IEEE/CVF International Conference on Computer Vision. pp. 7293--7302 (2021)

\bibitem{xie2020unsupervised}
Xie, Q., Dai, Z., Hovy, E., Luong, T., Le, Q.: Unsupervised data augmentation for consistency training. In: Advances in Neural Information Processing Systems (2020)

\bibitem{xie2020self}
Xie, Q., Luong, M.T., Hovy, E., Le, Q.V.: Self-training with noisy student improves imagenet classification. In: Proceedings of the IEEE/CVF Conference on Computer Vision and Pattern Recognition (2020)

\bibitem{xu2021dash}
Xu, Y., Shang, L., Ye, J., Qian, Q., Li, Y.F., Sun, B., Li, H., Jin, R.: Dash: Semi-supervised learning with dynamic thresholding. In: International Conference on Machine Learning (2021)

\bibitem{yang2019volumetric}
Yang, G., Ramanan, D.: Volumetric correspondence networks for optical flow. Advances in Neural Information Processing Systems  \textbf{32} (2019)

\bibitem{yun2021re}
Yun, S., Oh, S.J., Heo, B., Han, D., Choe, J., Chun, S.: Re-labeling imagenet: from single to multi-labels, from global to localized labels. In: Proceedings of the IEEE/CVF Conference on Computer Vision and Pattern Recognition. pp. 2340--2350 (2021)

\bibitem{zhang2021flexmatch}
Zhang, B., Wang, Y., Hou, W., Wu, H., Wang, J., Okumura, M., Shinozaki, T.: Flexmatch: Boosting semi-supervised learning with curriculum pseudo labeling. In: Advances in Neural Information Processing Systems (2021)

\bibitem{zhang2024tale}
Zhang, J., Herrmann, C., Hur, J., Polania~Cabrera, L., Jampani, V., Sun, D., Yang, M.H.: A tale of two features: Stable diffusion complements dino for zero-shot semantic correspondence. Advances in Neural Information Processing Systems  \textbf{36} (2024)

\bibitem{zhao2021multi}
Zhao, D., Song, Z., Ji, Z., Zhao, G., Ge, W., Yu, Y.: Multi-scale matching networks for semantic correspondence. In: Proceedings of the IEEE International Conference on Computer Vision. pp. 3354--3364 (2021)

\end{thebibliography}
\clearpage
{\noindent \large \textbf{Appendix} \\}

\appendix

\renewcommand\thefigure{\Alph{figure}}    
\setcounter{figure}{0}  
\renewcommand\thetable{\Alph{table}}    
\setcounter{table}{0}

\noindent In this appendix, we provide additional information that complements the materials in our main paper and various aspects of our research. Specifically, we include the following items:
\begin{itemize}
  \item \textbf{Novel Robustness Evaluation Benchmark.} We introduce a new robustness benchmark for semantic correspondence (dubbed \textbf{SPair-C}). To our knowledge, this is the first benchmark for evaluating the robustness of semantic correspondence learning methods.
  \item \textbf{Further Analyses.} We present an ablation study and further analyses to understand the effectiveness of our method. %
    \item \textbf{Further Training Details.} We present further training details that elaborate on our method of leveraging unlabeled data to boost performance.
  \item \textbf{Visualizations.} We showcase qualitative results by comparing our method with state-of-the-art methods. 

\end{itemize}

\section{Robustness Evaluation Benchmark}
\label{robustness}
We introduce a new corruption benchmark to verify the robustness of dense correspondence learning methods. This benchmark is dubbed SPair-C (\ie, a corrupted SPair-71k~\cite{min2019spair}), which was mentioned in the Robustness evaluation section of the main paper. Its purpose is to complement the existing dense correspondence learning task by providing a more challenging dataset for evaluating the robustness of models.

\noindent\textbf{Dataset details.} 
Since hardly corrupted or noise-injected images have been used to evaluate dense correspondence learning, we construct a new corruption robustness benchmark for semantic correspondence following the existing regime~\cite{hendrycks2019benchmarking}. The future applicability of a model can be determined by evaluating its robustness against corrupted images involving frequently observed corruption occurring in the wild.

\begin{figure*}[t]
\small
\centering
\includegraphics[width=1\linewidth]{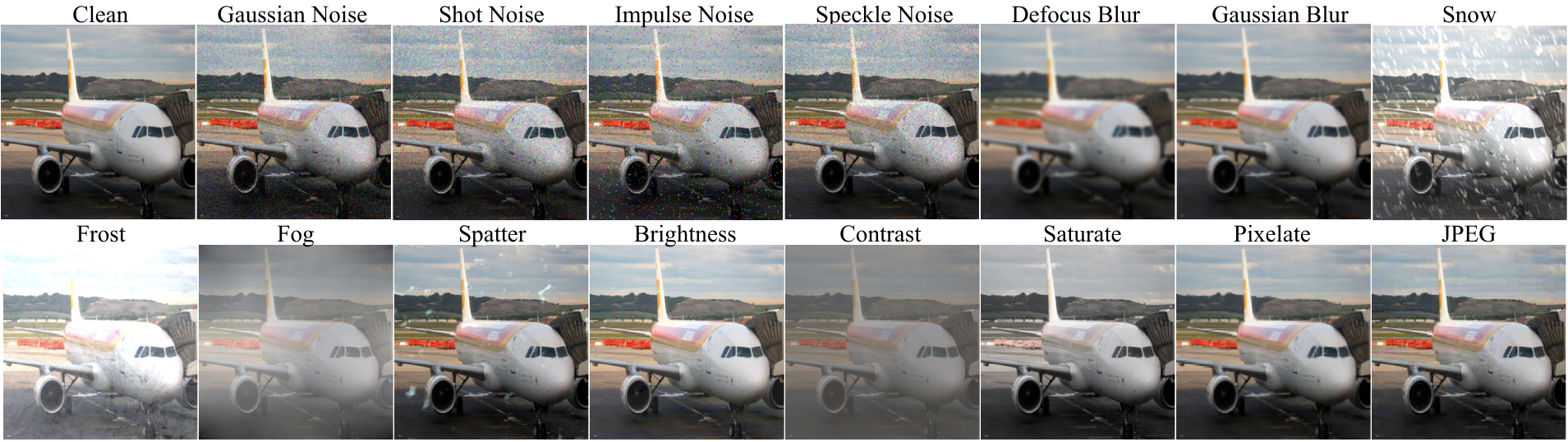}

\caption{\textbf{Visualization of corrupted images in SPair-C.} The corrupted images of one sample consist of types of algorithmically generated corruptions from noise, blur, weather, and digital categories. Each type of corruption has five levels of severity, resulting in 75 distinct corruptions.}
\label{fig:rob_types}
\end{figure*}

\begin{figure*}[t]
\small
\centering
\begin{subfigure}{\textwidth}
\centering
\includegraphics[width=1.0\linewidth]{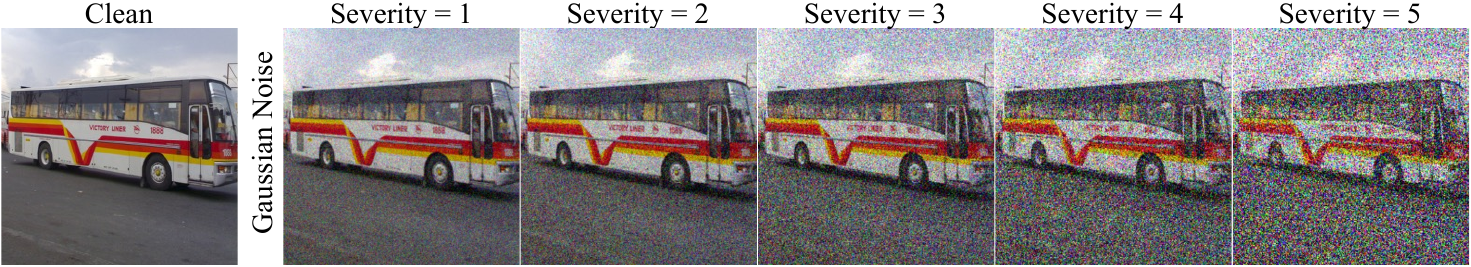}

\end{subfigure}%
\\

\caption{\textbf{Corrupted images with different severities.} We visualize a chosen image from SPair-71k~\cite{min2019spair} with severity from 1 to 5. The images get noisier as the severity increases.  }
\label{fig:rob_severity}
\end{figure*}

Fig.~\ref{fig:rob_types} shows 15 types of corruptions in the SPair-C dataset, selected among corruptions~\cite{hendrycks2019benchmarking} used for measuring robustness. We choose appropriate corruptions for pixel-level prediction from noise, blurred weather, and digital categories, which do not affect significant point changes after corruptions. The finalized categories are noise (Gaussian, shot, impulse, and speckle), blur (defocus and Gaussian), weather (snow, frost, fog, and spatter), and digital categories (brightness, contrast, saturate, pixelated, and JPEG). We apply the five levels of the severity $s$ per each type of corruption exemplified in Fig.~\ref{fig:rob_severity}. Therefore, a set of 75 common visual corruptions, which enable models to be fooled by small changes in the original image, are used for one test image. We use the codebase\footnote{\url{https://github.com/hendrycks/robustness}} to build the benchmark.


\begin{table}[t]
    \centering
    \small
\caption{\textbf{Impact of machine-annotated data.} 
We study the sensitivity of PCK as the quantity and quality of machine-annotated data change by the confidence threshold $\tau$. We adjust $\tau$ from 0.1 to 0.9 and confirm how PCK changes. CATs~\protect\cite{cho2021semantic} is used for this study. We observe the model can yield the maximal PCK with the tuned $\tau$ but shows insensitive PCKs to $\tau$.}
   \begin{tabular}{c|l}
        \toprule
        \multirow{2}{*}{Confidence threshold} & \multirow{2}{*}{PCK}
         \\
        ($\tau$) &  \\
        \midrule
        \textbf{$0.1$} & 52.7  \\
        \textbf{$0.3$} &52.6 \\
        \textbf{$0.5$} & 52.6 \\
        \textbf{$0.7$} &53.0  \\
        \textbf{$0.9$} &52.3  \\
        \bottomrule
    \end{tabular}%
    \label{tab:ab_conf}
\end{table}

\begin{table*}[t]
\centering
\caption{\textbf{PCK comparison with state-the-art methods under varying tolerance ($\alpha$)}. We report PCK for different $\alpha$ of the state-of-the-art methods on SPair-71k~\protect\cite{min2019spair}. Numbers in bold denote the best. \ours outperforms all the methods by a large margin.}
\begin{tabular}{c|cccccccccc}
\toprule
$\alpha$     & 0.01  & 0.02  & 0.03  & 0.04  & 0.05  & 0.06  & 0.07  & 0.08  & 0.09  & 0.1   \\ \midrule
CATs~\cite{cho2021semantic}      & 1.93   & 7.0  & 13.8  & 20.9   & 27.7 & 33.6 & 38.6 & 43.0  & 46.8 & 49.9 \\ 
CATs++~\cite{cho2021semantic}   &4.3 & 14.6 & 25.0 & 33.9 & 40.8 & 46.4 & 50.8 & 54.3 & 57.3 & 59.8 \\
SemiMatch~\cite{kim2022semimatch} &2.1 & 7.7 & 15.0 & 22.4& 29.0  & 34.8 &39.7&43.9 & 47.6 & 50.7 \\ 
SCORRSAN~\cite{huang2022learning} & 3.6 & 12.1 & 21.3 &29.3 & 35.8 & 41.0 & 45.2 & 48.8 & 51.7 & 55.3\\
\midrule
\ours (ours) &\textbf{6.1}  & \textbf{18.5} & \textbf{29.9} & \textbf{38.6} & \textbf{45.1} & \textbf{50.0} & \textbf{53.9} & \textbf{57.0} & \textbf{59.7} & \textbf{62.0}\\
\bottomrule
\end{tabular}

\label{tab:PCK_variation}
\end{table*}

\section{Further Analyses}
\label{analysis}
In this section, we first ablate our model regarding the confidence threshold. Then, we analyze PCK concerning the tolerance threshold $\alpha$. Finally, We examine our proposed method \ours by training it, varying the ratio of unlabeled images to labeled training images.%

\noindent\textbf{Ablation study on confidence threshold.}
The confidence threshold $\tau$ in Eq.(8) is a critical hyper-parameter that determines the quantity and quality of machine-annotated data used for training. To investigate the relationship between the confidence threshold and model performance, we conduct an ablation study with the threshold at intervals of 0.2 across a wide range of confidence thresholds from 0.1 to 0.9. We use CATs~\cite{cho2021semantic} for this study.

As shown in Tab.~\ref{tab:ab_conf}, while the threshold value of 0.7 produces the highest PCK value, other values do not significantly degrade performance and still produce higher PCK values than the baseline's PCK 49.9, which do not use our method. 
The PCK trend suggests that the quantity of machine-annotated data is more critical than its quality, as demonstrated by the lower PCK value at the higher confidence threshold (\eg, $\tau=0.9$) compared to that at the lower threshold (\eg, $\tau \leq0.3$).  Moreover, the low sensitivity to threshold values suggests that our novel data itself contributes to improving model performance.

\begin{table}[t] 
\centering
\vspace{1em}
\caption{\textbf{PCK of \ours trained with a varying fraction of labeled images.} We observe that \ours is not heavily dependent on a large fraction of labeled images. It achieves high PCK even when trained with only 20\% labeled data from the entire labeled images, which amounts to approximately 0.04\% of the unlabeled data.}
\small
\tabcolsep=0.2cm
\begin{tabular}{@{}c|c|c|c|c@{}}
\toprule 
\multicolumn{5}{c}{\textbf{PCK} with fraction of labeled data \scriptsize{(baseline: 49.9)} } \\ \midrule
0.2 & 
0.4 & 
0.6 &
0.8 & 
1.0 \\
\midrule
50.2 \scriptsize{(+0.3)} & 50.5 \scriptsize{(+0.5)} & 50.7 \scriptsize{(+0.8)} & 51.1 \scriptsize{(+1.2)} & 52.0 \scriptsize{(+2.1)}\\
\bottomrule
\end{tabular}

\label{tab:labeled_unlabeled}
\end{table}

\noindent\textbf{PCK analysis.}
We report PCK results with the various tolerance thresholds from 0.1 to 0.01 on SPair-71k in comparison between ours and the competitive baselines~\cite{coates2011analysis,cho2022cats++,kim2022semimatch,huang2022learning} in Tab.~\ref{tab:PCK_variation}. 
The results demonstrate that our expanded keypoint correspondences, amplified at both pixel-level and image-level, enable the trained model to more accurately estimate the correspondences, as evidenced by the significant PCK gaps with much stricter tolerance criteria (smaller $\alpha$ values), compared to other methods. Moreover, the results show that as the tolerance threshold of PCK ($\alpha$) decreases, the gap with the baseline~\cite{cho2022cats++} does not narrow but widens. For example, at $\alpha = 0.1$, the performance gap is 2.2, but at $\alpha=0.05$, it is 4.3, showing a gap approximately twice as large. Therefore, this demonstrates that our predicted point correspondences closely approximate the GT point correspondences. %

\noindent\textbf{Analysis on using labeled/unlabeled data.} 
We conduct further experiments by varying the ratios of unlabeled and existing labeled training data ratios. We aim to determine the necessary amount of labeled data for performance and understand the dependency on labeled data. Table.~\ref{tab:labeled_unlabeled} offers the following observations: 1) the minimum amount of labeled data required is quite low, and 2) there is a low reliance on labeled data. 
For example, when using 20\% images of the entire labeled data, which corresponds to approximately 0.04\% of the unlabeled data, \ours outperforms the baseline (\ie, 49.9), trained with the whole labeled data in a supervised manner, on SPair-71k.%

\section{Further Details on Training}

\noindent\textbf{Unlabeled images for training}
We use the labeled data in the training set of SPair-71k~\cite{min2019spair} along with the unlabeled data in PASCAL VOC 2012~\cite{everingham2015pascal}, which is the source of the SPair-71k. We assign images for each object class according to the corresponding classification labels. Only non-overlapping images in the validation and test set of SPair-71k are used for training to avoid cheating. Furthermore, images in the 'dining table' and 'sofa' classes are not included in the same way that the dining table and sofa categories are not used as in SPair-71k. Since SPair-71k was built to have similar numbers of labeled data for each class, we balance the number of unlabeled data for each class to ensure that their distribution matches that of the labeled data. 
We use different batches of unlabeled data in every iteration to diversify the training sample.

\begin{figure}[ht!]
\centering
    \begin{subfigure}{1\linewidth}
    \centering
        \includegraphics[width=0.48\linewidth]{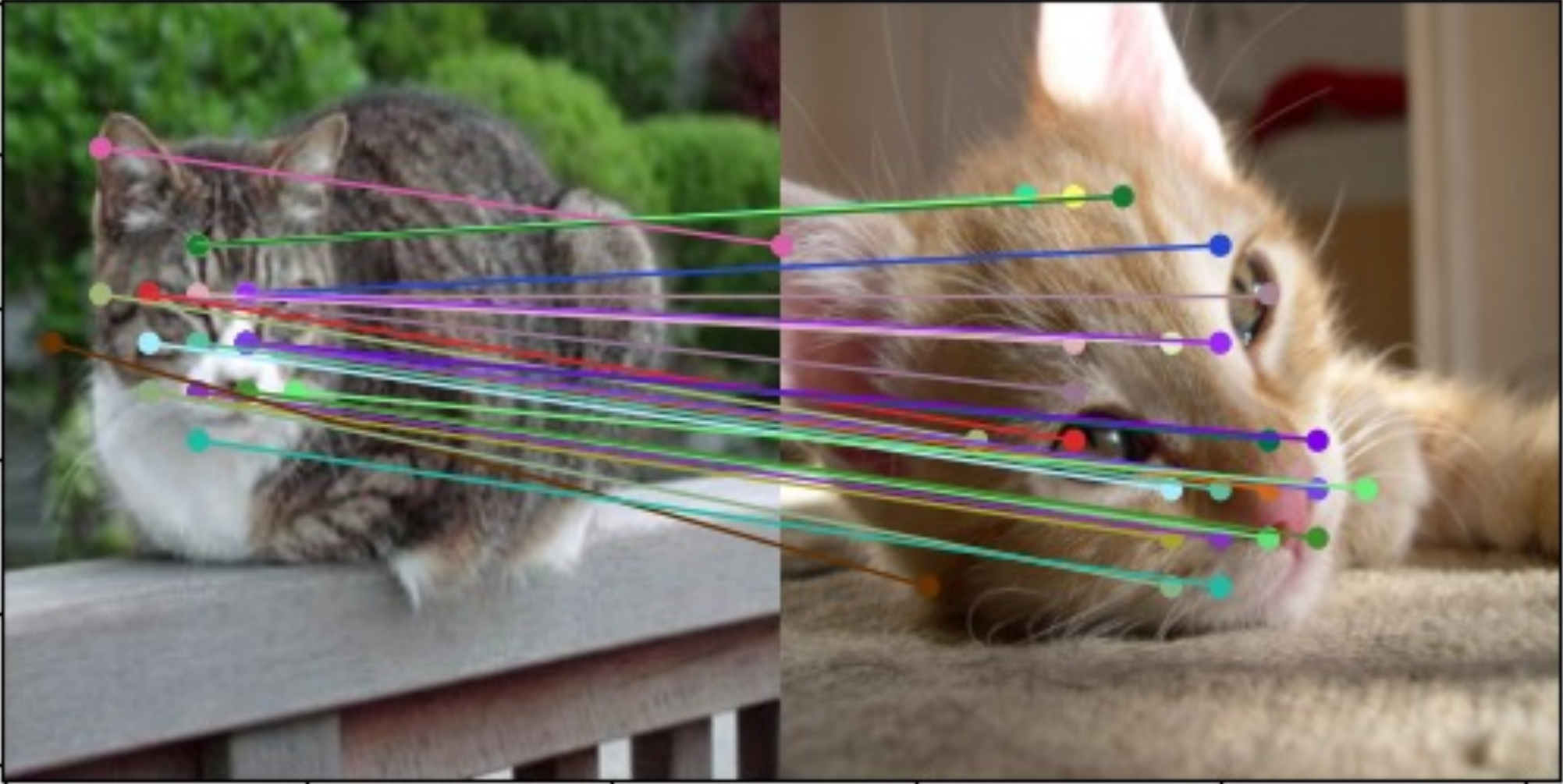} 
        \includegraphics[width=0.48\linewidth]{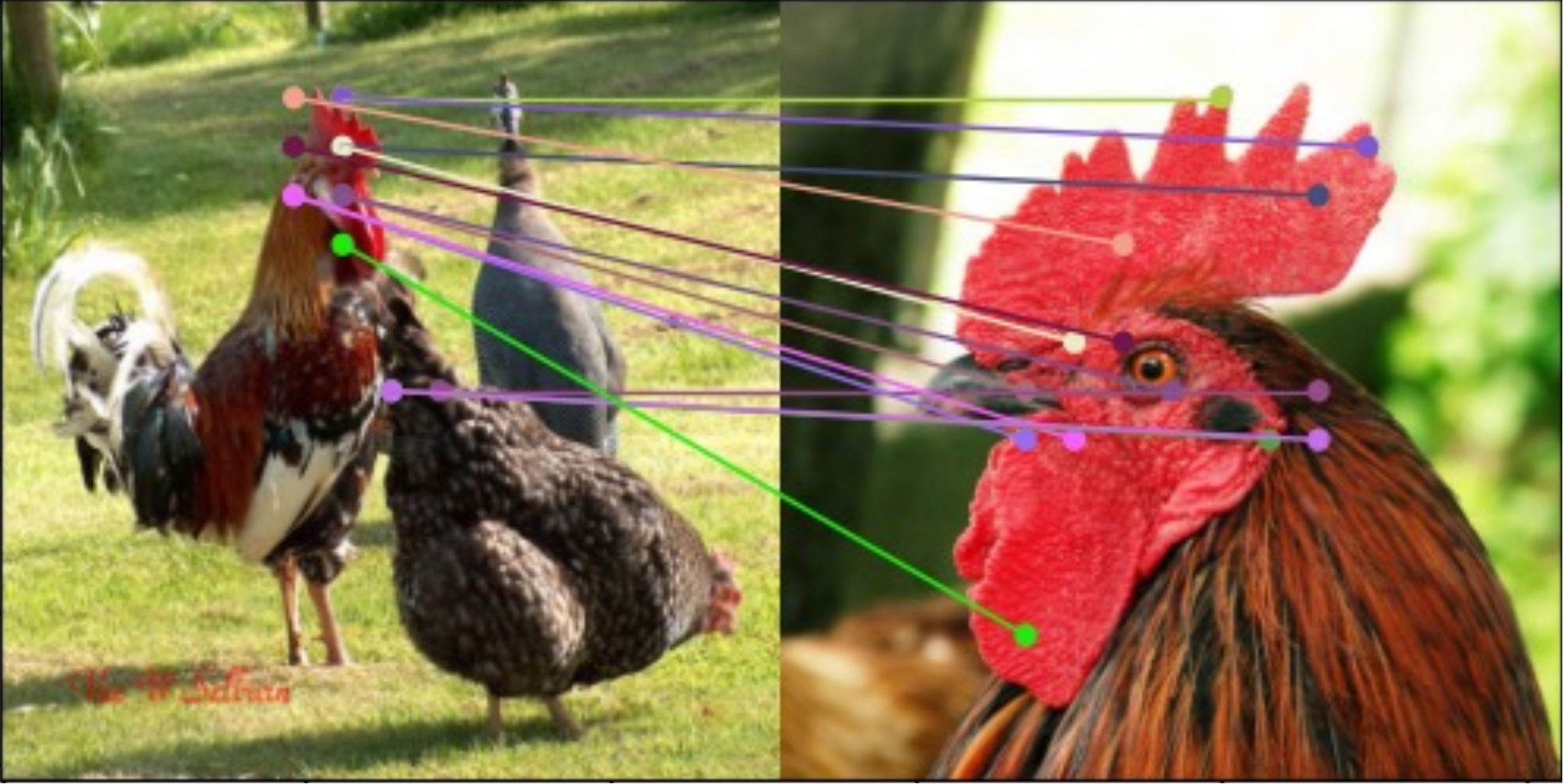}
    \end{subfigure}
\caption{\textbf{Qualitative example of handling unseen images}.
We demonstrate the applicability of \ours using newly acquired data beyond the training dataset. This includes images for both existing and unseen classes, with the images sourced from ImageNet~\protect\cite{russakovsky2015imagenet}. The top two images display matching results for cat images, while the bottom images feature the unseen class, Hen. The results indicate the strong applicability of our method, as evidenced by the high accuracy of the correspondence in both cases.}
\label{fig:seen_unseen}
\end{figure}

\section{Visualizations}
\noindent\textbf{Handling unseen data.} 
We qualitatively verify the generalizability of our method by examining the labels generated by \ours for newly captured data from the ImageNet dataset~\cite{russakovsky2015imagenet}, which were not included in our training dataset.
As shown in Fig.~\ref{fig:seen_unseen}, our model produces high-quality correspondences for both data from known classes (\eg, Cat) and even previously unseen classes (\eg, Hen). This demonstrates the high potential to extend our method by incorporating newly captured data into the existing unlabeled data for training or evaluation.

\begin{figure*}[t!]
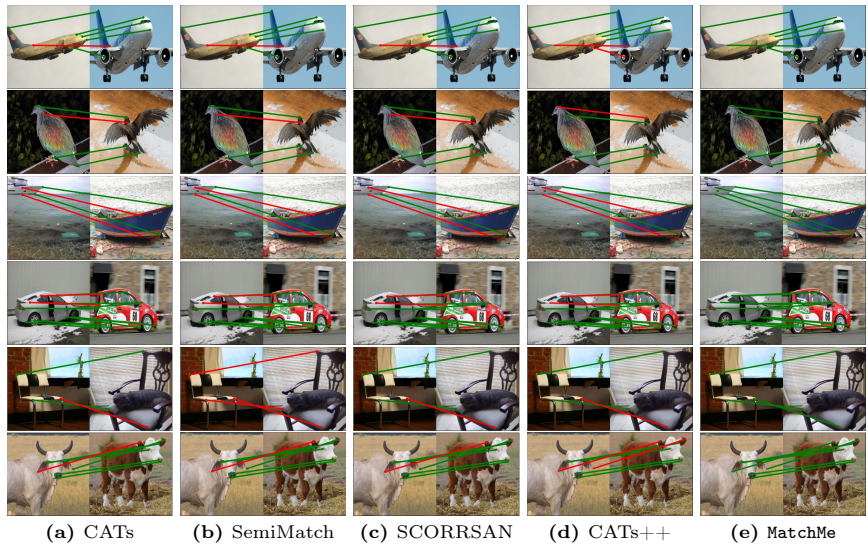

  \centering
  \footnotesize
  \begin{subfigure}[b]{0.18\textwidth}
    \centering
    \includegraphics[width=\textwidth]{appendix_figure/qual_sota/qual_sota-1.pdf}
  \end{subfigure}
  \begin{subfigure}[b]{0.18\textwidth}
    \centering
    \includegraphics[width=\textwidth]{appendix_figure/qual_sota/qual_sota-2.pdf}
  \end{subfigure}
  \begin{subfigure}[b]{0.18\textwidth}
    \centering
    \includegraphics[width=\textwidth]{appendix_figure/qual_sota/qual_sota-3.pdf}
  \end{subfigure}
  \begin{subfigure}[b]{0.18\textwidth}
    \centering
    \includegraphics[width=\textwidth]{appendix_figure/qual_sota/qual_sota-4.pdf}
  \end{subfigure}
    \begin{subfigure}[b]{0.18\textwidth}
    \centering
    \includegraphics[width=\textwidth]{appendix_figure/qual_sota/qual_sota-5.pdf}
  \end{subfigure} \\
   \begin{subfigure}[b]{0.18\textwidth}
    \centering
    \includegraphics[width=\textwidth]{appendix_figure/qual_sota/qual_sota-11.pdf}
  \end{subfigure}
  \begin{subfigure}[b]{0.18\textwidth}
    \centering
    \includegraphics[width=\textwidth]{appendix_figure/qual_sota/qual_sota-12.pdf}
  \end{subfigure}
  \begin{subfigure}[b]{0.18\textwidth}
    \centering
    \includegraphics[width=\textwidth]{appendix_figure/qual_sota/qual_sota-13.pdf}
  \end{subfigure}
  \begin{subfigure}[b]{0.18\textwidth}
    \centering
    \includegraphics[width=\textwidth]{appendix_figure/qual_sota/qual_sota-14.pdf}
  \end{subfigure}
    \begin{subfigure}[b]{0.18\textwidth}
    \centering
    \includegraphics[width=\textwidth]{appendix_figure/qual_sota/qual_sota-15.pdf}
  \end{subfigure} \\
    \begin{subfigure}[b]{0.18\textwidth}
    \centering
    \includegraphics[width=\textwidth]{appendix_figure/qual_sota/qual_sota-16.pdf}
  \end{subfigure}
  \begin{subfigure}[b]{0.18\textwidth}
    \centering
    \includegraphics[width=\textwidth]{appendix_figure/qual_sota/qual_sota-17.pdf}
  \end{subfigure}
  \begin{subfigure}[b]{0.18\textwidth}
    \centering
    \includegraphics[width=\textwidth]{appendix_figure/qual_sota/qual_sota-18.pdf}
  \end{subfigure}
  \begin{subfigure}[b]{0.18\textwidth}
    \centering
    \includegraphics[width=\textwidth]{appendix_figure/qual_sota/qual_sota-19.pdf}
  \end{subfigure}
    \begin{subfigure}[b]{0.18\textwidth}
    \centering
    \includegraphics[width=\textwidth]{appendix_figure/qual_sota/qual_sota-20.pdf}
  \end{subfigure} \\
     \begin{subfigure}[b]{0.18\textwidth}
    \centering
    \includegraphics[width=\textwidth]{appendix_figure/qual_sota/qual_sota-26.pdf}
  \end{subfigure}
  \begin{subfigure}[b]{0.18\textwidth}
    \centering
    \includegraphics[width=\textwidth]{appendix_figure/qual_sota/qual_sota-27.pdf}
  \end{subfigure}
  \begin{subfigure}[b]{0.18\textwidth}
    \centering
    \includegraphics[width=\textwidth]{appendix_figure/qual_sota/qual_sota-28.pdf}
  \end{subfigure}
  \begin{subfigure}[b]{0.18\textwidth}
    \centering
    \includegraphics[width=\textwidth]{appendix_figure/qual_sota/qual_sota-29.pdf}
  \end{subfigure}
    \begin{subfigure}[b]{0.18\textwidth}
    \centering
    \includegraphics[width=\textwidth]{appendix_figure/qual_sota/qual_sota-30.pdf}
  \end{subfigure} \\
      \begin{subfigure}[b]{0.18\textwidth}
    \centering
    \includegraphics[width=\textwidth]{appendix_figure/qual_sota/qual_sota-31.pdf}
  \end{subfigure}
  \begin{subfigure}[b]{0.18\textwidth}
    \centering
    \includegraphics[width=\textwidth]{appendix_figure/qual_sota/qual_sota-32.pdf}
  \end{subfigure}
  \begin{subfigure}[b]{0.18\textwidth}
    \centering
    \includegraphics[width=\textwidth]{appendix_figure/qual_sota/qual_sota-33.pdf}
  \end{subfigure}
  \begin{subfigure}[b]{0.18\textwidth}
    \centering
    \includegraphics[width=\textwidth]{appendix_figure/qual_sota/qual_sota-34.pdf}
  \end{subfigure}
    \begin{subfigure}[b]{0.18\textwidth}
    \centering
    \includegraphics[width=\textwidth]{appendix_figure/qual_sota/qual_sota-35.pdf}
  \end{subfigure} 
      \begin{subfigure}[b]{0.18\textwidth}
    \centering
    \includegraphics[width=\textwidth]{appendix_figure/qual_sota/qual_sota-36.pdf}
    \caption{CATs}
  \end{subfigure}
  \begin{subfigure}[b]{0.18\textwidth}
    \centering
    \includegraphics[width=\textwidth]{appendix_figure/qual_sota/qual_sota-37.pdf}
   \caption{SemiMatch}
  \end{subfigure}
  \begin{subfigure}[b]{0.18\textwidth}
    \centering
    \includegraphics[width=\textwidth]{appendix_figure/qual_sota/qual_sota-38.pdf}
   \caption{SCORRSAN}
  \end{subfigure}
  \begin{subfigure}[b]{0.18\textwidth}
    \centering
    \includegraphics[width=\textwidth]{appendix_figure/qual_sota/qual_sota-39.pdf}
   \caption{CATs++~}
  \end{subfigure}
    \begin{subfigure}[b]{0.18\textwidth}
    \centering
    \includegraphics[width=\textwidth]{appendix_figure/qual_sota/qual_sota-40.pdf}
   \caption{\ours}
  \end{subfigure} \\
  \caption{\textbf{Qualitative results on SPair-71k in comparison with SOTA methods (cont'd) } The point-to-point matches are drawn by linking key point pairs with line segments. {Green} and {red} lines denote correct and incorrect predictions with respect to the ground-truth pairs, respectively. We observe that ours performs much better compared with the counterparts across all the sample image pairs.}
  \label{fig:qual_sota_appendix}
\end{figure*}

\begin{figure*}[t!]
     \begin{subfigure}[b]{0.2\textwidth}
    \centering
    \includegraphics[width=\textwidth]{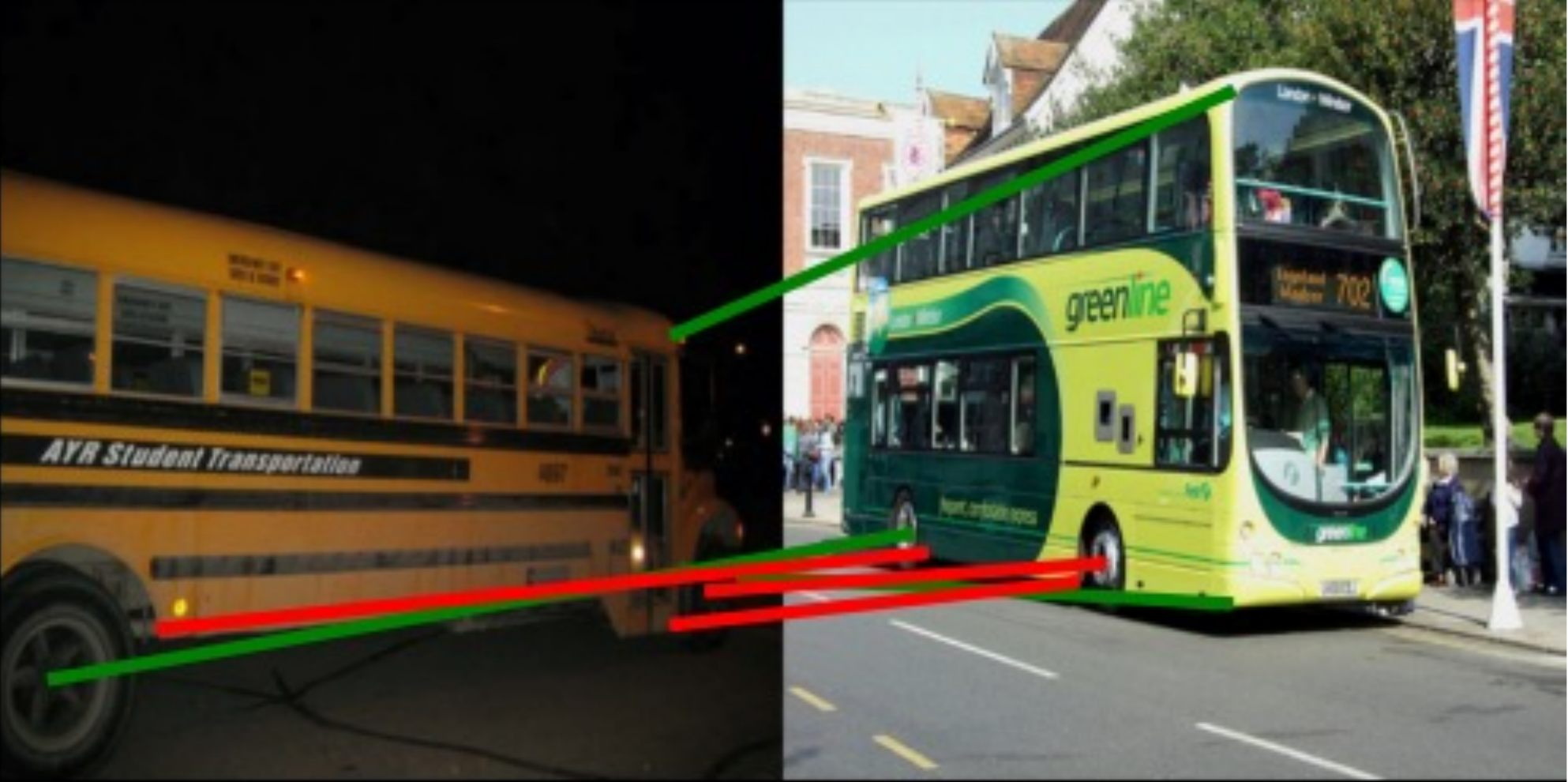}
  \end{subfigure}\hfill
  \begin{subfigure}[b]{0.2\textwidth}
    \centering
    \includegraphics[width=\textwidth]{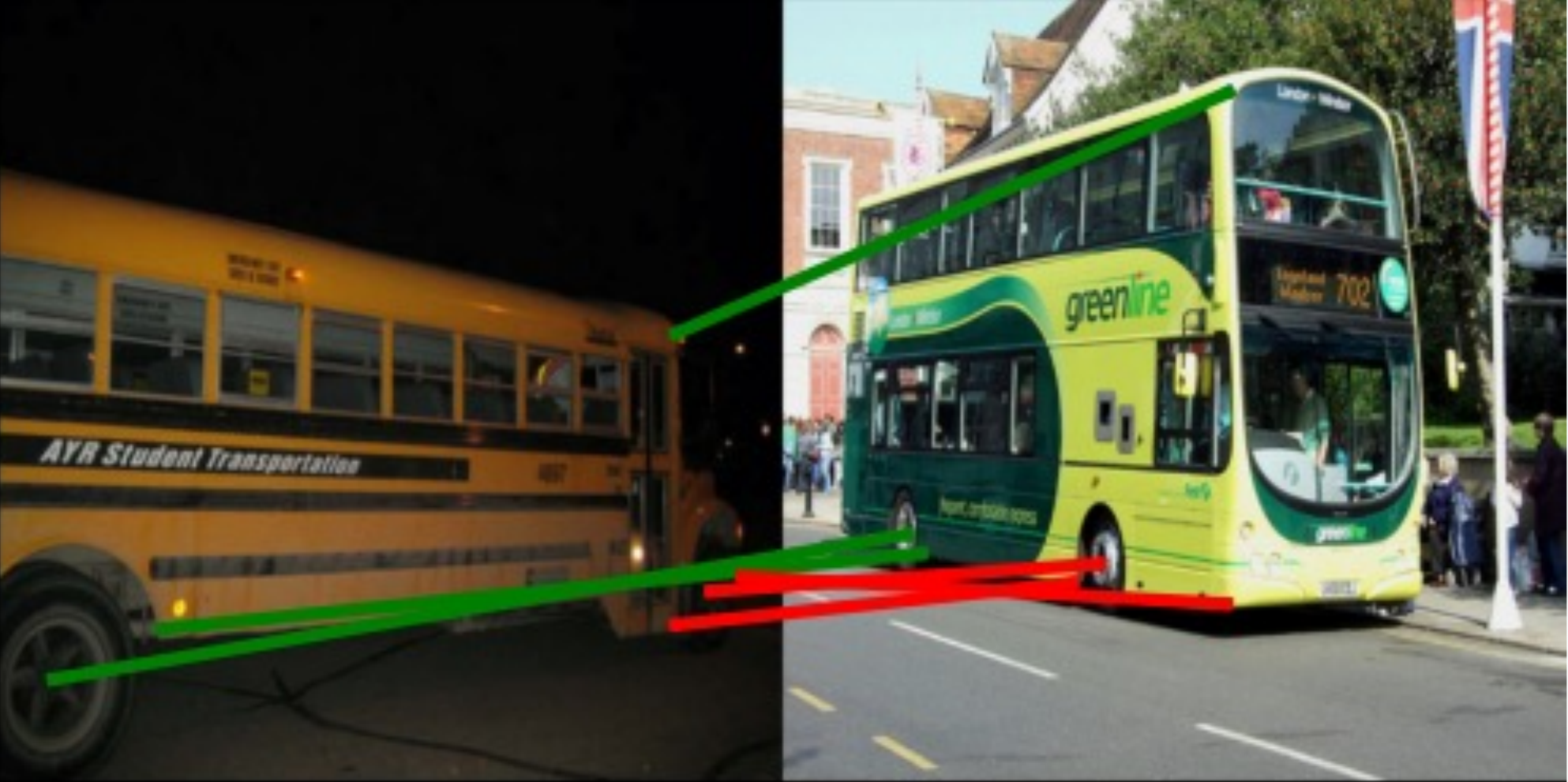}
  \end{subfigure}\hfill
  \begin{subfigure}[b]{0.2\textwidth}
    \centering
    \includegraphics[width=\textwidth]{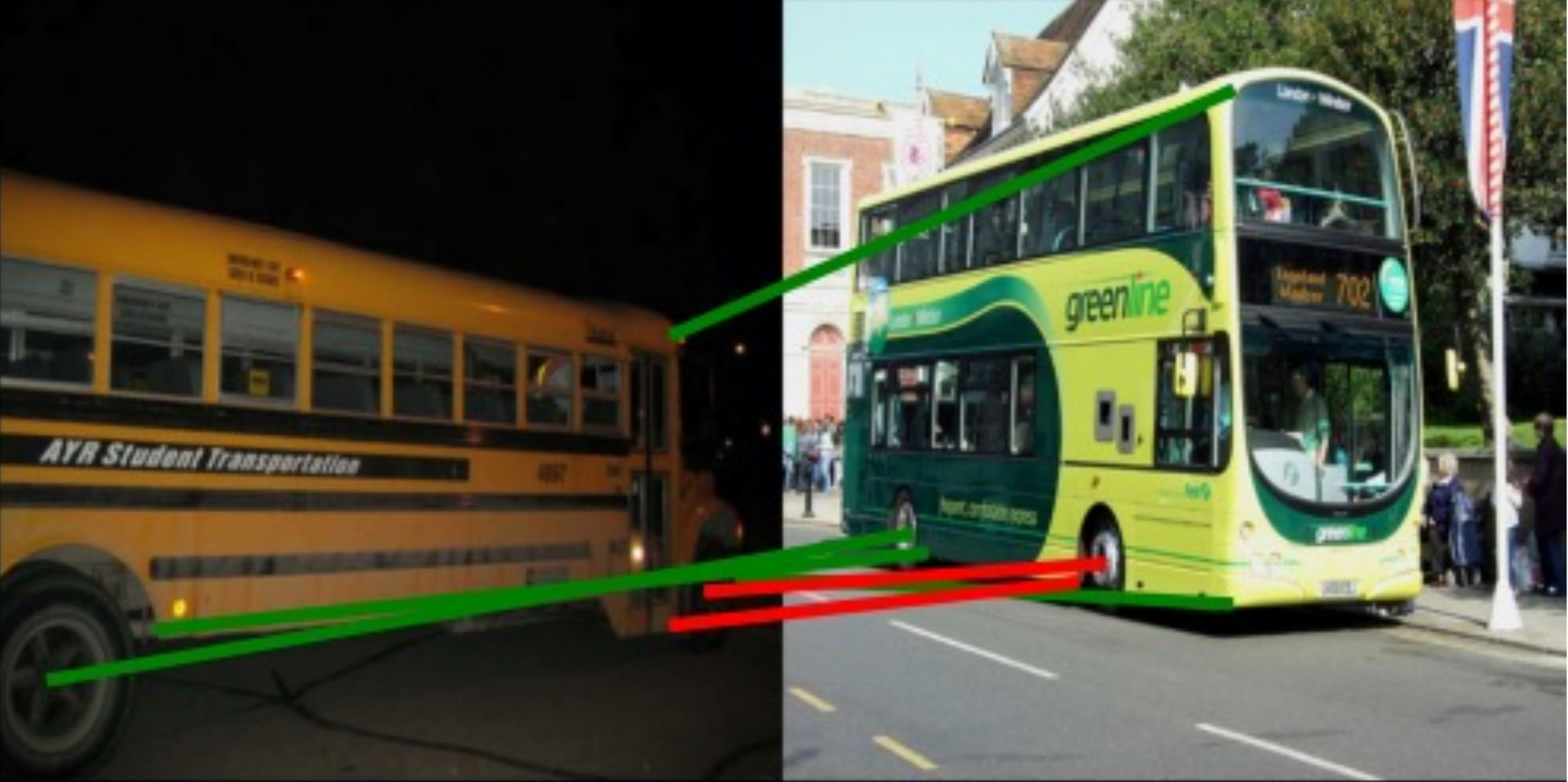}
  \end{subfigure}\hfill
  \begin{subfigure}[b]{0.2\textwidth}
    \centering
    \includegraphics[width=\textwidth]{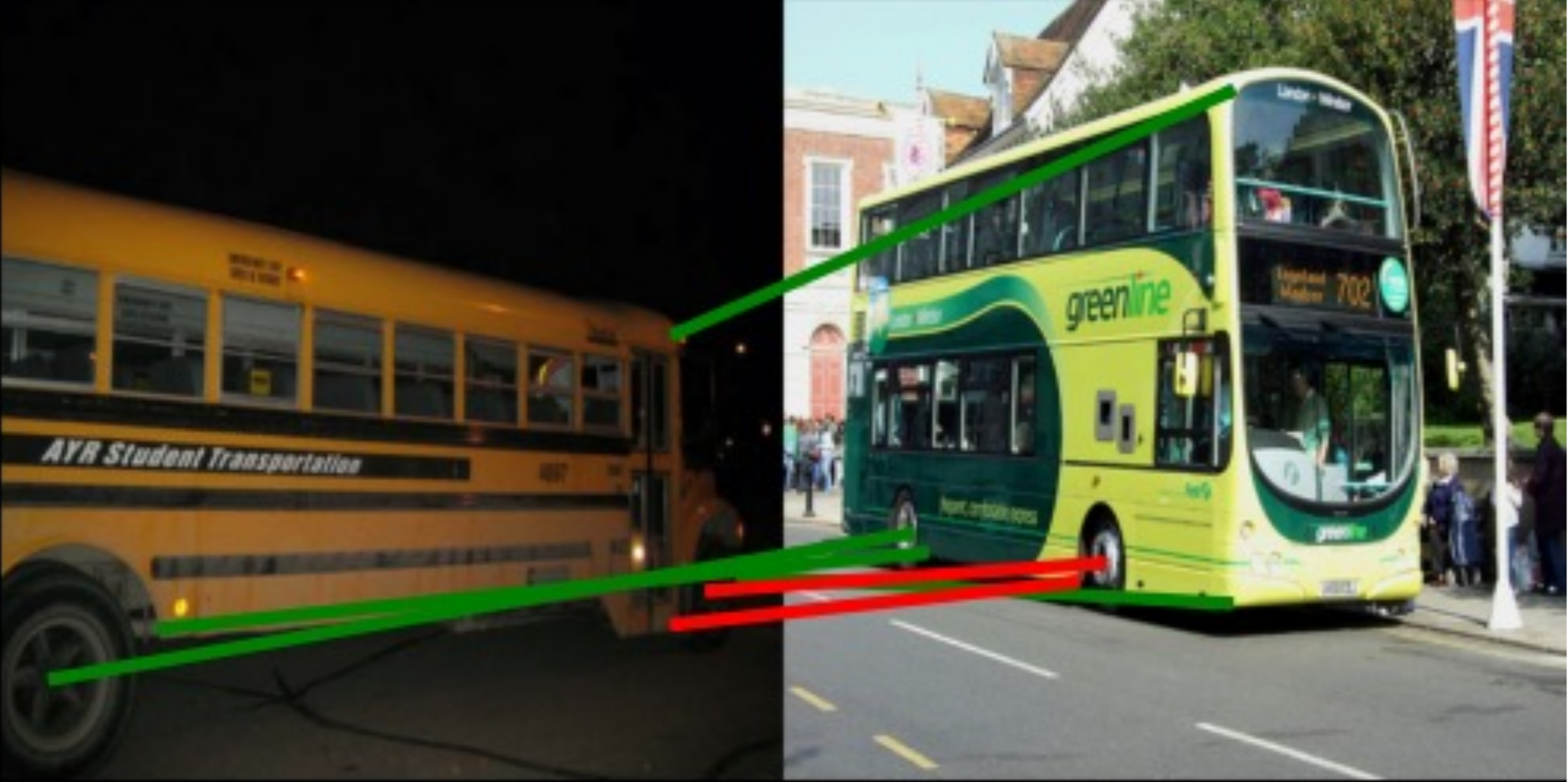}
  \end{subfigure}\hfill
    \begin{subfigure}[b]{0.2\textwidth}
    \centering
    \includegraphics[width=\textwidth]{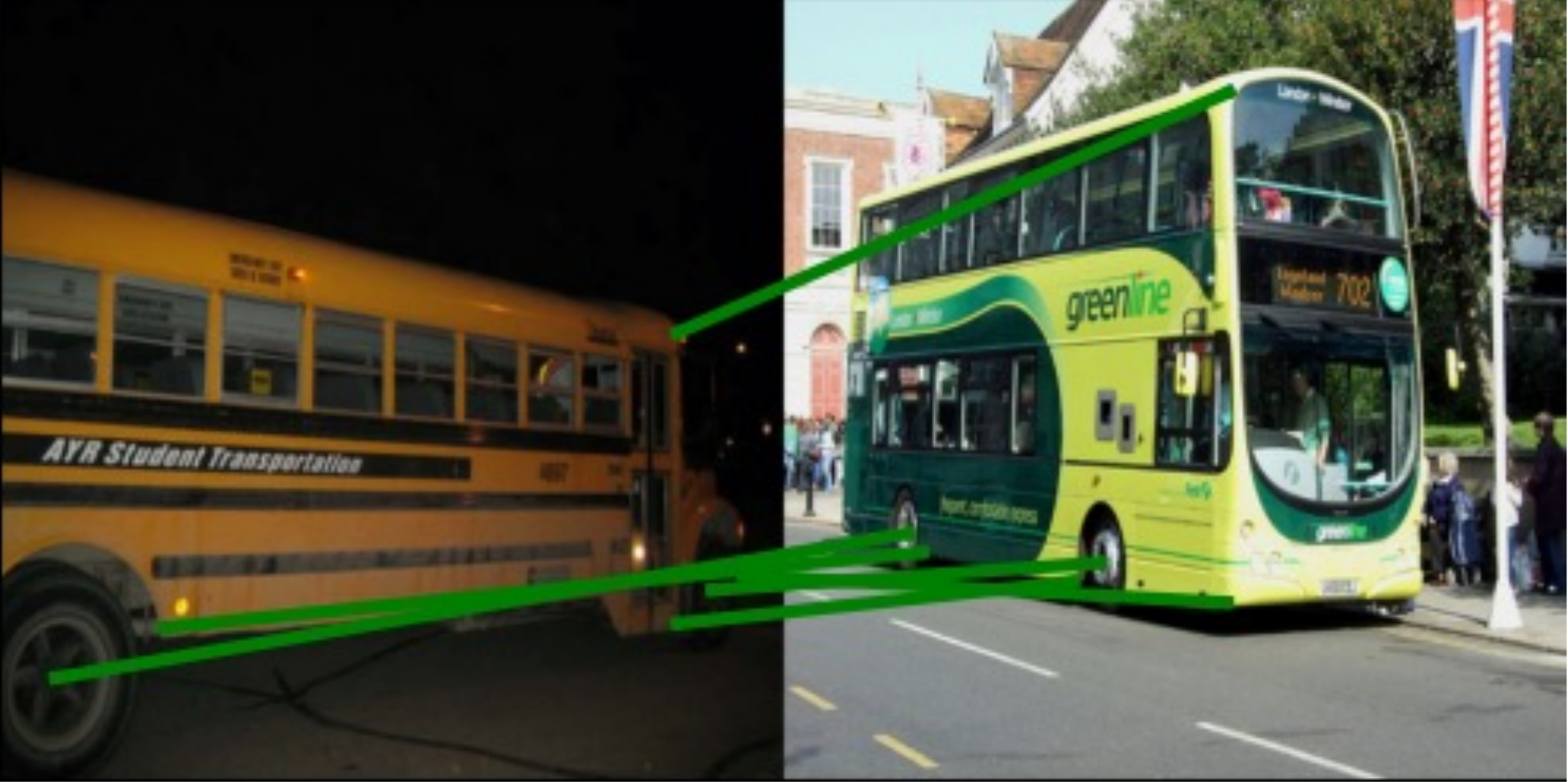}
  \end{subfigure}\hfill \\
      \begin{subfigure}[b]{0.2\textwidth}
    \centering
    \includegraphics[width=\textwidth]{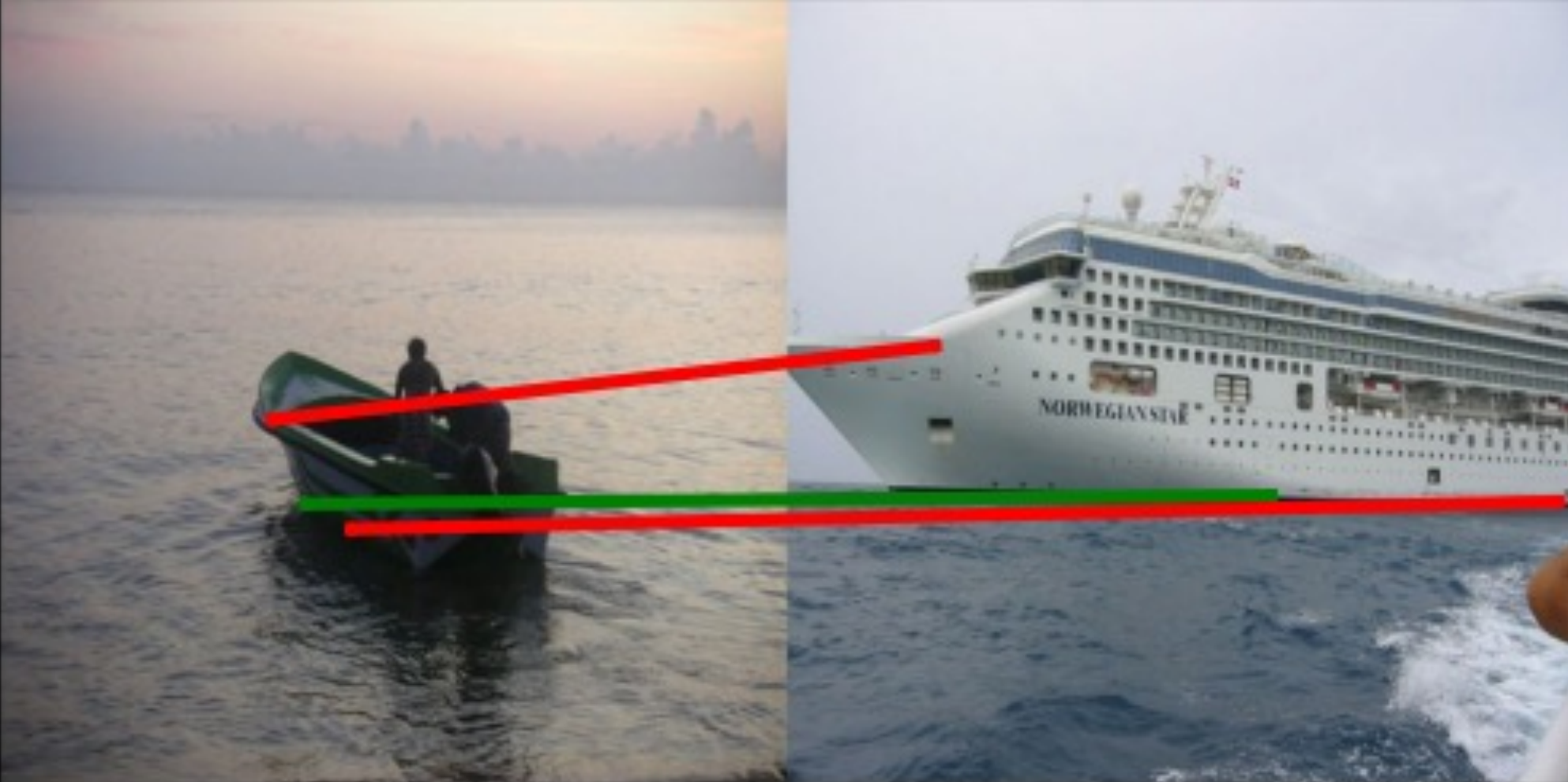}
  \end{subfigure}\hfill
  \begin{subfigure}[b]{0.2\textwidth}
    \centering
    \includegraphics[width=\textwidth]{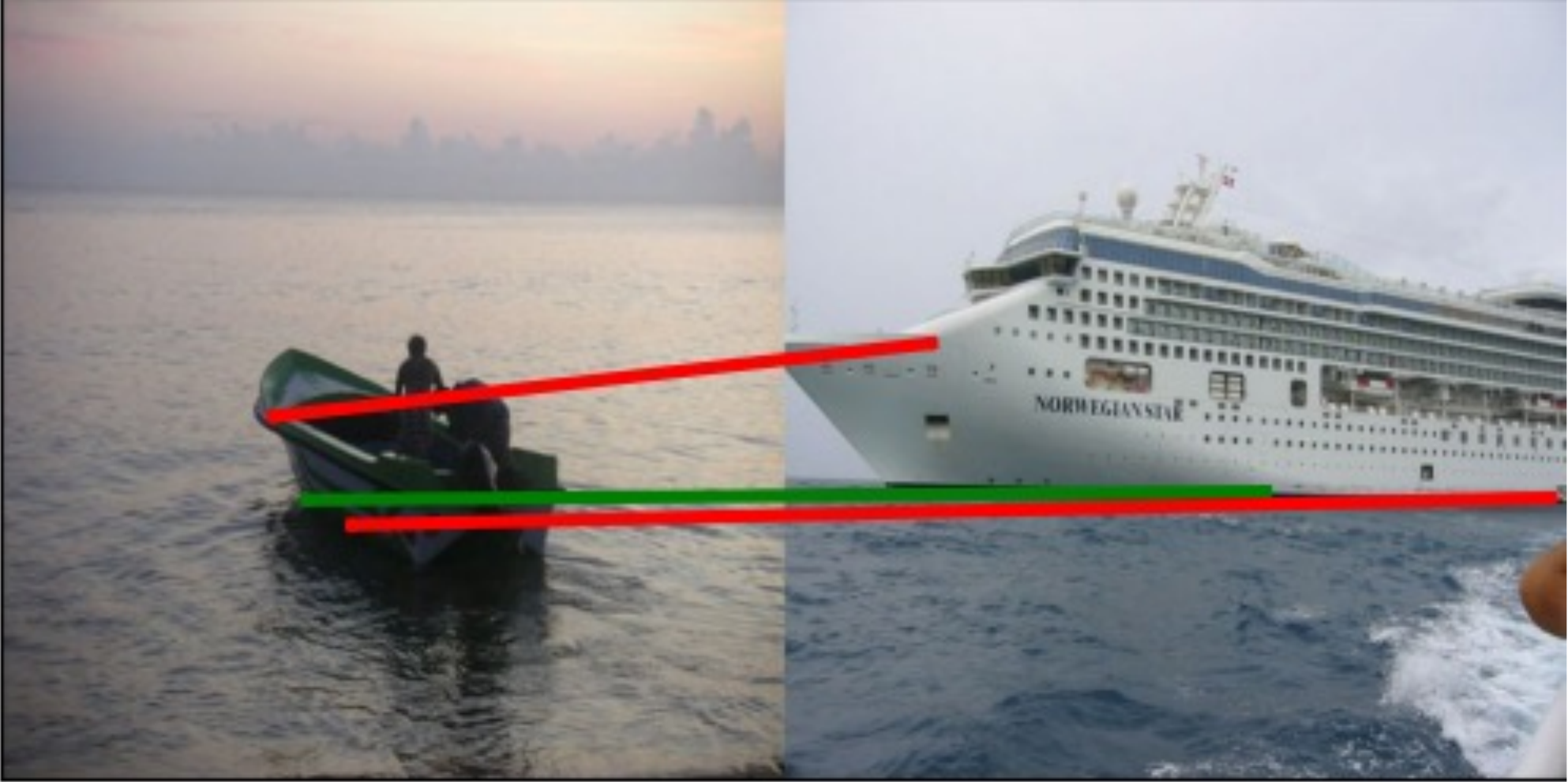}
  \end{subfigure}\hfill
  \begin{subfigure}[b]{0.2\textwidth}
    \centering
    \includegraphics[width=\textwidth]{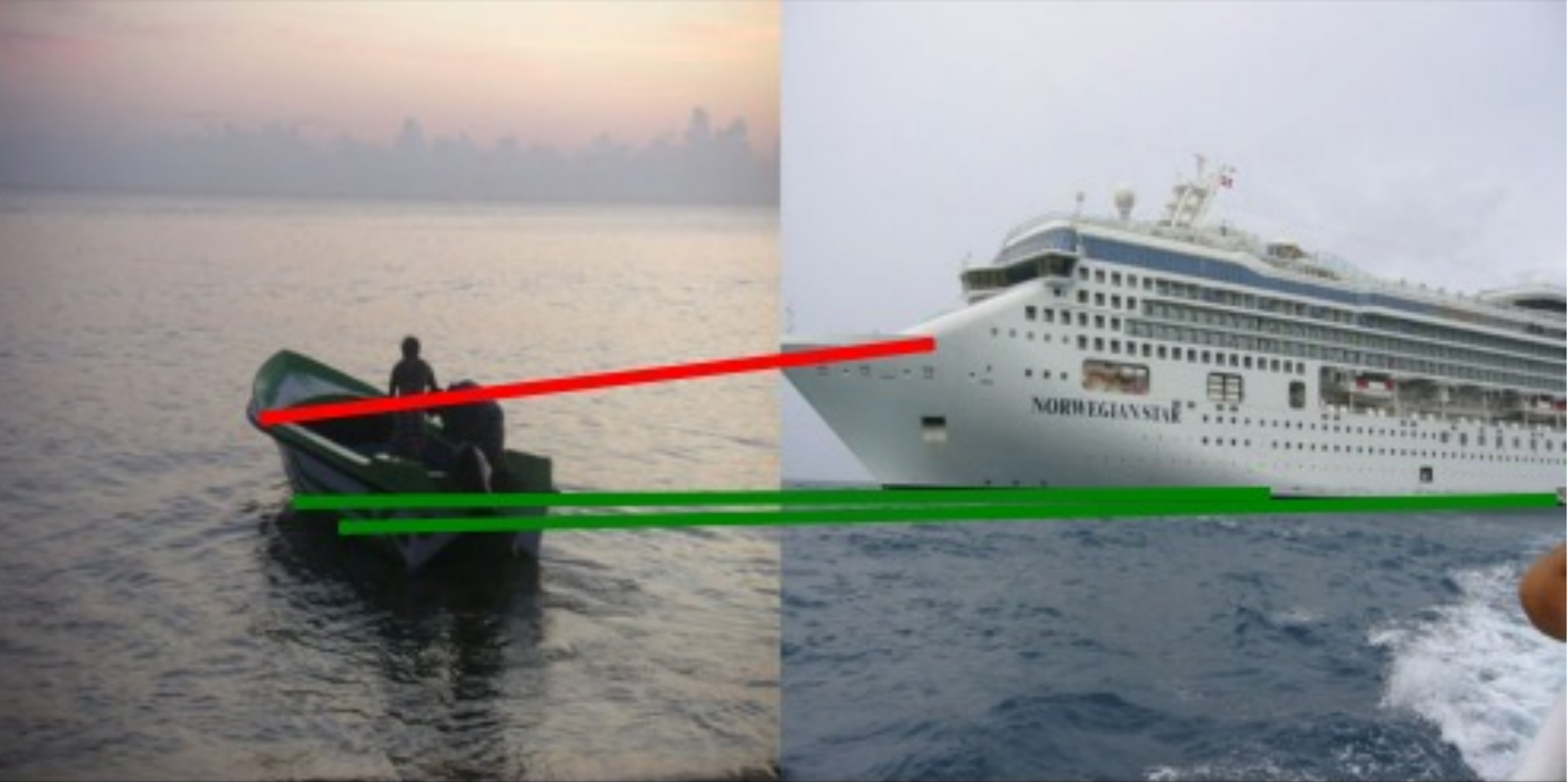}
  \end{subfigure}\hfill
  \begin{subfigure}[b]{0.2\textwidth}
    \centering
    \includegraphics[width=\textwidth]{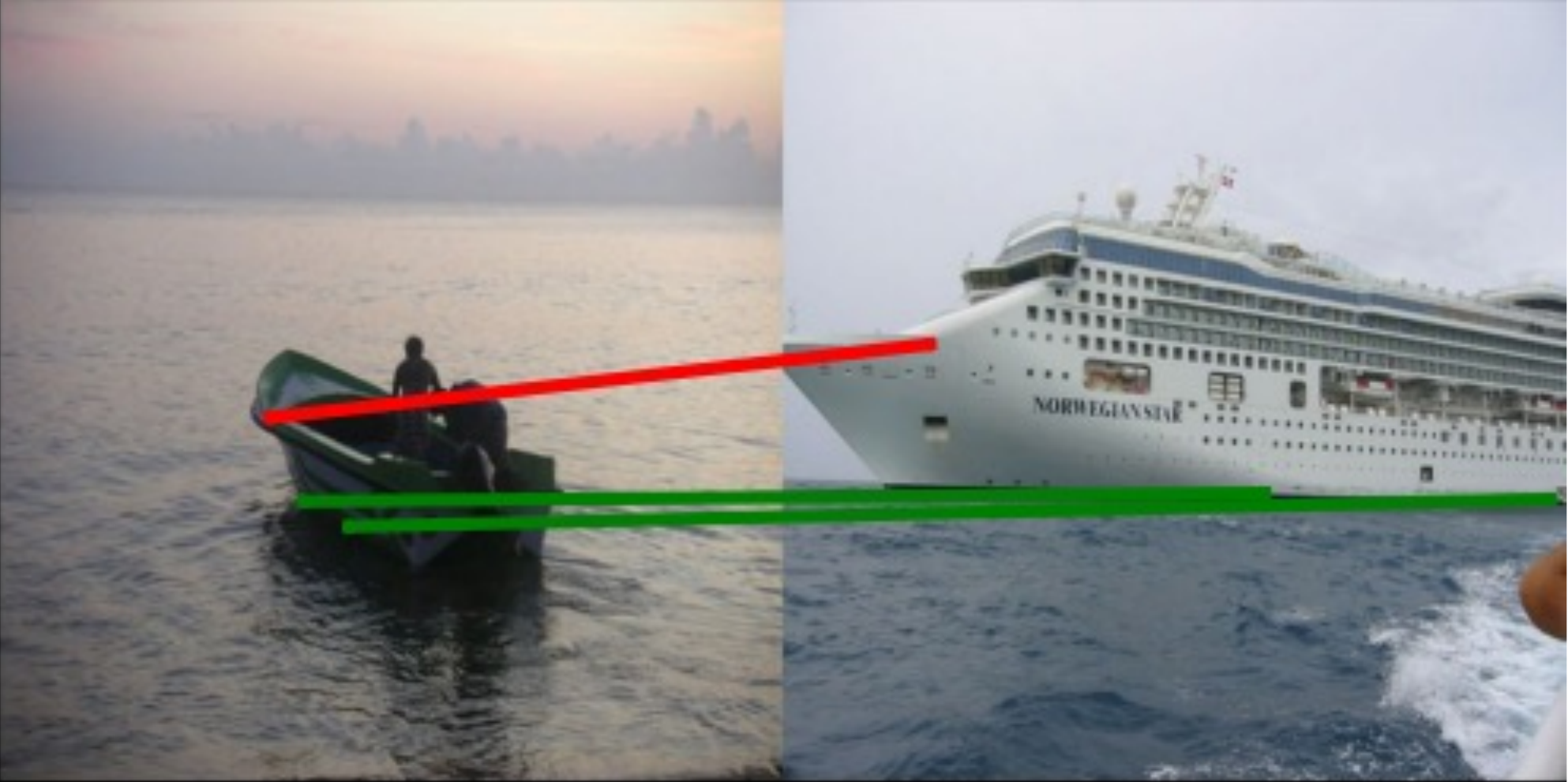}
  \end{subfigure}\hfill
    \begin{subfigure}[b]{0.2\textwidth}
    \centering
    \includegraphics[width=\textwidth]{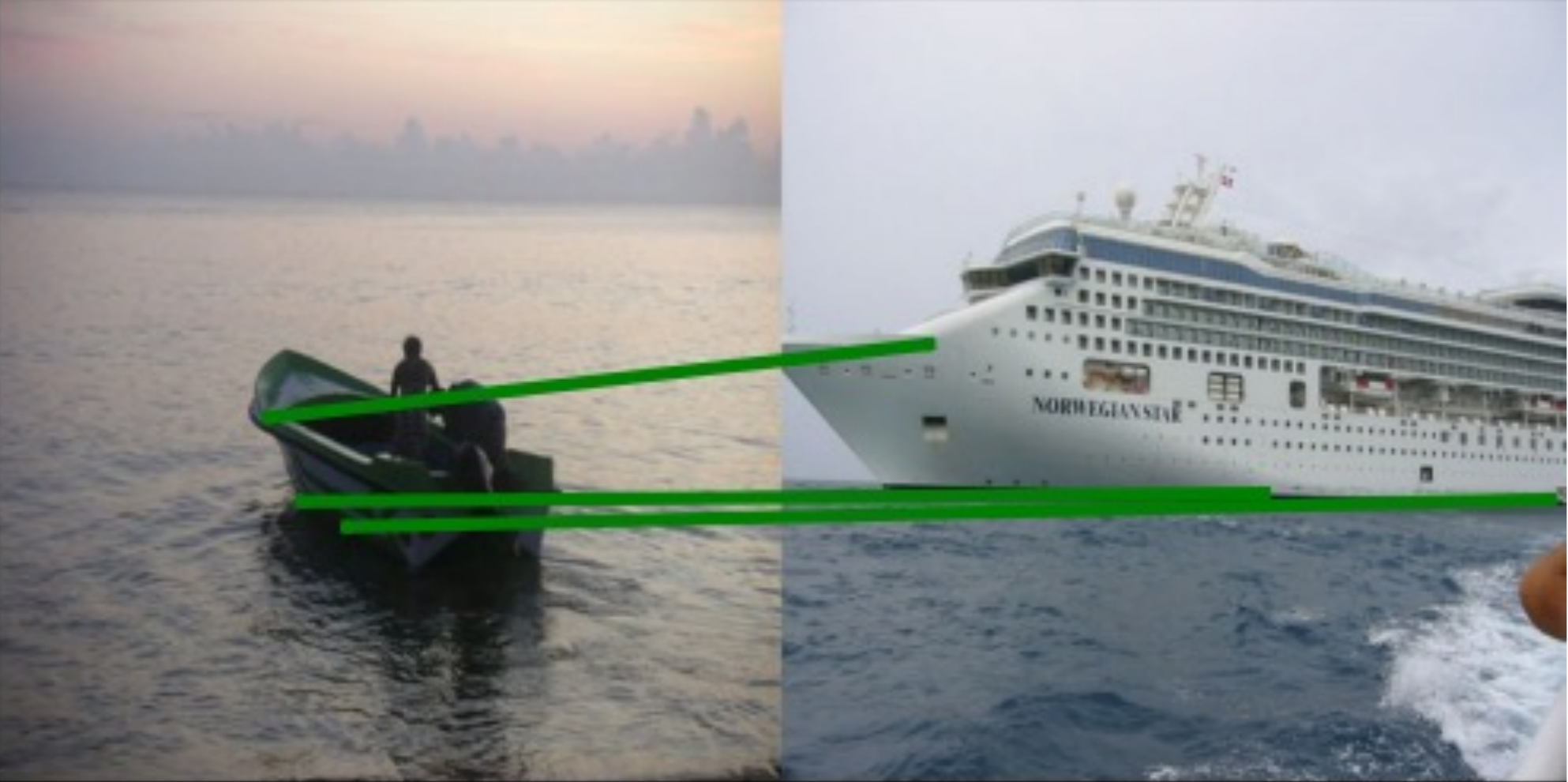}
  \end{subfigure}\hfill \\
      \begin{subfigure}[b]{0.2\textwidth}
    \centering
    \includegraphics[width=\textwidth]{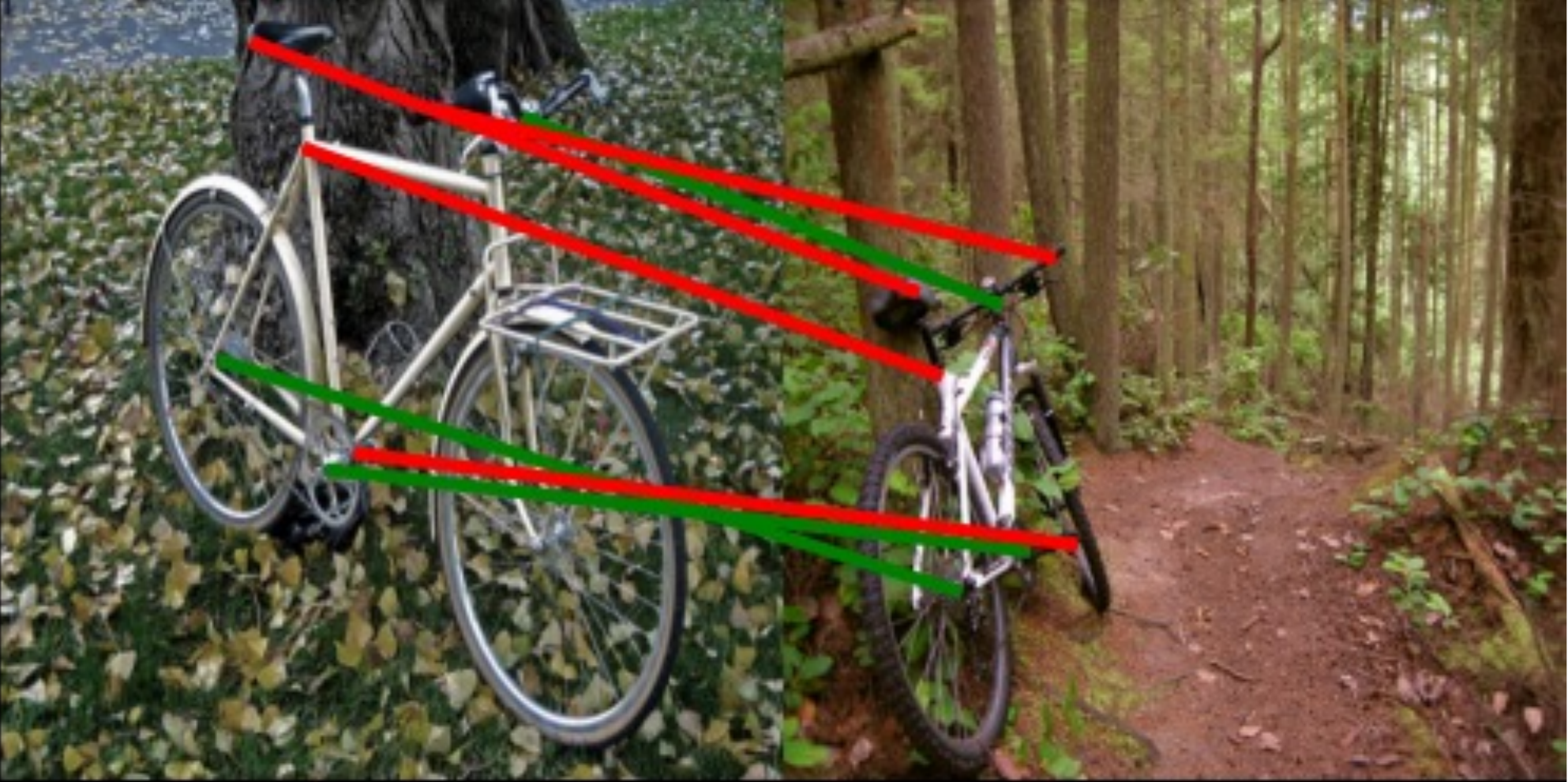}
  \end{subfigure}\hfill
  \begin{subfigure}[b]{0.2\textwidth}
    \centering
    \includegraphics[width=\textwidth]{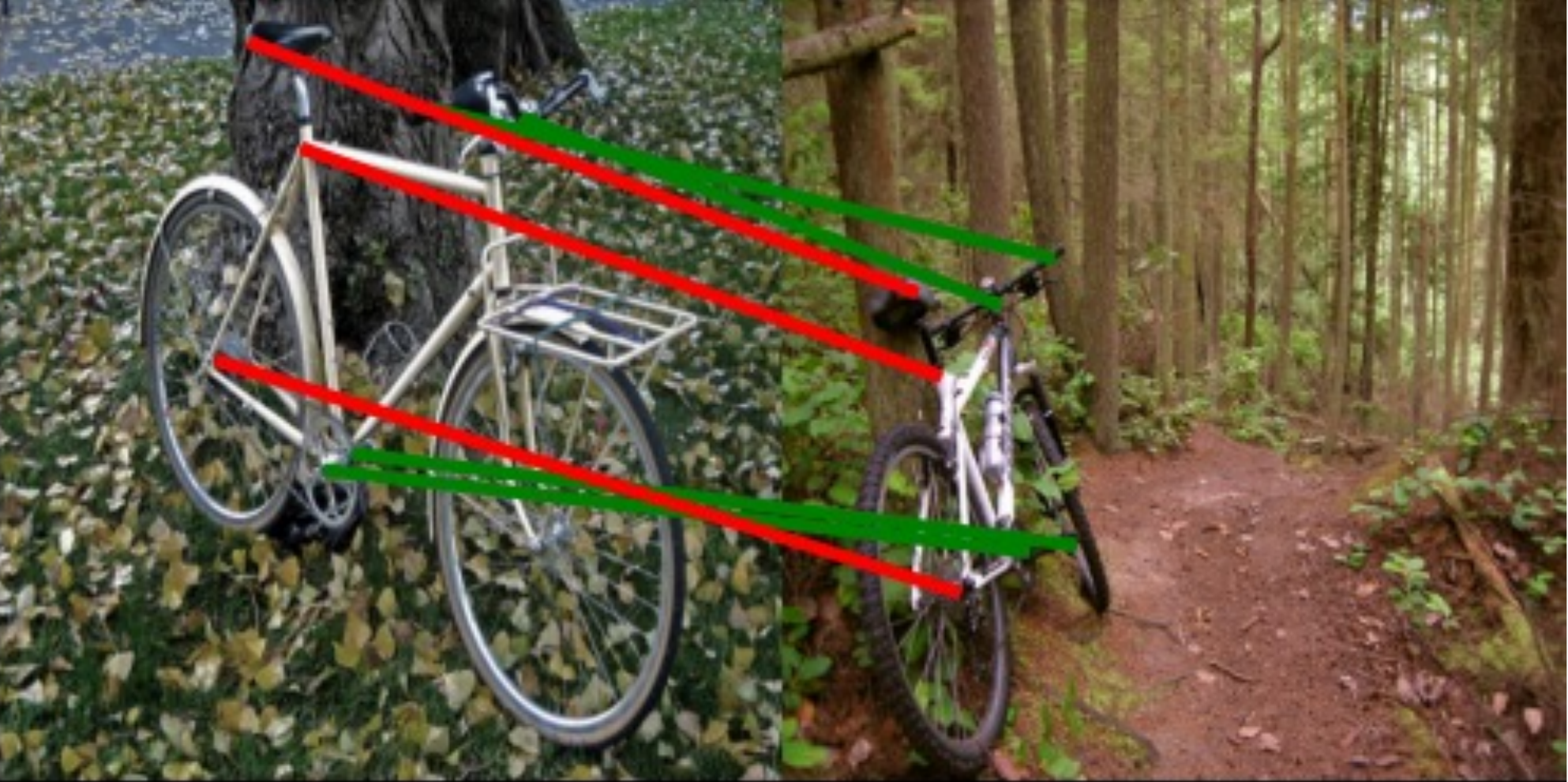}
  \end{subfigure}\hfill
  \begin{subfigure}[b]{0.2\textwidth}
    \centering
    \includegraphics[width=\textwidth]{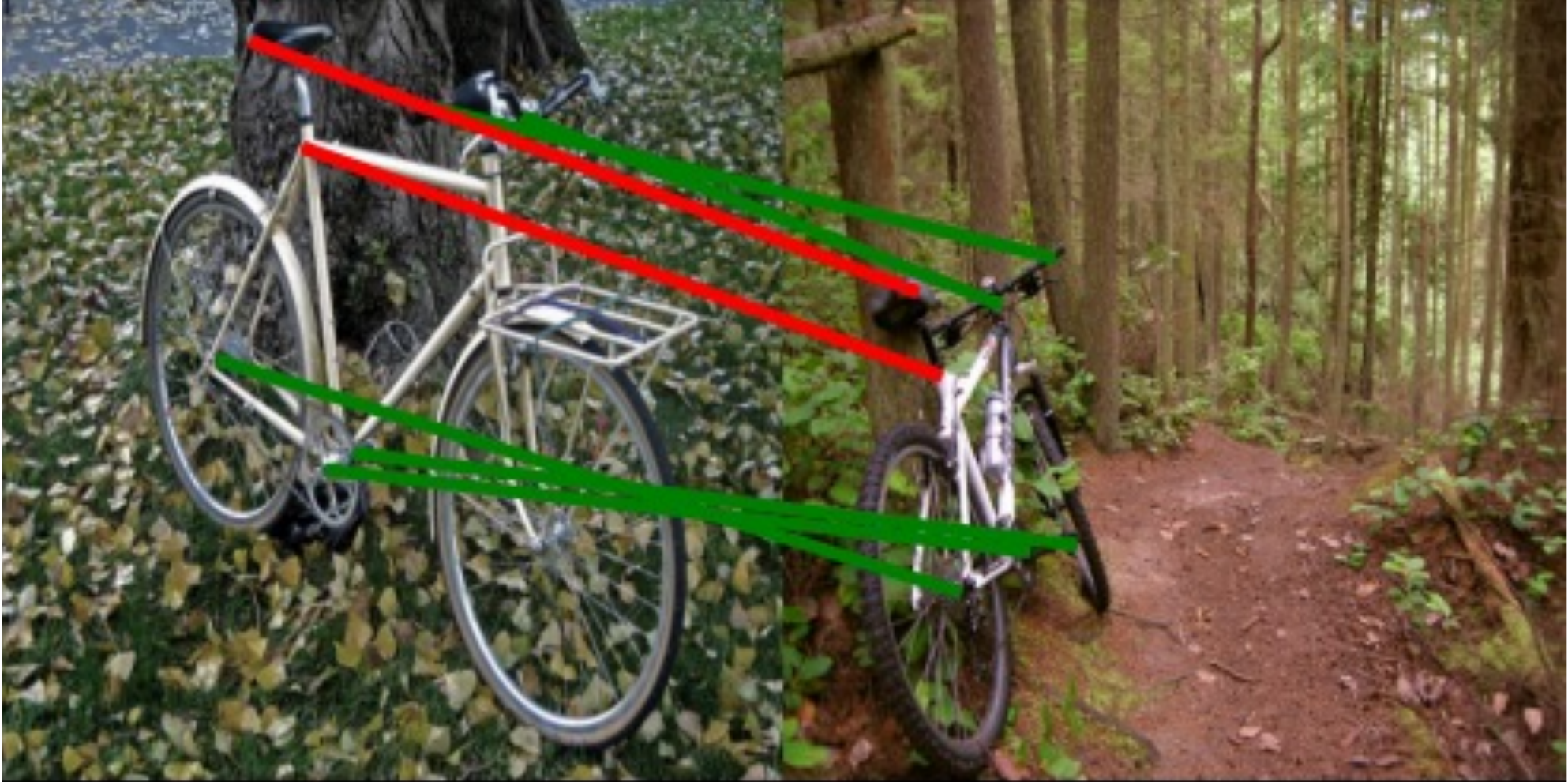}
  \end{subfigure}\hfill
  \begin{subfigure}[b]{0.2\textwidth}
    \centering
    \includegraphics[width=\textwidth]{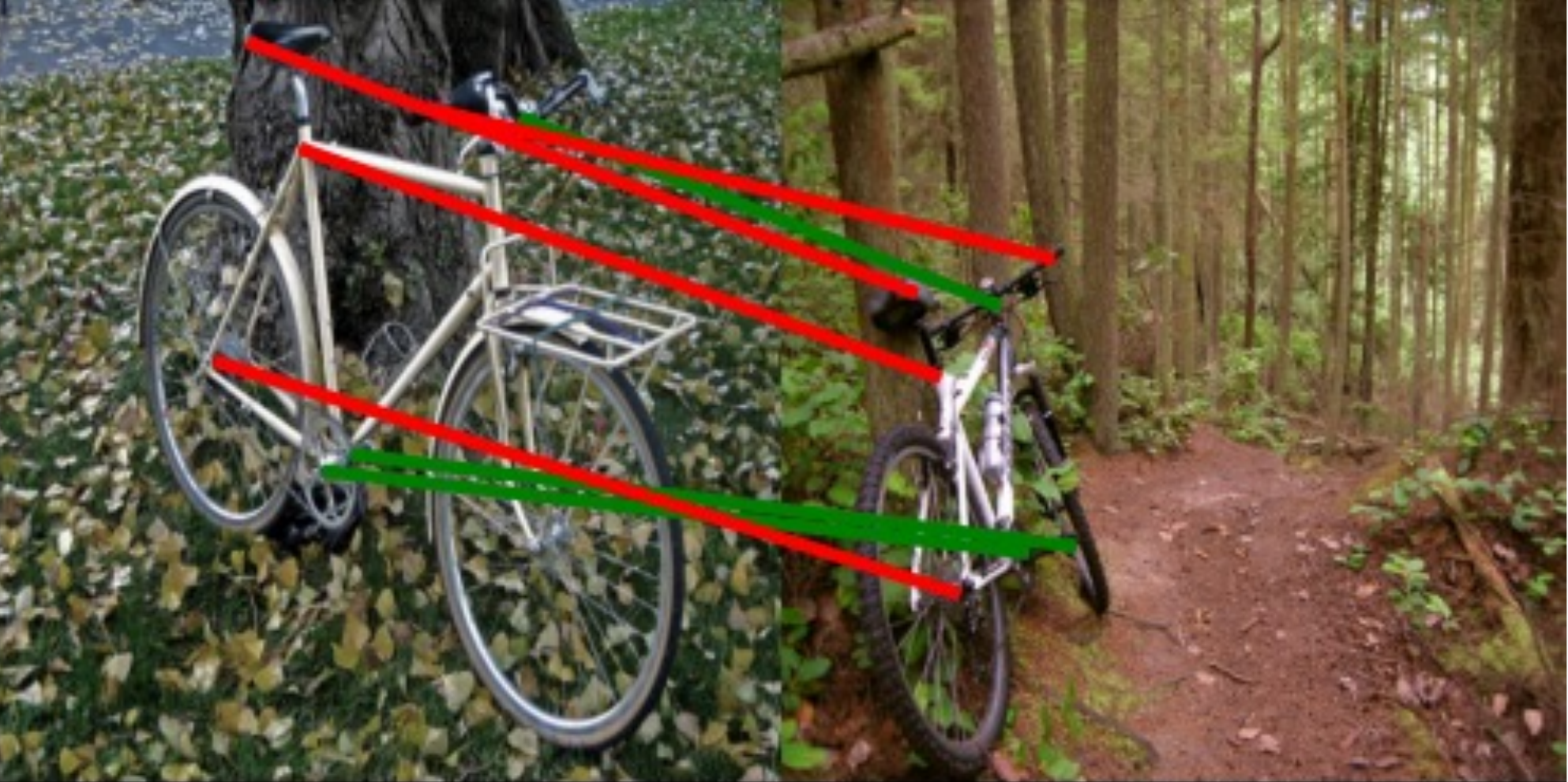}
  \end{subfigure}\hfill
    \begin{subfigure}[b]{0.2\textwidth}
    \centering
    \includegraphics[width=\textwidth]{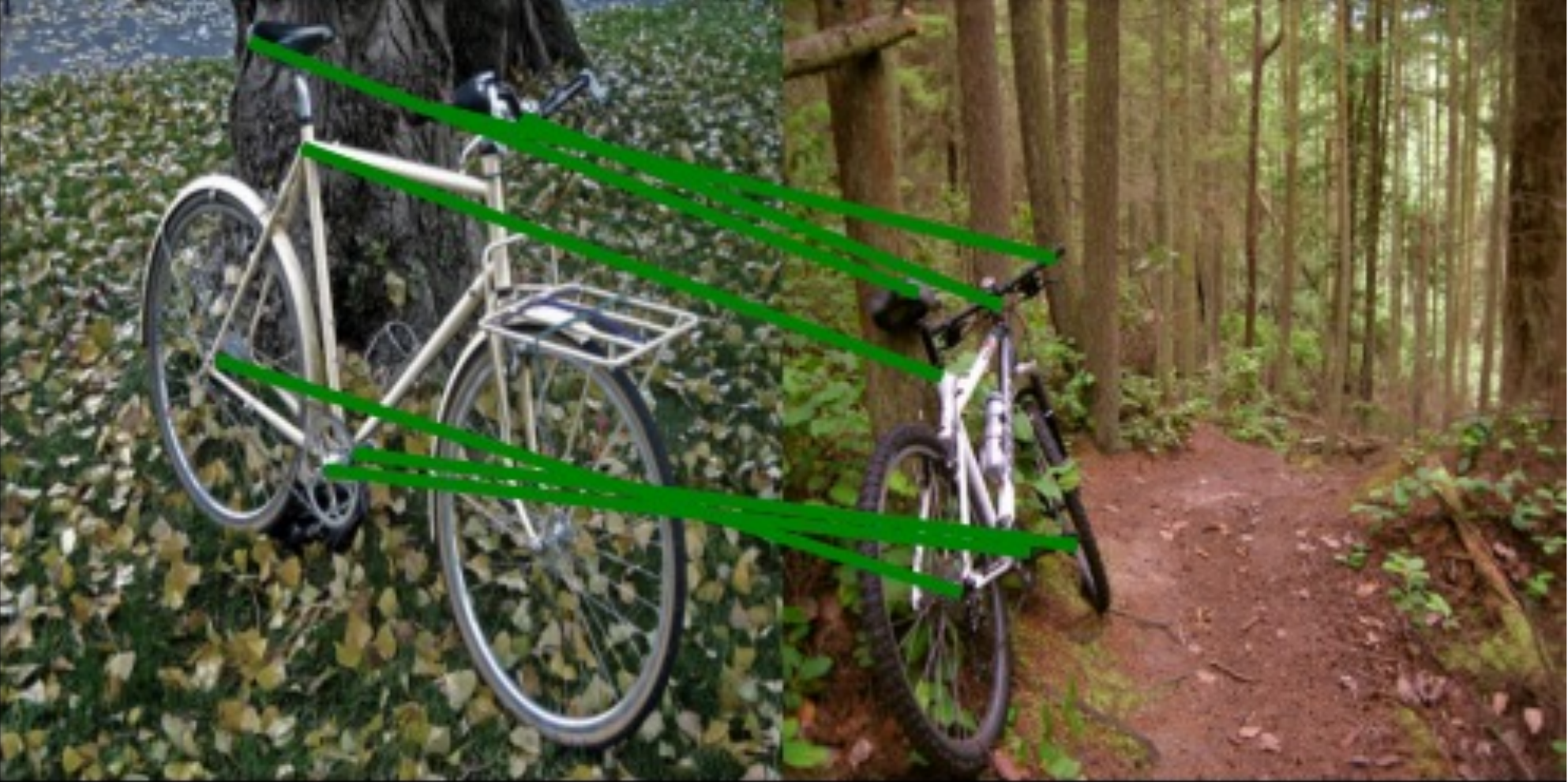}
  \end{subfigure}\hfill \\
  \begin{subfigure}[b]{0.2\textwidth}
    \centering
    \includegraphics[width=\textwidth]{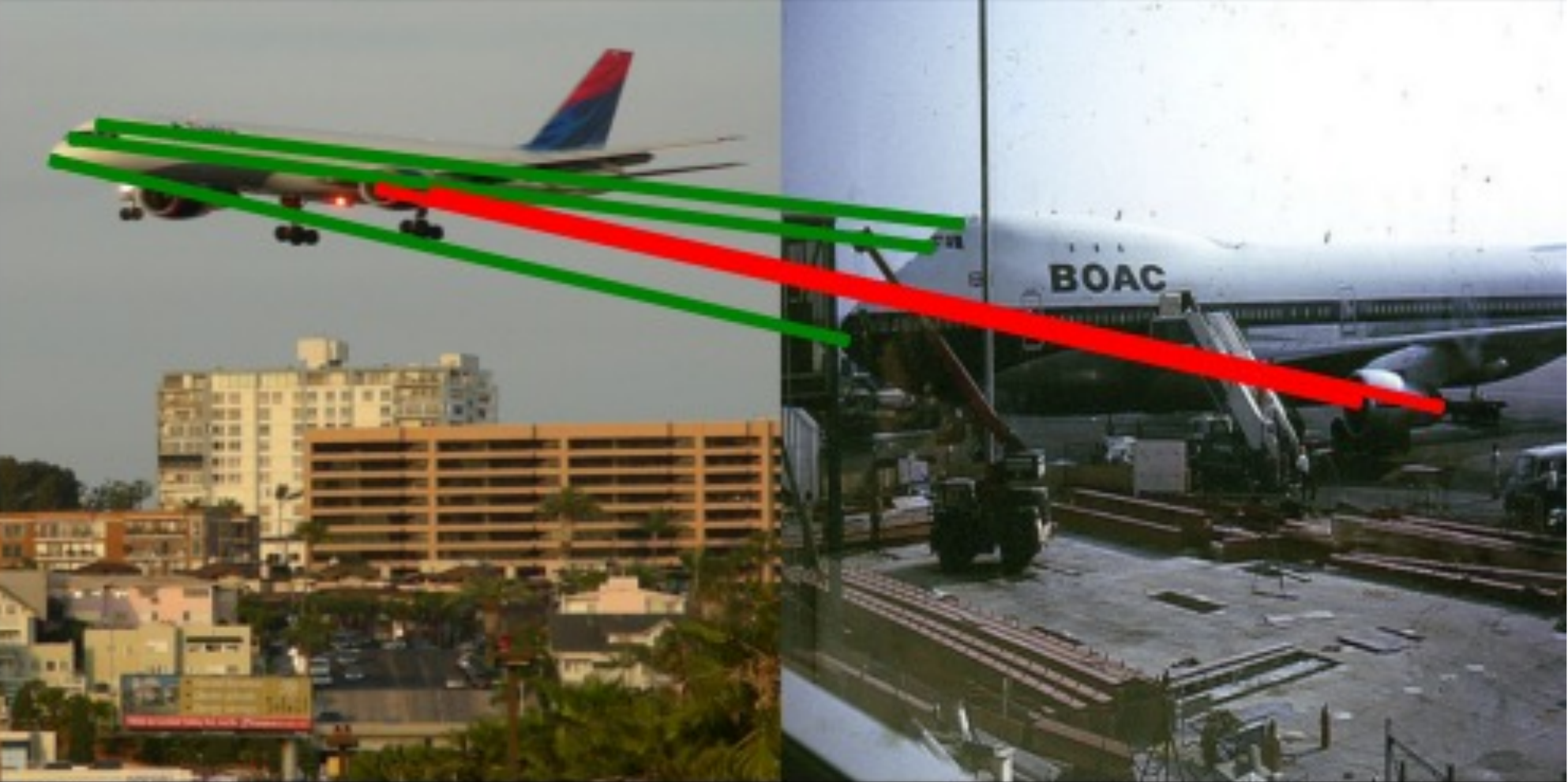}
    \caption{CATs++ (Baseline) }
  \end{subfigure}\hfill
  \begin{subfigure}[b]{0.2\textwidth}
    \centering
    \includegraphics[width=\textwidth]{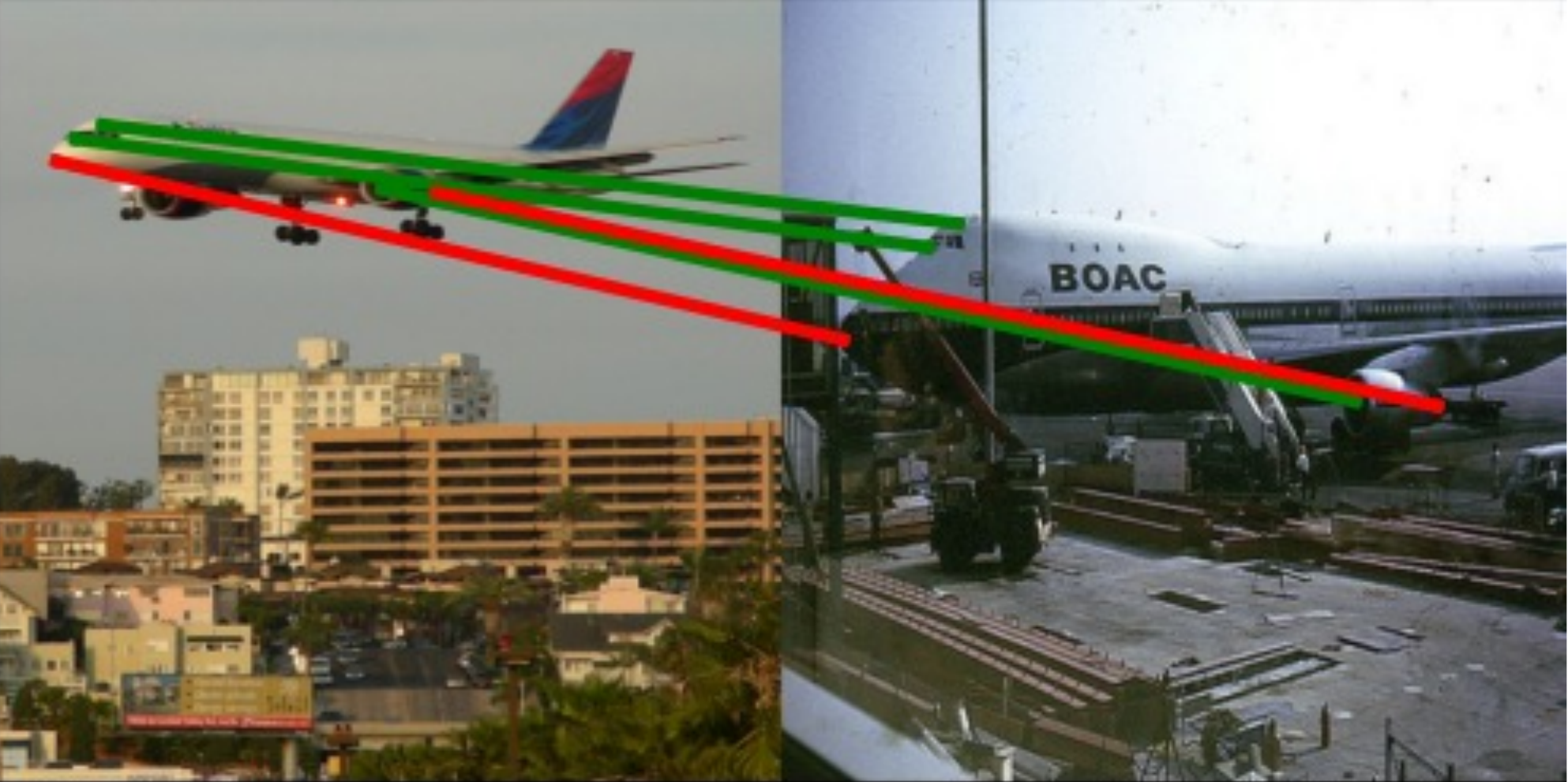}
   \caption{(a) + CNNGeoU}
  \end{subfigure}\hfill
  \begin{subfigure}[b]{0.2\textwidth}
    \centering
    \includegraphics[width=\textwidth]{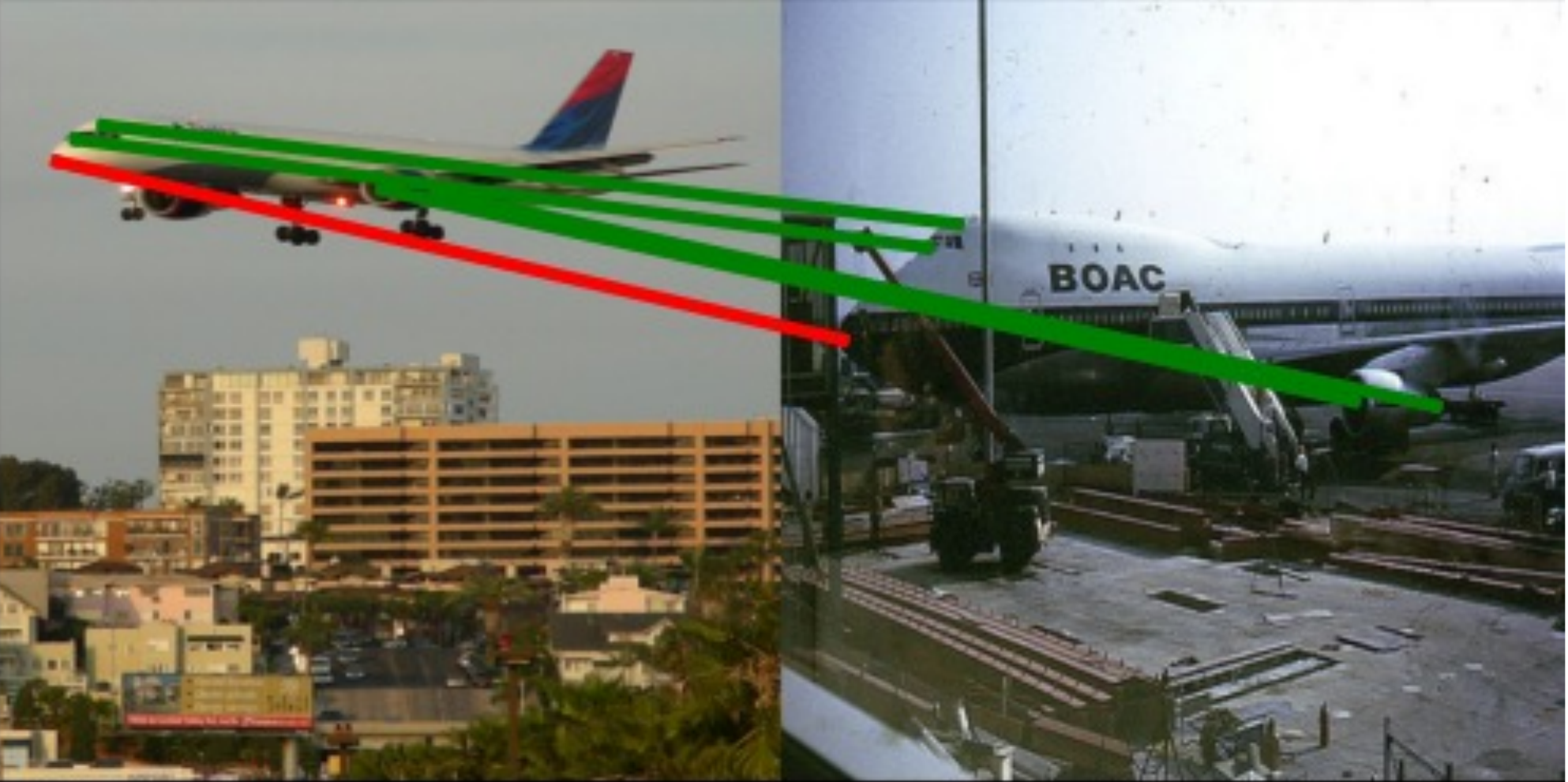}
   \caption{(a) + \\ PWarpC}
  \end{subfigure}\hfill
  \begin{subfigure}[b]{0.2\textwidth}
    \centering
    \includegraphics[width=\textwidth]{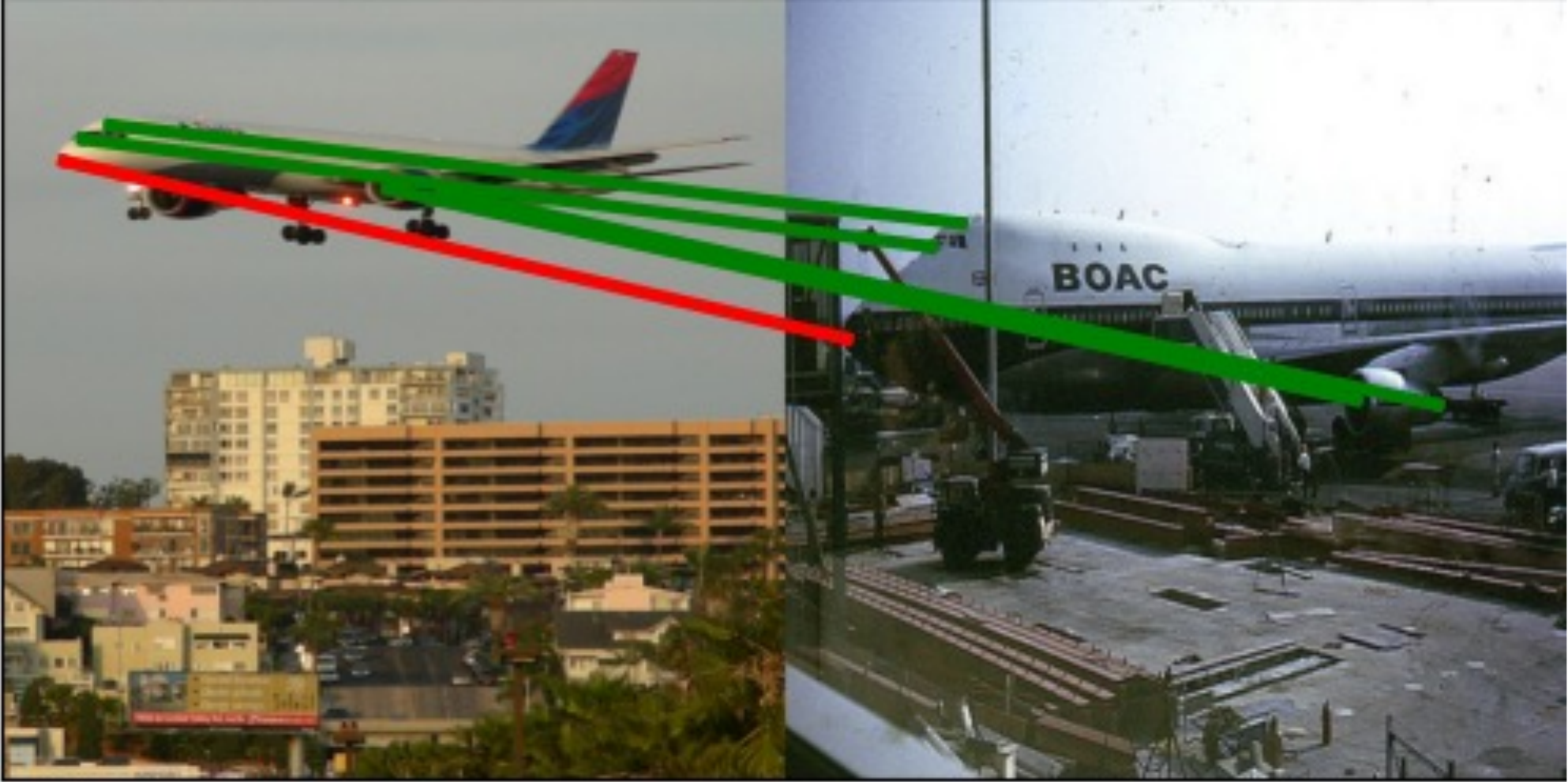}
   \caption{(a) + SCORRSAN}

  \end{subfigure}\hfill
    \begin{subfigure}[b]{0.2\textwidth}
    \centering
    \includegraphics[width=\textwidth]{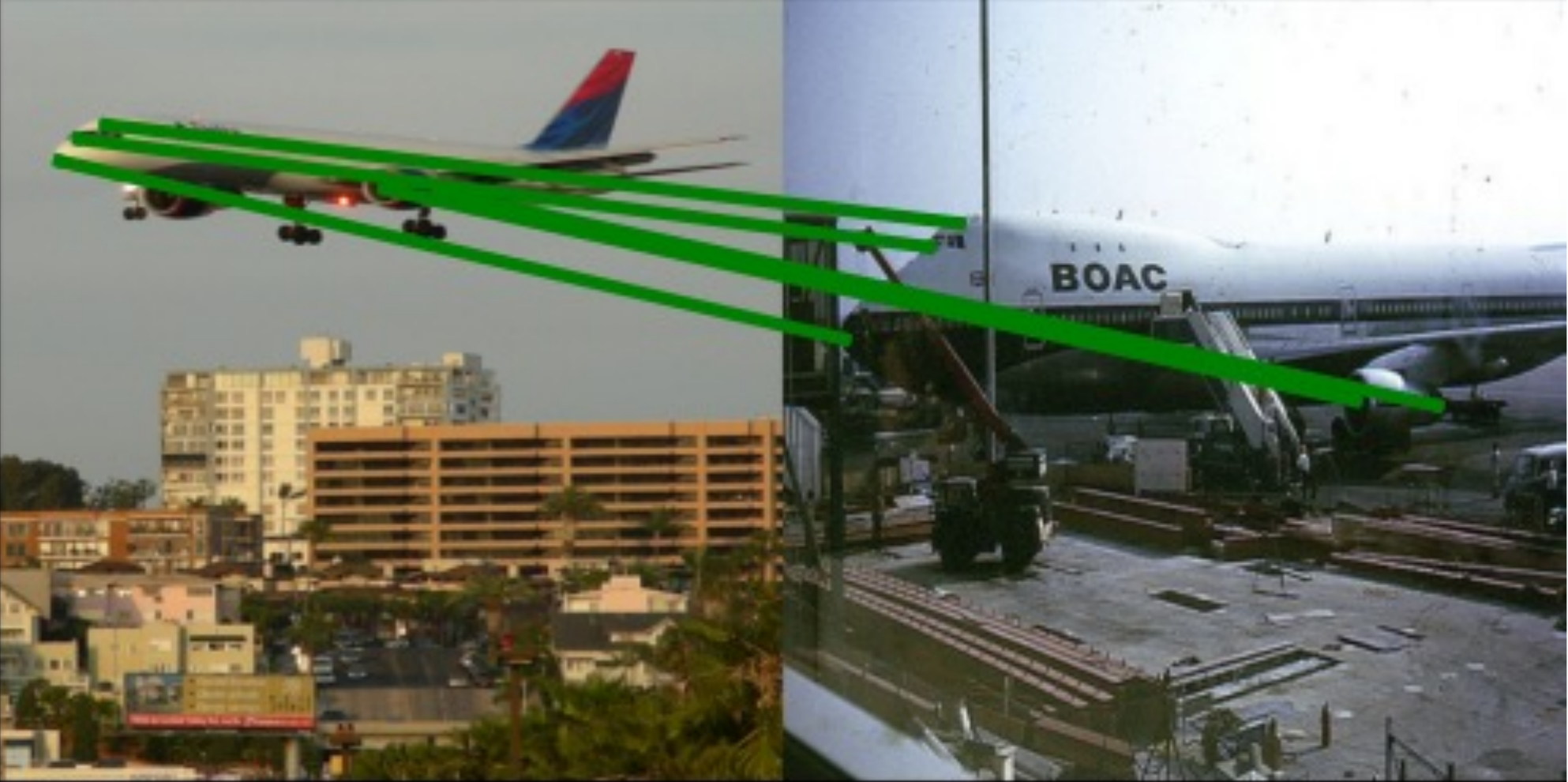}
   \caption{(a)+ \ours (Ours)}
  \end{subfigure}\hfill \\
\vspace{-1em}
  \caption{\textbf{Qualitative results on SPair-71k~\protect\cite{min2019spair} in comparison with other semi-supervised methods:} For a fair comparison, we use (a) the fixed baseline CATs++~\protect\cite{cho2022cats++} for all semi-supervised methods, (b) + CNNGeoU~\protect\cite{laskar2018semi}, (c) + PWarpC~\protect\cite{truong2022probabilistic}, (d) + SCORRSAN~\protect\cite{huang2022learning}, and (e) + \ours. The point-to-point matches are drawn by linking key point pairs with line segments. {Green} and {red} lines denote correct and incorrect predictions with respect to the ground-truth pairs, respectively. We observe that ours performs much better compared with the counterparts across all the sample image pairs.}
  \label{fig:qual_semi}
\end{figure*}

\noindent\textbf{Comparison with State-of-the-arts.}
Alongside the qualitative results shown in Fig.3 in the main paper, we offer further visualizations of example pairs with their predicted matches for \ours and the highly competitive methods in both the supervised and semi-supervised regimes: CATs~\cite{cho2021semantic}, CATs++~\cite{cho2022cats++}, SemiMatch~\cite{kim2022semimatch}, SCORRSAN~\cite{huang2022learning}. As shown in Fig.~\ref{fig:qual_sota_appendix}, our approach produces more accurate estimations of correspondences between image pairs across various object classes and differences in variation factors compared with other methods.

\noindent\textbf{Comparison with semi-supervised methods.}
We show qualitative results to complement the aspect of the controlled experiments for the learning methods section of the main paper. For a fair comparison, all methods are trained under the same network architecture, suggested in CATs++~\cite{cho2022cats++}. As shown in Fig.~\ref{fig:qual_semi}, we predict correct correspondences, even in challenging samples that exhibit significant differences in scale and viewpoint between image pairs, unlike other methods, which tend to produce incorrect predictions for such samples.

\begin{figure*}[t!]
    \begin{subfigure}[b]{0.5\textwidth}
    \centering
\includegraphics[width=1\textwidth]{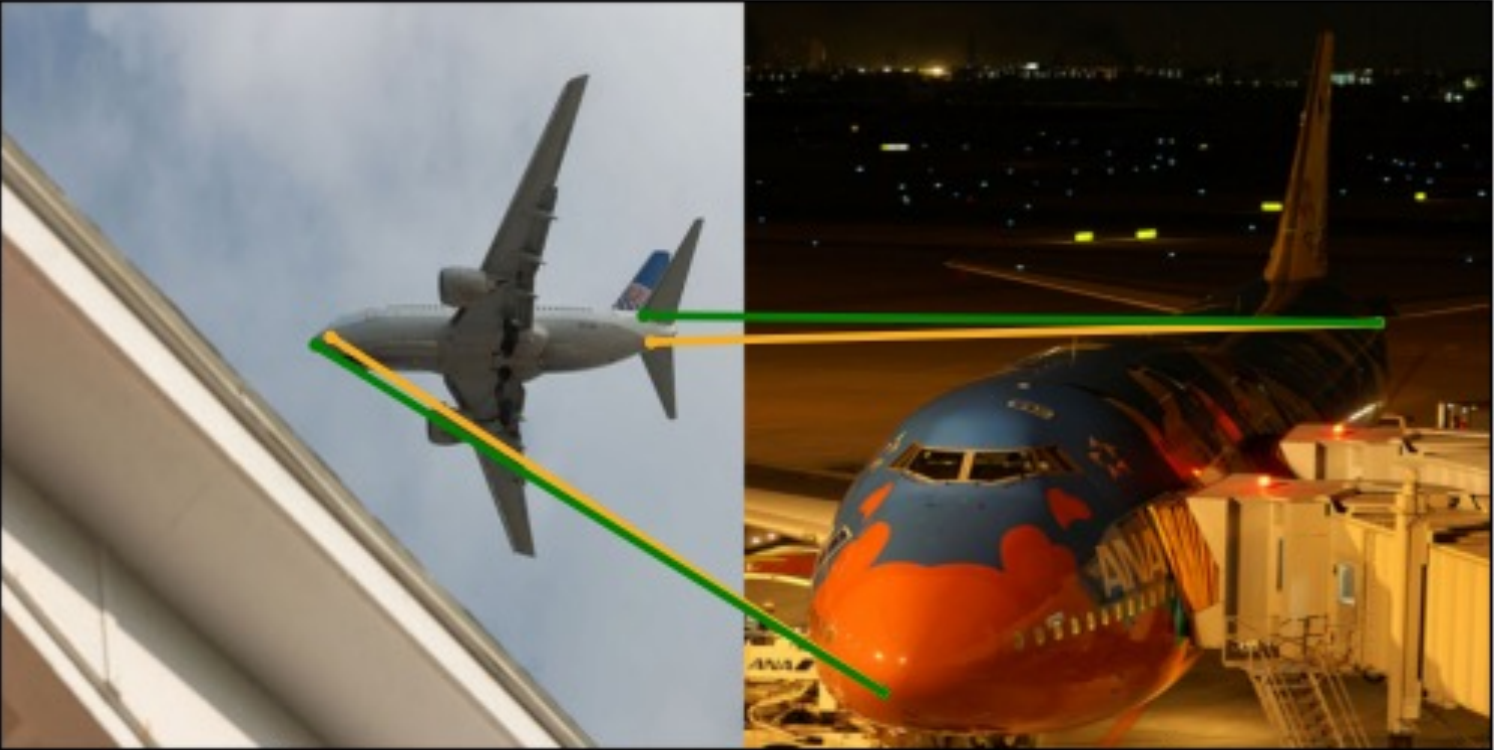} \hfill
\vspace{-1.1em}
\includegraphics[width=1\textwidth]{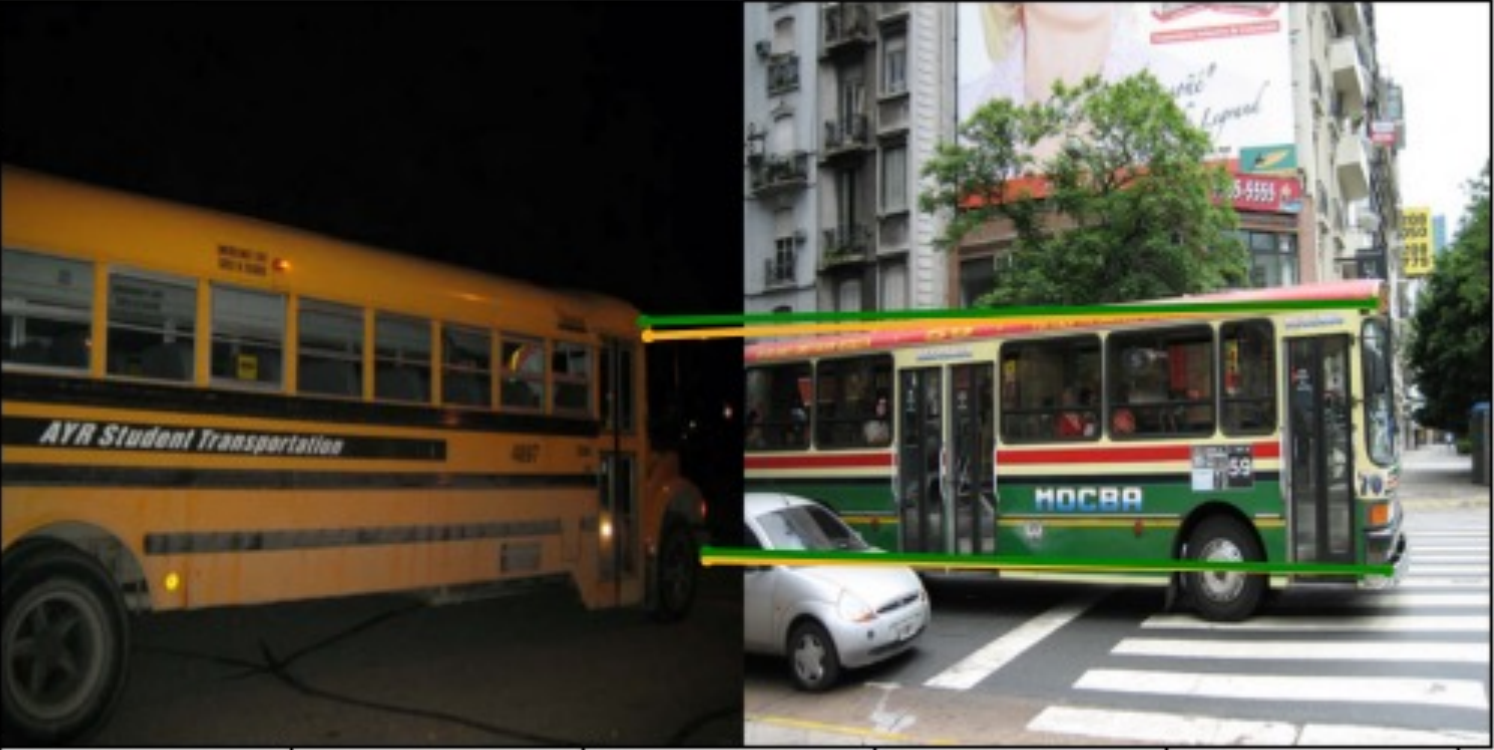} \hfill
\vspace{-1.1em}
\includegraphics[width=1\textwidth]{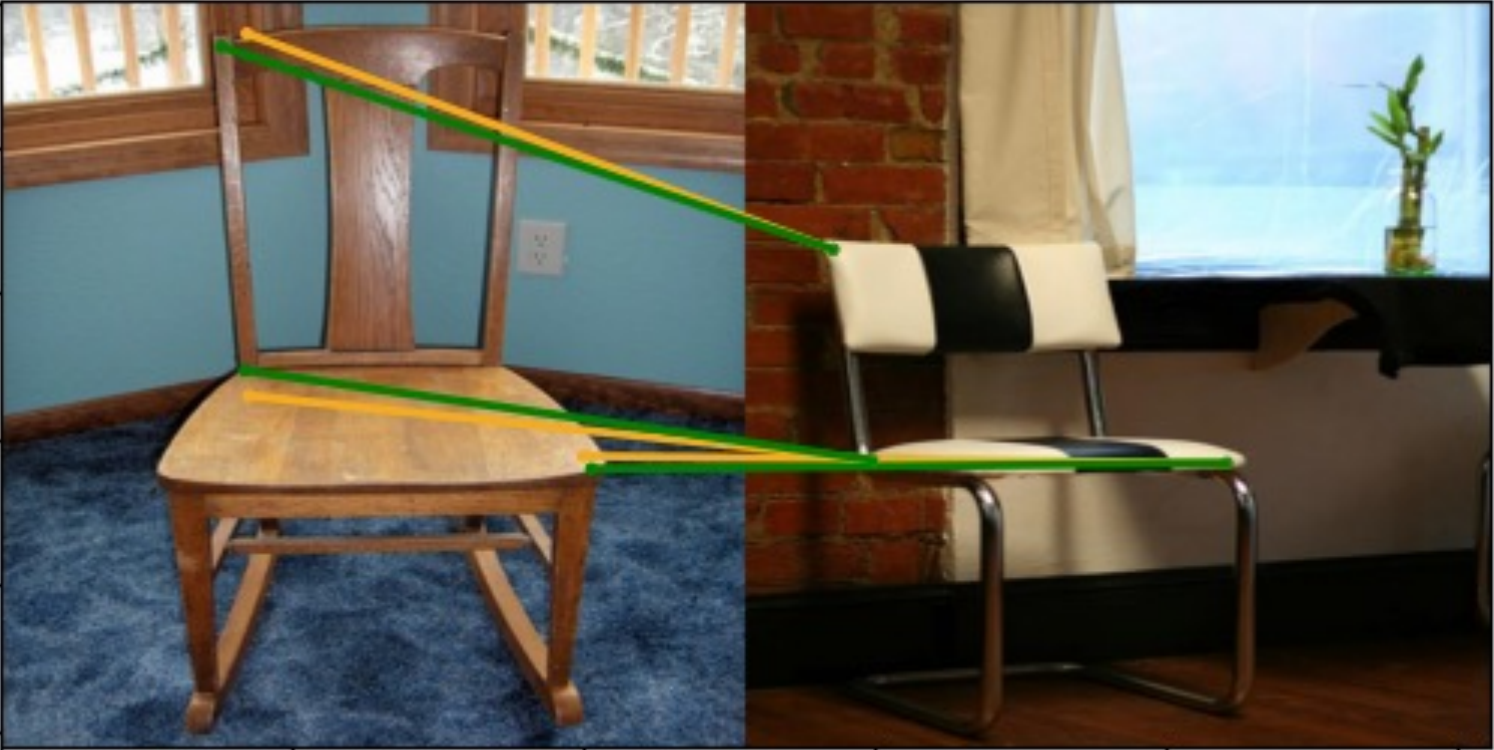} \hfill
\vspace{-1.1em}
\includegraphics[width=1\textwidth]{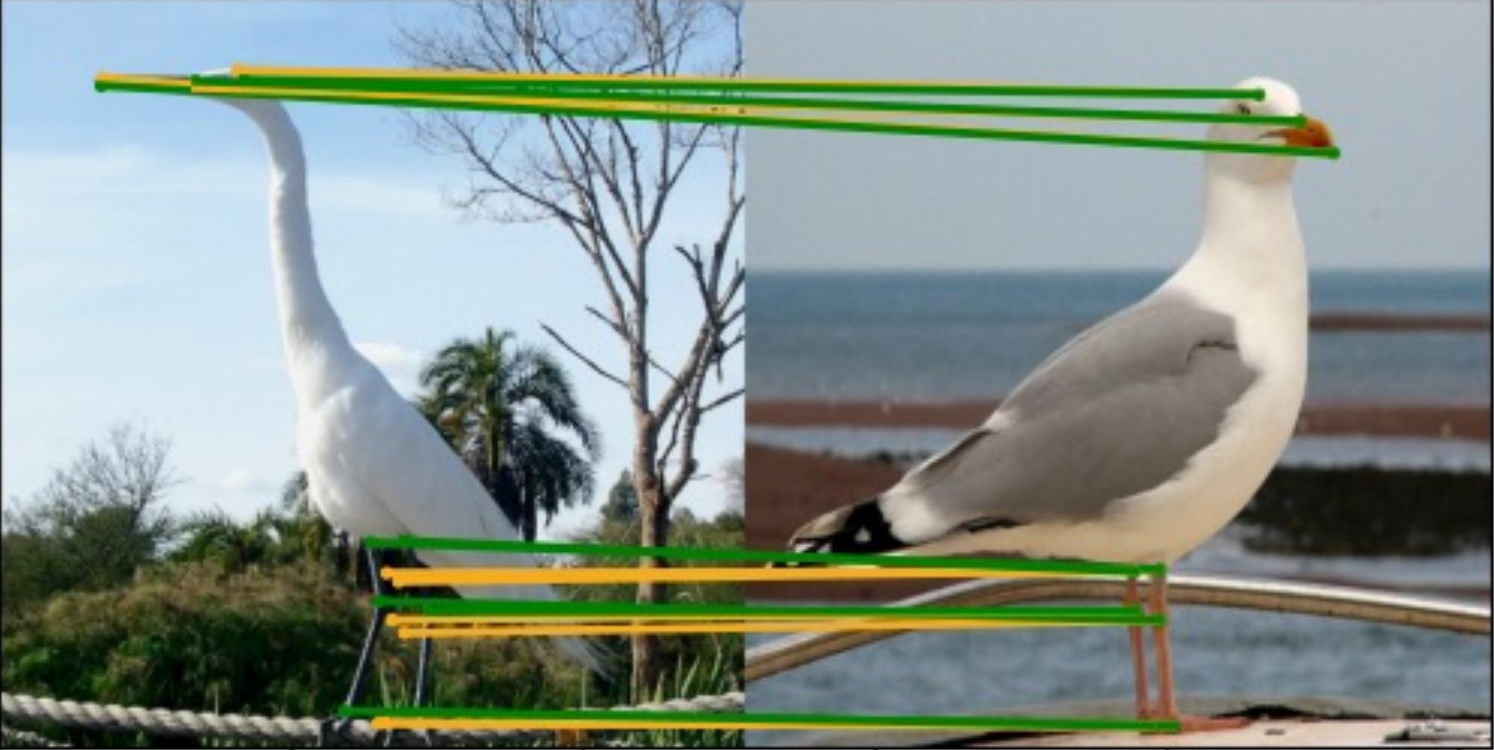} \hfill
\vspace{-1.1em}

        \caption{Baseline}     
            \end{subfigure}%
\begin{subfigure}[b]{0.5\textwidth}
\centering
\includegraphics[width=1\textwidth]{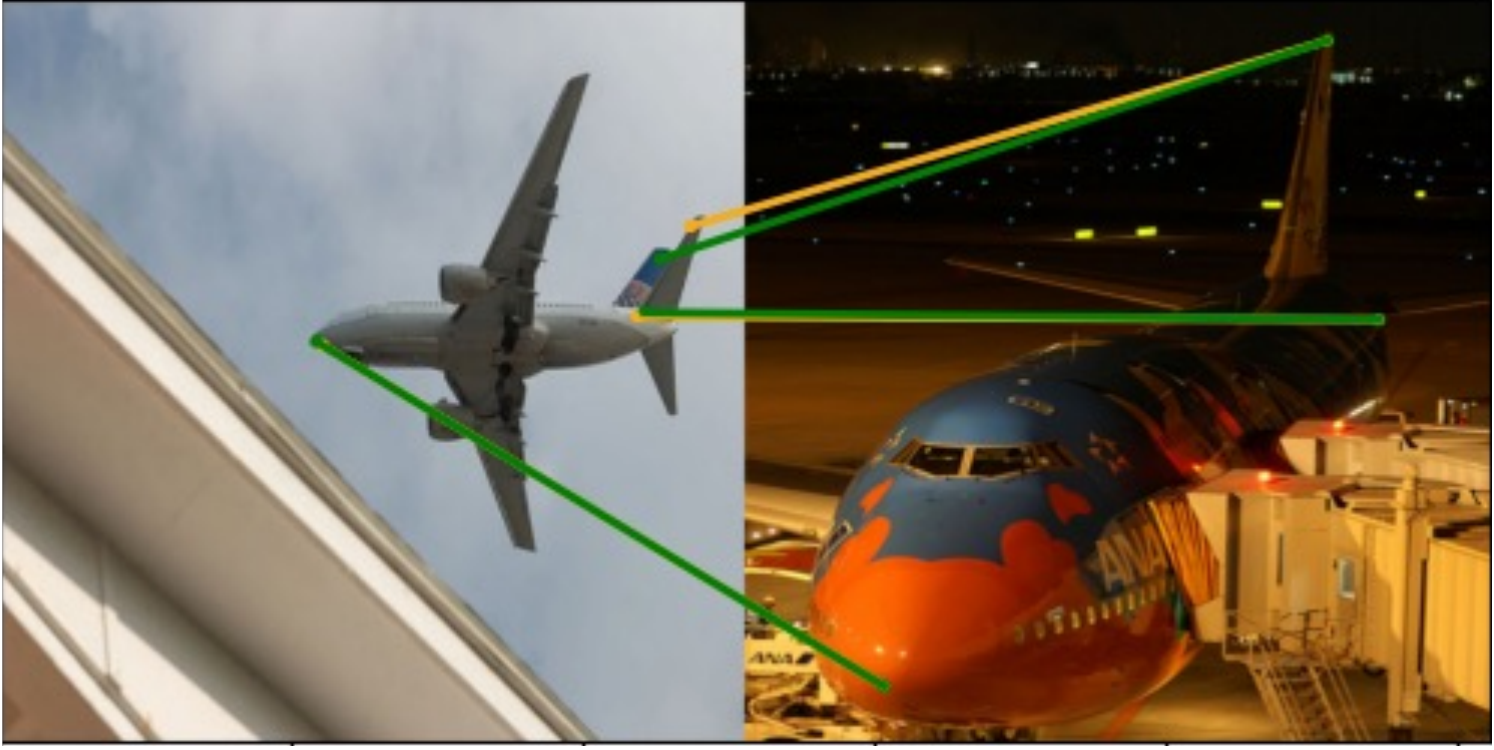} \hfill
\vspace{-1.1em}
\includegraphics[width=1\textwidth]{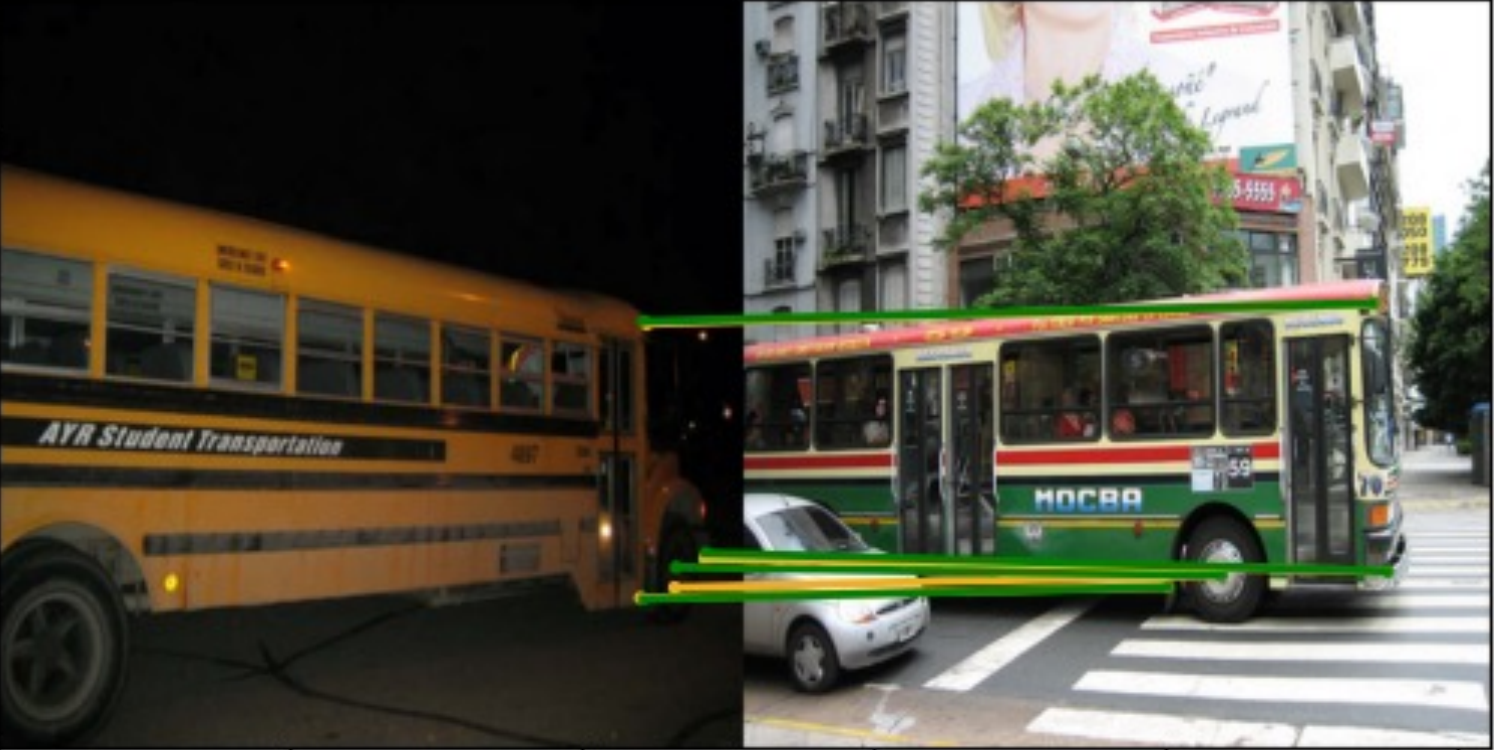} \hfill
\vspace{-1.1em}
\includegraphics[width=1\textwidth]{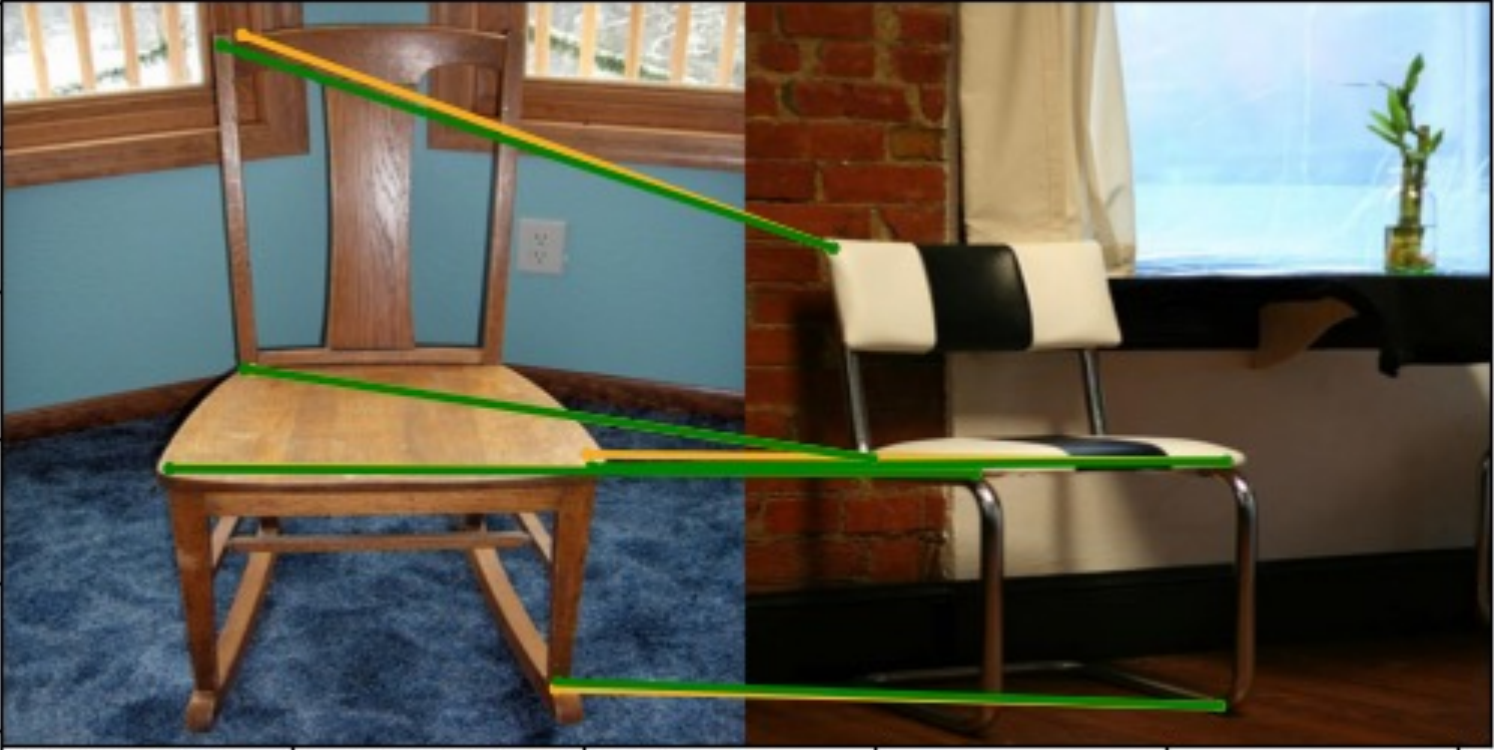} \hfill
\vspace{-1.1em}
\includegraphics[width=1\textwidth]{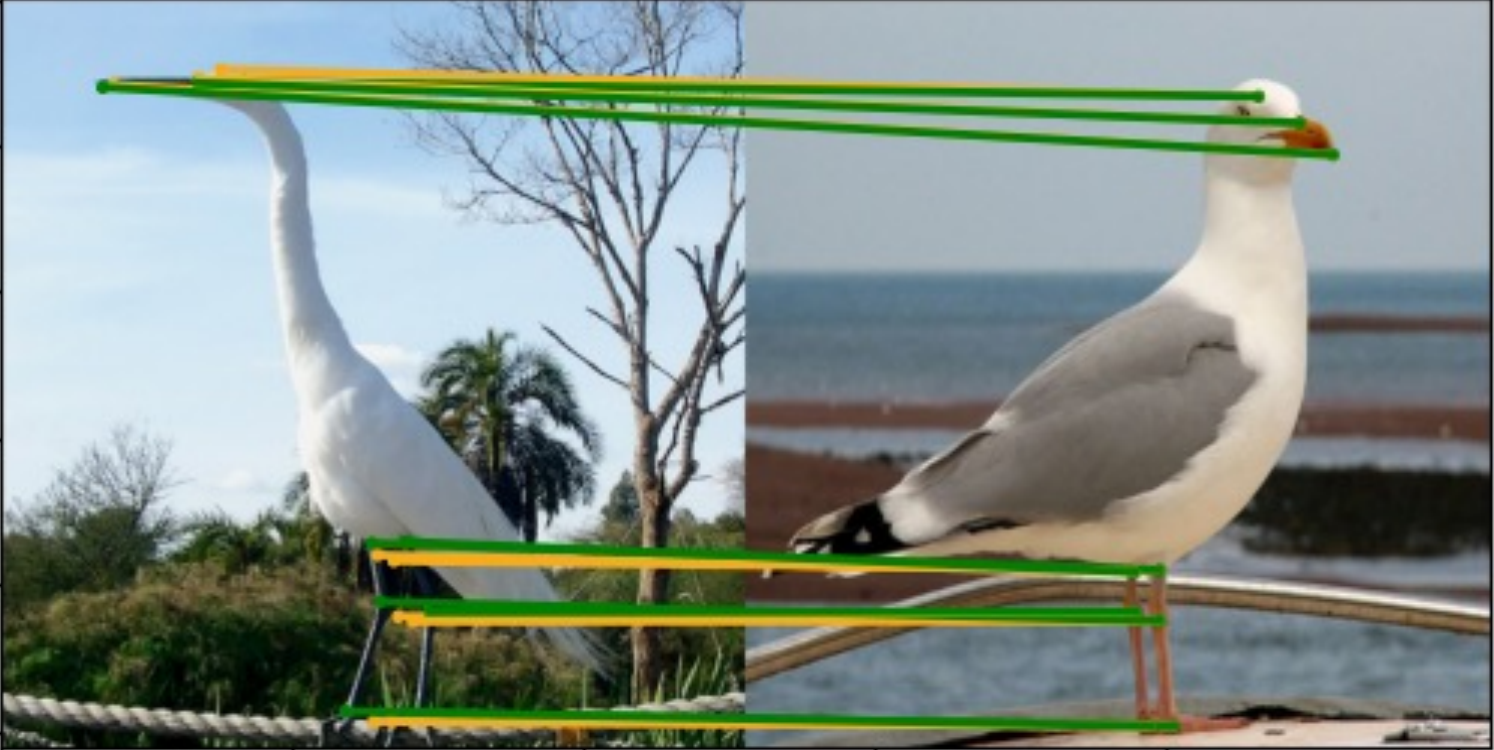} \hfill
\vspace{-1.1em}

        \caption{\ours}                
            \end{subfigure}
\caption{\textbf{Visualization of the difference between correctly predicted points and ground truth (GT) points on SPair-71k.} The GT points in the left images corresponding to the GT points in the right images for each image pair are marked in {green} lines, and the predicted point correspondences are marked in {yellow} lines. The closer the predicted correspondence to the GT correspondence is, the more accurate the prediction. Notice that if only the green line is visible, the predicted and GT point correspondences are perfectly matched.} 
\vspace{-1em}
\label{fig:vis_diff}
\end{figure*}

\begin{figure*}[t!]
    \begin{subfigure}[b]{0.5\textwidth}
    \centering
\includegraphics[width=1\textwidth]{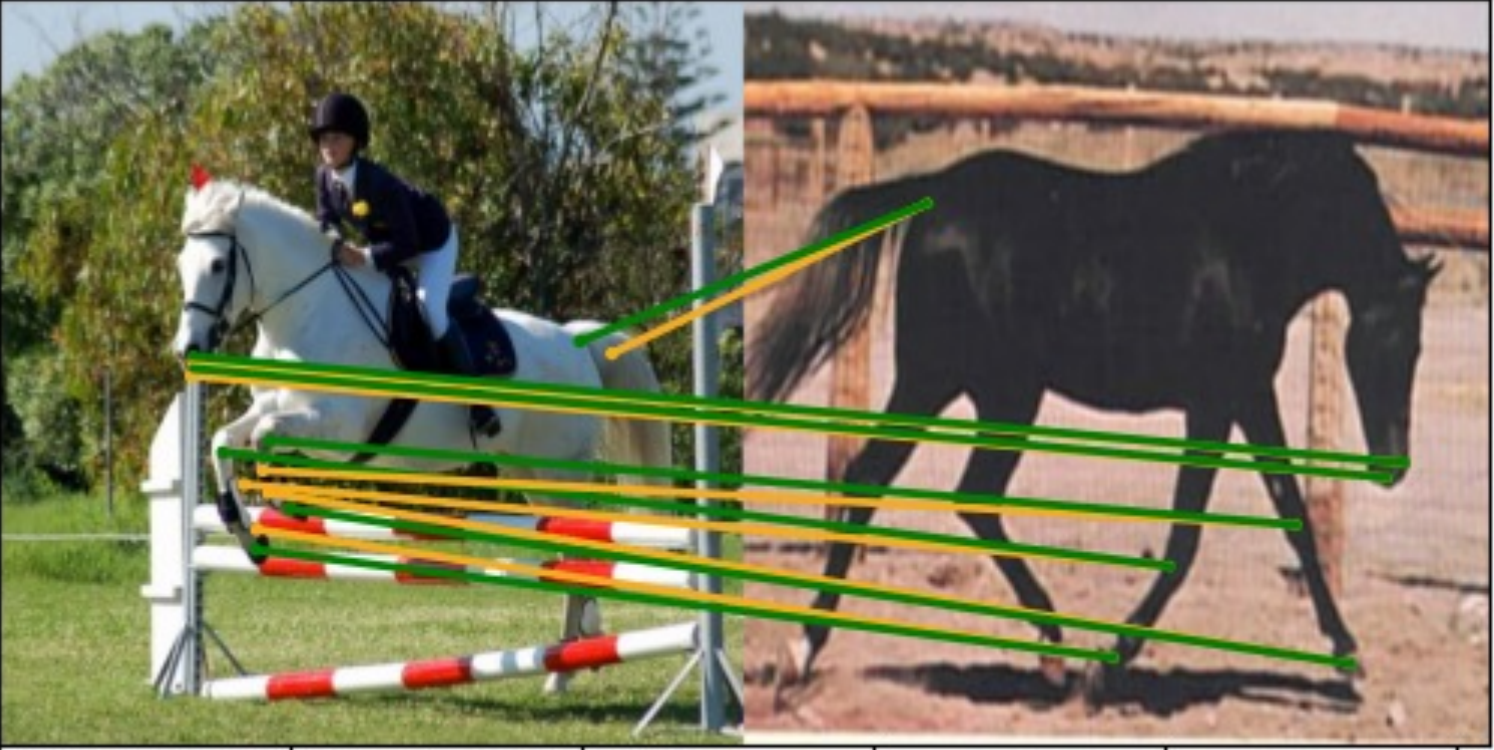} \hfill
\vspace{-1.1em}

\includegraphics[width=1\textwidth]{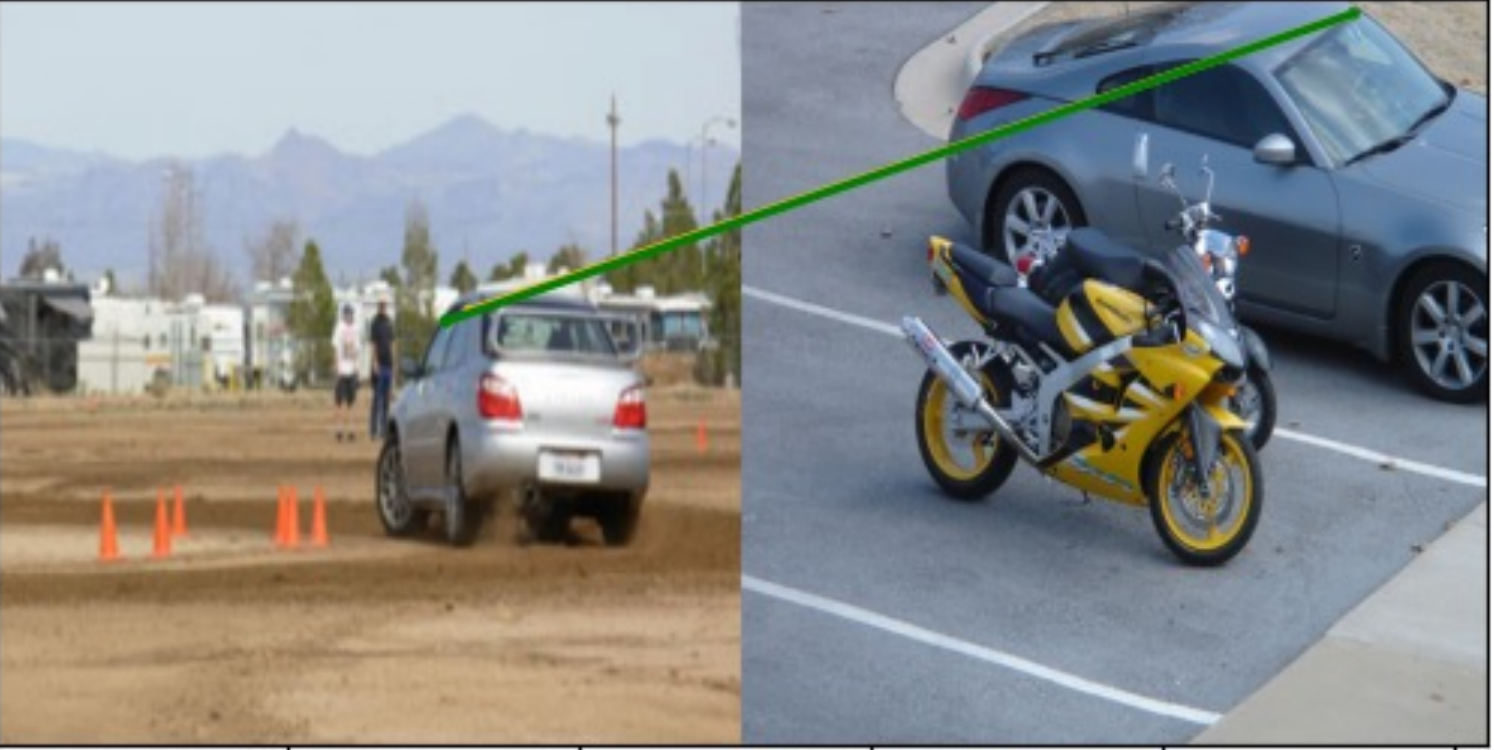} \hfill
\vspace{-1.1em}

\includegraphics[width=1\textwidth]{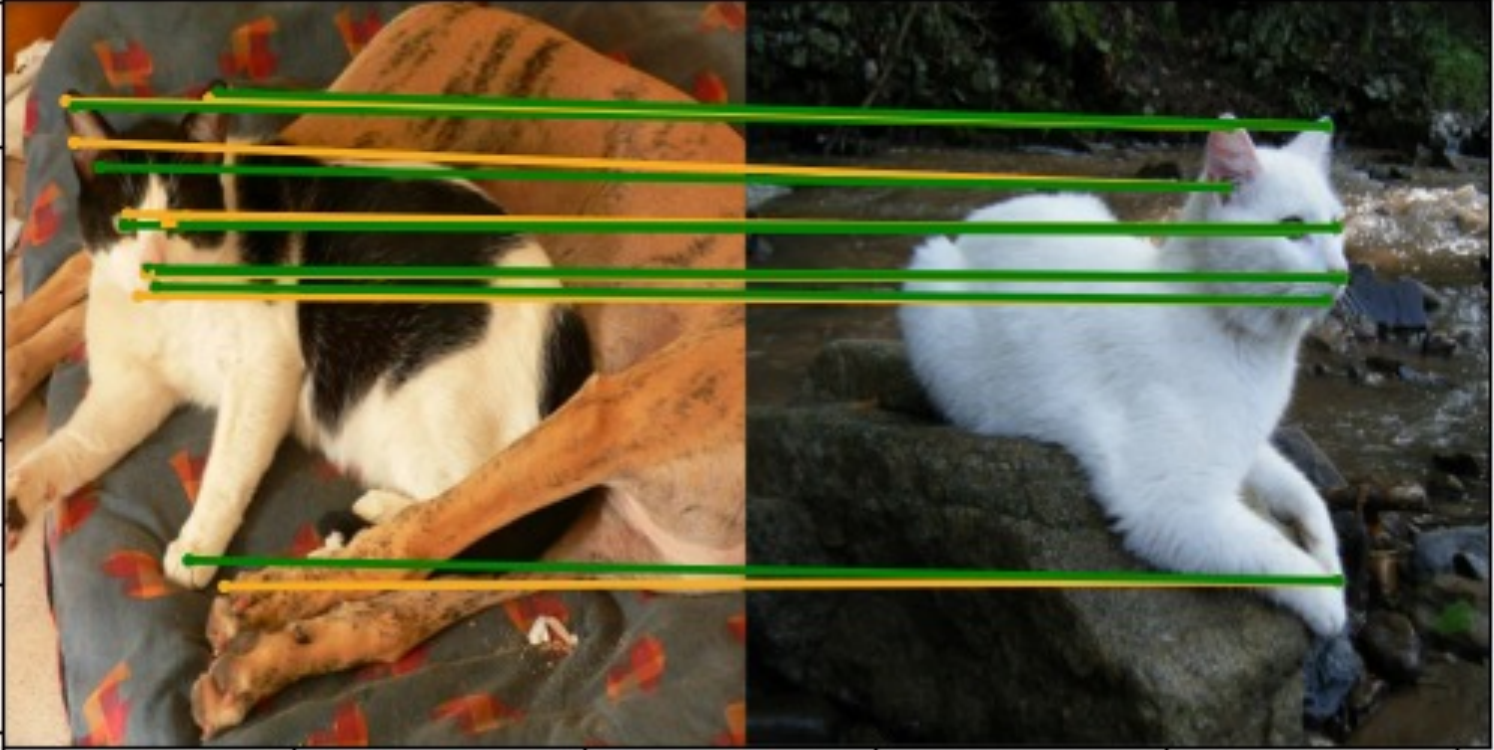} \hfill
\vspace{-1.1em}

\includegraphics[width=\textwidth]{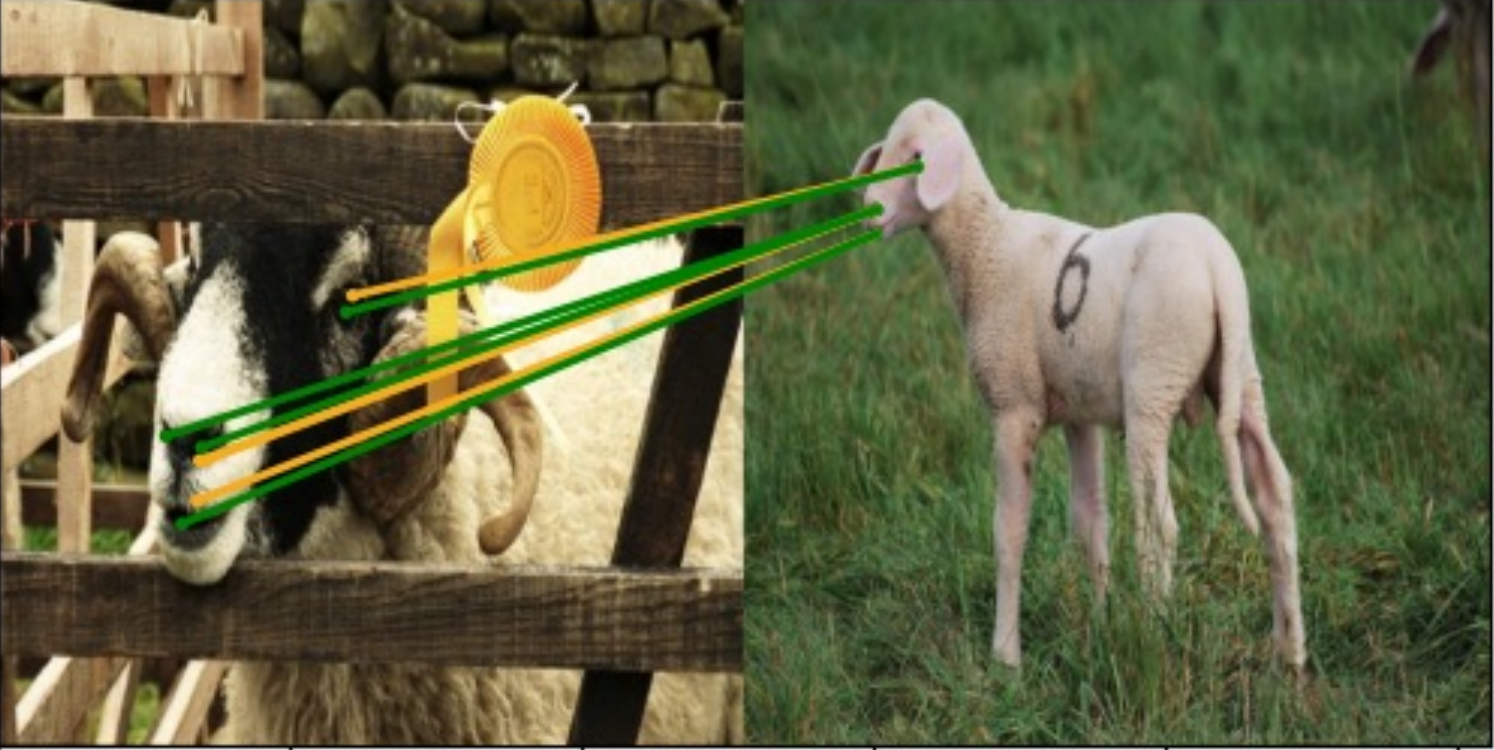} \hfill
\vspace{-1.1em}

        \caption{Baseline}     
            \end{subfigure}%
\begin{subfigure}[b]{0.5\textwidth}
\centering
\includegraphics[width=1\textwidth]{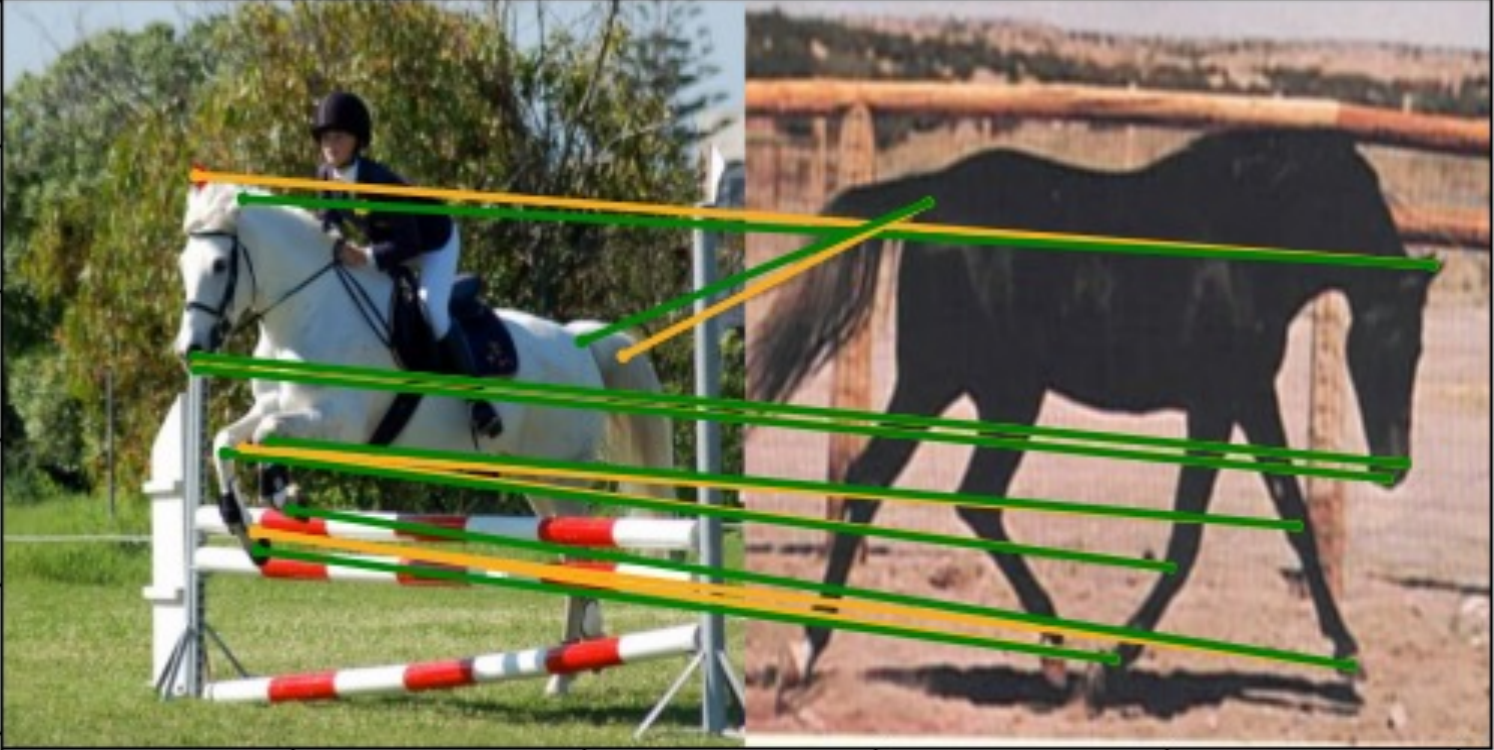} \hfill
\vspace{-1.1em}

\includegraphics[width=1\textwidth]{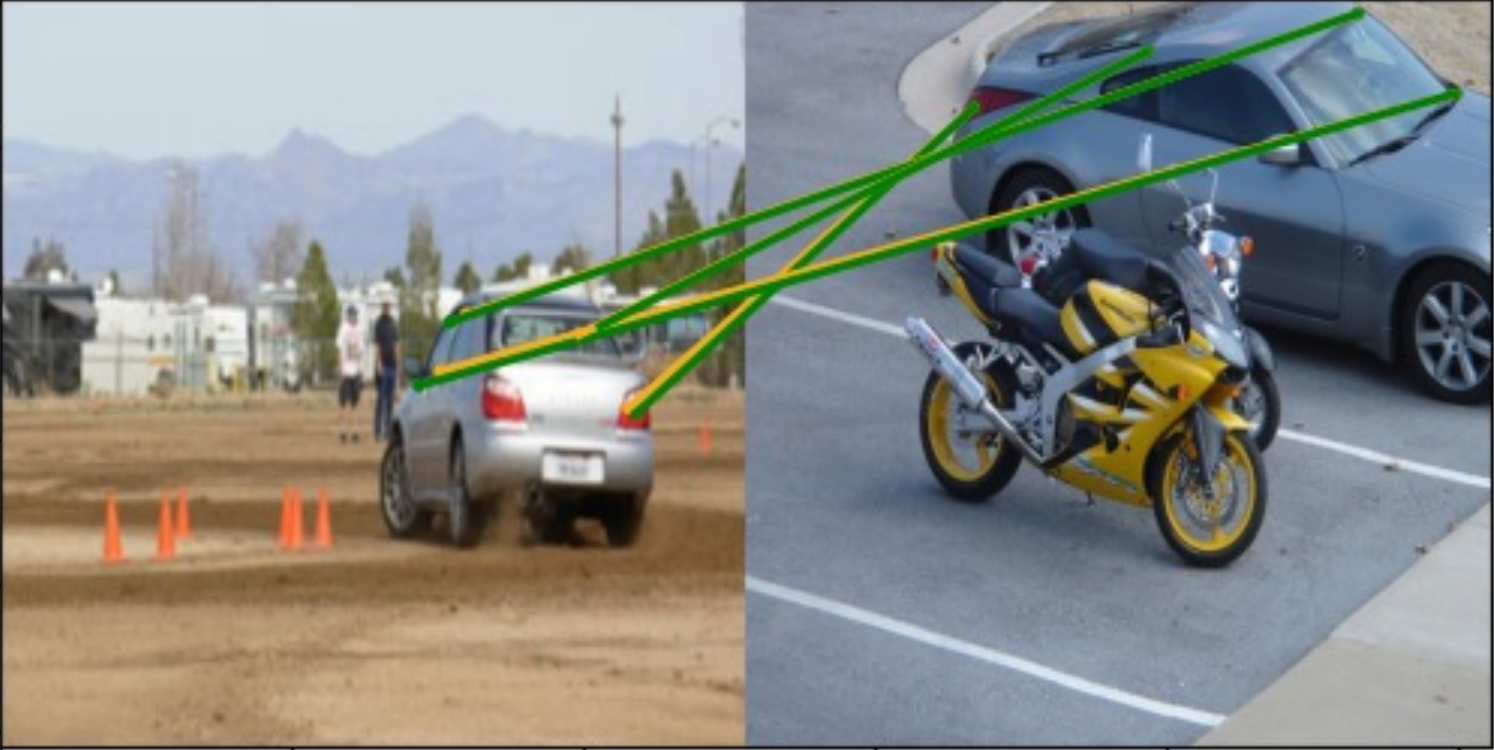} \hfill
\vspace{-1.1em}

\includegraphics[width=1\textwidth]{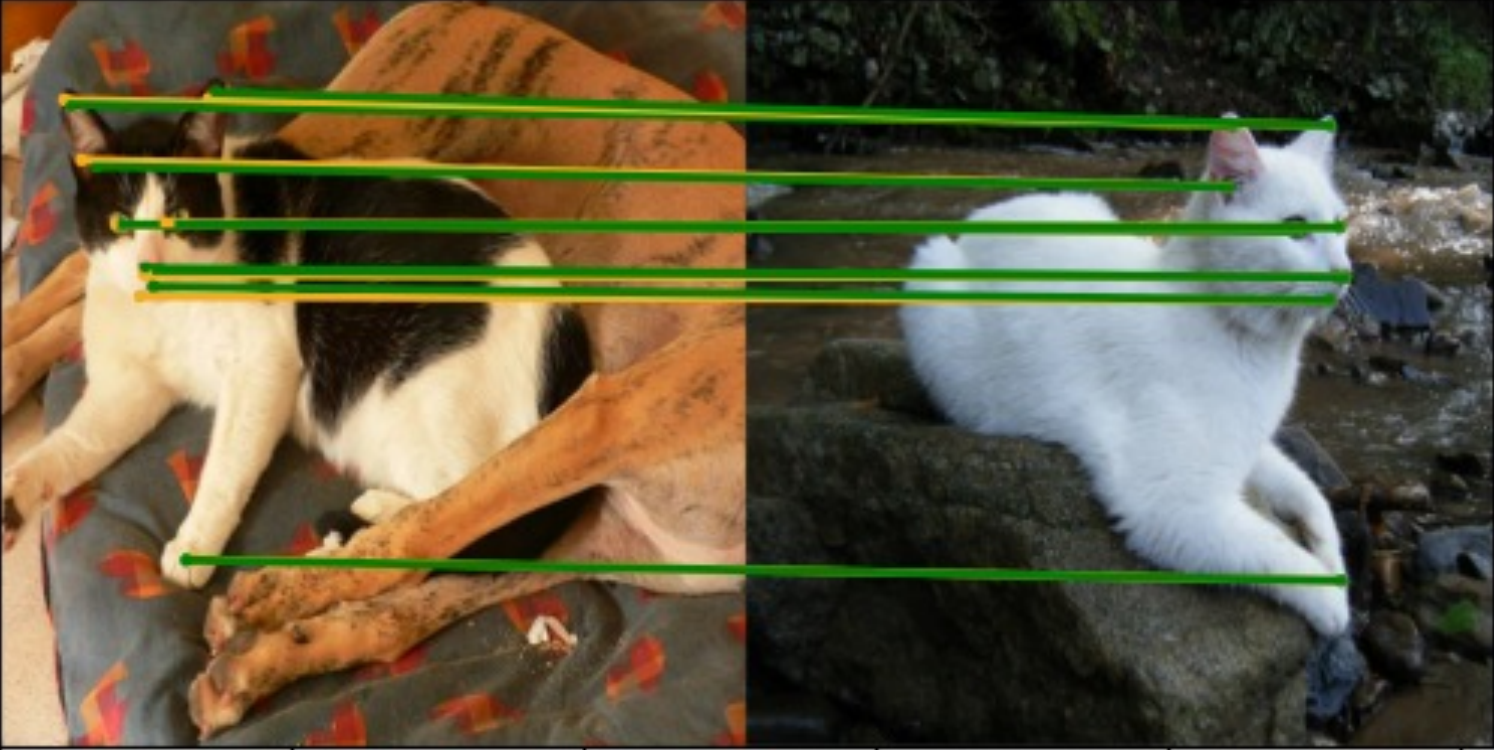} \hfill
\vspace{-1.1em}

\includegraphics[width=\textwidth]{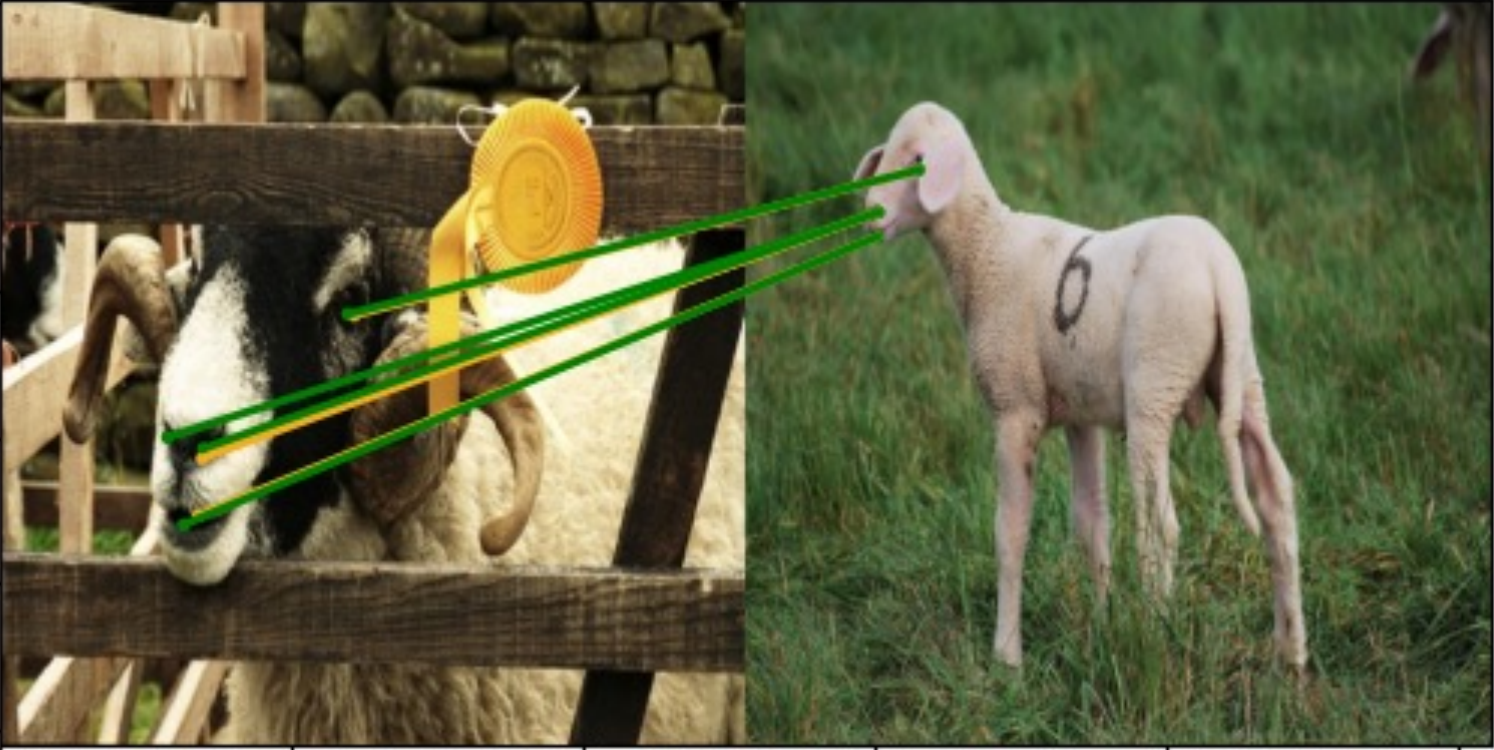} \hfill
\vspace{-1.1em}

        \caption{\ours}                
            \end{subfigure}
    \vspace{-1.5em}
\caption{\textbf{Visualization of the difference between correctly predicted points and ground truth (GT) points on SPair-71k (cont'd).} The GT points in the left images corresponding to the GT points in the right images for each image pair are marked in {green} lines, and the predicted point correspondences are marked in {yellow} lines. The closer the predicted correspondence to the GT correspondence is, the more accurate the prediction. Notice that if only the green line is visible, the predicted and GT point correspondences are perfectly matched.} 
\vspace{-.5em}
\label{fig:vis_diff_2}
\end{figure*}

\noindent\textbf{Qualitative PCK analysis.}
In addition to quantifying model performance through the analysis of PCK values, we demonstrate the superiority of our method by visualizing its predictive quality in Fig.~\ref{fig:vis_diff} and Fig.~\ref{fig:vis_diff_2}. We compare the differences between the correctly predicted and the ground truth (GT) point correspondences at $\alpha = 0.1$, as indicated by yellow and green colors, respectively. This visualization illustrates how many more points our model predicts correctly as well as how closely our model's predictions align with the correct GT key points, even at the extreme points compared to the baseline~\cite{cho2022cats++}. Specifically, the example of the sheep class, having the most minor categorical PCK difference compared to the baseline, illustrates that, even though the PCK achieved by our method is similar to that of the baseline, the quality of the predicted correspondence is superior. 
This outstanding performance can be attributed to the fact that our method generates machine-annotated point correspondences, providing diverse and rare supervisions that are difficult to obtain through the limited amount of manually annotated GT key points.

\end{document}